\pdfoutput=1

\documentclass[10pt,twocolumn,letterpaper]{article}

\usepackage[pagenumbers]{cvpr} %

\usepackage{cuted} %

\usepackage{currfile} %

\usepackage{caption} %

\usepackage{marvosym}

\definecolor{cvprblue}{rgb}{0.21,0.49,0.74}
\usepackage[pagebackref,breaklinks,colorlinks,allcolors=cvprblue]{hyperref}

\def\confName{CVPR}
\def\confYear{2027}

\title{Seeing the Edit: Grounded Reasoning and Intent for Instruction-Based Image Edit Detection}

\author{
Van-Loc Nguyen$^{1,4,*,\dagger}$ \quad
Hongbo Wang$^{2,3,*}$ \quad
AprilPyone MaungMaung$^{3,*}$ \\
Junyu Lu$^{5}$ \quad
Minh-Triet Tran$^{1,4}$ \quad
Isao Echizen$^{2,3,6,\ddagger}$ \\[0.6em]
$^{1}$University of Science, Ho Chi Minh City, Vietnam \quad
$^{2}$The University of Tokyo, Japan \\
$^{3}$National Institute of Informatics, Japan \quad
$^{4}$Vietnam National University, Ho Chi Minh City, Vietnam\\
$^{5}$Dalian University of Technology, China\\
$^{6}$The Graduate University for Advanced Studies, SOKENDAI, Japan\\
{\small $^{*}$Equal contribution} \quad
{\small $^{\dagger}$Work done during the internship at National Institute of Informatics, Japan} \quad
{\small $^{\ddagger}$Corresponding author} \\
{\tt\small nvloc@selab.hcmus.edu.vn, \{hongbo, pyone\}@nii.ac.jp,} \\
{\tt\small dutljy@mail.dlut.edu.cn, tmtriet@fit.hcmus.edu.vn, iechizen@nii.ac.jp}
}

\begin{document}
\maketitle
\begin{strip}
\centering
\includegraphics[width=\linewidth]{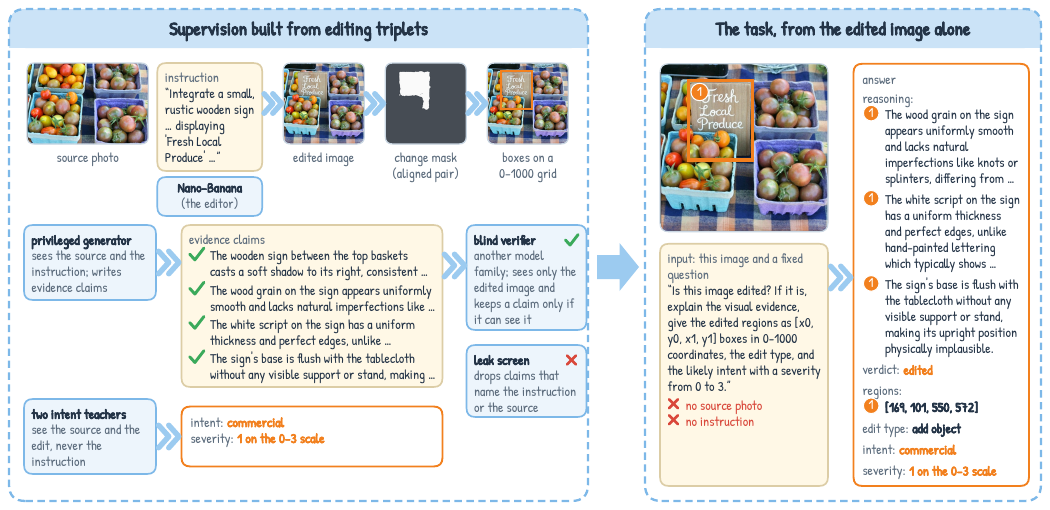}
\captionsetup{hypcap=false}
\captionof{figure}{\textbf{Overview.} Left: supervision from one Pico-Banana-400K~\cite{qian2026pico} triplet, with claims kept only when a blind verifier sees them in the edit. Right: the task, from the edited image alone.}
\captionsetup{hypcap=true}
\label{fig:\currfilebase}%
\end{strip}
\begin{abstract}

Instruction-based editors change a photo from one sentence, and the result looks like a real photo with one thing changed. Because they regenerate the whole image, the local traces that forensic detectors look for largely disappear, and a verdict alone says neither what was changed nor why. To answer both questions, we formulate edit detection with four outputs from the edited image alone: a verdict, the edited regions, a short reasoning grounded in them, and the edit's plausible intent and harm. Earlier explainable detectors predict no intent. We build the supervision automatically from Pico-Banana-400K: a generator that sees the source and the instruction writes evidence claims, a blind verifier keeps those it sees in the edit, and teachers that never see the instruction label intent. It covers 253K edits in five tiers, 242K with verified reasoning, and 171K authentic photos, with held-out edit types and a human audit. On its local and regional edits, no released detector exceeds 58 AUC; fine-tuned on them, a 3B model reaches 93.1 and forensic detectors 98.7--100. Three findings follow. A trace test with identical processing shows that the fine-tuned detectors respond mainly to the editor's whole-image trace: they detect the editor's output even with the edit undone, and they fail on other editors. A model that reasons before its verdict falls near chance although its verdict rests on the cited regions; writing the verdict first with a reweighted loss reaches 87.2 with all four outputs. Intent is predicted above the majority class on seen types only. Detection with reasons is thus within reach on one editor, but it does not yet carry to other editors, nor does intent to new edit types. Code and pipeline are released. Project page: \href{https://seeing-the-edit.vercel.app}{https://seeing-the-edit.vercel.app}.

\end{abstract}    
\section{Introduction}
\label{sec:intro}

An instruction-based image editor changes a photo from a sentence: remove the sign, make the man older, put the scene in winter, etc. The result is a real photo with one thing changed, and the question it raises is not only whether it was edited but \textit{what was changed} and \textit{what the change is for}. A removed warning sign and a swapped dessert are both edits; only one of them matters if the image is believed. A verdict alone answers neither question, and a mask answers only the first. 

These editors also break the assumption behind most forensic detectors. A splice or an inpainting leaves local traces, a seam, a resampling pattern, a second compression, and detectors are trained to find them~\cite{Chen2021MVSS,Guillaro_2023_CVPR,kong2025pim}. An instruction-based editor regenerates the whole image, so those traces largely disappear: with their released weights, no detector, forensic MLLM or zero-shot MLLM we test reaches 60 AUC on our edits (\cref{sec:exp-indomain}). What regeneration leaves instead is the editor's own trace over the whole image. Detectors fine-tuned on one editor learn it, reach 93--100 AUC on that editor's edits and fall to 41--70 on another editor's edits of the same photos (\cref{sec:exp-generalization}).

Existing datasets do not ask. Forensic benchmarks give labels and masks~\cite{dong2013casia,hsu2006detecting,wen2016coverage}; editing datasets give the triplet of source, instruction and edit to train editors~\cite{brooks2023instructpix2pix,zhang2023magicbrush,qian2026pico}; explainable forensic models write a rationale for their verdict~\cite{xu2025fakeshield,Huang_2025_CVPR,huang2026sofakebenchmarkingexplainingsocial}, but the rationales they train on are not checked against the image, and none of them predicts the purpose of an edit.

We formulate image edit detection as a task with four outputs from the edited image alone: a verdict, the edited regions as boxes, a short reasoning whose claims point to those regions, and the plausible intent of the edit with the harm it could do if believed (\cref{sec:task-formulation}; \cref{fig:teaser}, right). The model never sees the instruction or the source photo, because neither is available when a suspicious image is found. We build the supervision automatically from Pico-Banana-400K~\cite{qian2026pico} (\cref{sec:automatic}; \cref{fig:teaser}, left): masks and boxes from the aligned pair; claims written by a generator that sees the source and the instruction, then kept only when a blind verifier from another model family can see them in the edited image, with claims that leak the instruction or the source removed; and intent labels from two teachers that never see the instruction, whose polite wording would bias every label toward benign. The two classes are processed nearly alike, so that size, format and compression do not give the label away. The data covers 253K edits of 31 trainable and 4 held-out edit types in five tiers, 242K of them with verified reasoning, and 171K authentic photos with rationales of their own, and humans audit the masks, the reasoning and the reproducibility of intent judgments (\cref{sec:annotation}).

On its 106K local and regional edits, we benchmark three groups of existing methods and fine-tune one 3B model~\cite{bai2025qwen25vl} in variants that differ in what they are trained to write (\cref{sec:models}). We use them to ask what a detector learns from such data, and report three findings (\cref{sec:experiments}). \emph{Detection on one editor is not edit detection.} Trained to write the verdict and the boxes, the model reaches 93.1 AUC on edits of the training types and 91.3 on held-out types, 35 points above the best existing detector. But a trace test with identical processing shows that it, and far more so the fine-tuned detectors, respond to the editor's whole-image trace: on a second editor's edits of the same photos it keeps 66.9 and they range from 41 to 70, and all of them fail on MagicBrush's DALL-E 2 edits~\cite{zhang2023magicbrush} (\cref{tab:transfer}). \emph{Reasoning decides the verdict without improving it.} Trained to write its reasoning before the verdict, the same model falls to chance, because the reasoning fixes the verdict early, and writing the verdict first recovers most of the loss only with a reweighted training loss; a region-masking test shows that its verdict does rest on the regions it cites, including the ones it hallucinates on authentic photos, and a rationale swap shows that the verdict follows the reasoning, not the image. \emph{Intent does not transfer.} It is predicted above the majority class on seen edit types (64.2\% against 58.6\%) and not on held-out types.

Our contributions are: (1) a task formulation and an automatic pipeline that writes reasoning checked against the edited image and intent labels that never depend on the instruction; (2) a dataset of 253K edits in five tiers with masks, boxes, reasoning and intent, with held-out edit types and a human audit; (3) a benchmark that tests existing detectors and MLLMs on these edits, on held-out edit types and on the edits of other editors; and (4) a trace test that separates the edited content from the editor's whole-image trace, training variants that locate the cost of grounded reasoning, and interventions showing that a model's verdict rests on the evidence it cites. We release the code and the pipeline; the images must be rebuilt from Pico-Banana under its license (\cref{sec:limitations}).

\section{Related work}
\label{sec:related-work}

\paragraph{Image manipulation detection and localization.}
\label{sec:related-imdl} Classic benchmarks pair spliced, copied or inpainted photos with binary masks of the tampered regions~\cite{dong2013casia,hsu2006detecting,wen2016coverage}. Image forgery detection and localization (IFDL) methods trained on them look for low-level traces that a local manipulation leaves behind~\cite{Chen2021MVSS,Guillaro_2023_CVPR,Guo_2023_CVPR,kong2025pim}. Detectors of fully generated images instead look for traces of the generator, such as diffusion reconstruction error~\cite{wang2023dire}. An instruction-based editor sits between the two cases: it starts from a real photo but regenerates the whole image.

\paragraph{MLLM-based explainable forensics.}

\label{sec:related-mllm}

Recent detectors build on multimodal large language models (MLLMs) to explain their verdicts. FakeShield explains and localizes tampering in edited photos~\cite{xu2025fakeshield}, SIDA detects, localizes and explains social-media deepfakes~\cite{Huang_2025_CVPR}, and So-Fake-R1 is trained with reinforcement learning (RL) to explain and localize forgeries~\cite{huang2026sofakebenchmarkingexplainingsocial}. RL with format and box rewards is now the common recipe for such models~\cite{udayanga2026vlmedits,wang2026forgeryvcr}; we apply it to our grounded-reasoning variant RG, which gives RG+RL (\cref{sec:models-variants}). For fully synthetic images, LEGION, FakeXplain and FakeVLM explain artifacts, the first two grounded in the image~\cite{kang2025legion,ji2026fakexplain,wen2025spot}, and LOKI benchmarks MLLMs as detectors~\cite{ye2025loki}. These explanations describe what looks wrong; none predict what the edit is for or how much harm it could do. Our reasoning also differs in how it is made: each claim is checked by a blind verifier from another model family, and claims that leak the instruction or the source photo are removed (\cref{sec:automatic}).

\paragraph{Instruction-based editing datasets.}
\label{sec:related-editing}

Editing datasets pair a source photo, an instruction and the edited result to train editors: InstructPix2Pix generates such triplets synthetically~\cite{brooks2023instructpix2pix}, MagicBrush annotates multi-turn edits by hand~\cite{zhang2023magicbrush}, and UltraEdit, Emu Edit and HQ-Edit scale them up~\cite{zhao2024ultraedit,sheynin2024emu,hui2024hq}. Pico-Banana-400K edits Open Images~\cite{Kuznetsova2020} photos with Nano-Banana under a taxonomy of edit types~\cite{qian2026pico}, and our edits come from it (\cref{sec:automatic}). We use these triplets in reverse: the edited image is the input, and the instruction and source photo are withheld (\cref{sec:task-formulation}). MagicBrush's hand-annotated edits also serve as a cross-dataset test. EditScout trains an MLLM to localize diffusion-based edits on MagicBrush and tests on edit types it did not see~\cite{nguyen2024editscout}; it has no detection verdict, reasoning or intent, and its edits come from older inpainting editors. \Cref{sec:supp-related} reviews further related work, including harm and intent in synthetic media~\cite{luo2021newsclippings,shao2023dgm4,liu2025mmfakebench,marchal2024misuse}.

\section{Task formulation}
\label{sec:task-formulation}

\subsection{Input}
\label{sec:task-input}

The model sees one image and a fixed question (\cref{sec:supp-prompt-ours}). It does not see the edit instruction, which names the edit, or the source photo; neither is available when a suspicious image is found online. Edited and authentic images are standardized to the same size tiers and the same JPEG encoding (\cref{sec:automatic}), so file properties do not reveal the class; \cref{sec:supp-limitations} states where the processing of the two classes still differs.

\subsection{Outputs}
\label{sec:task-outputs}

The task fixes the outputs, a reasoning and a structured answer, not their order, which we vary (\cref{sec:models-variants}). For an image $I$ the output contains:
\begin{enumerate}
\item \textbf{Detection:} a verdict $v \in \{\text{edited}, \text{authentic}\}$.
\item \textbf{Localization:} for an edited image, the edited regions as boxes $B = \{b_1, \dots, b_K\}$ on a $0$--$1000$ grid over the image, one per region of the edit mask (\cref{sec:automatic}).
\item \textbf{Grounded reasoning:} a few short claims about visible evidence; a claim about a region points to its box.
\item \textbf{Intent:} the plausible uses of the edit, not its author's motive, among seven classes (benign aesthetic, commercial, privacy, misinformation, document or text tampering, identity manipulation, humor or meme; not exclusive), the most plausible one, and a severity $s \in \{0, 1, 2, 3\}$. Severity is the potential harm if the edited image were believed real (\cref{sec:supp-intent}). The reference labels are written by two teacher MLLMs of different families, which see the source photo and the edit but not the instruction (\cref{sec:automatic}).
\end{enumerate}
The structured answer also names the edit type, one of the 13 seen in training, with its category (object, scene, text or human); for an edit of a held-out type (\cref{sec:exp-setup}) no answer is correct, so the type is not scored there. For an authentic image the output is the verdict and a reasoning that explains why nothing looks edited. The experiments use local (L) and regional (R) edits, which have a region to localize; whole-image edits (a filter, a change of lighting or style) are tested only in \cref{sec:supp-gsx}.

\section{Automatic data pipeline}
\label{sec:automatic}

\begin{figure*}[t]
    \centering
    \includegraphics[width=\textwidth]{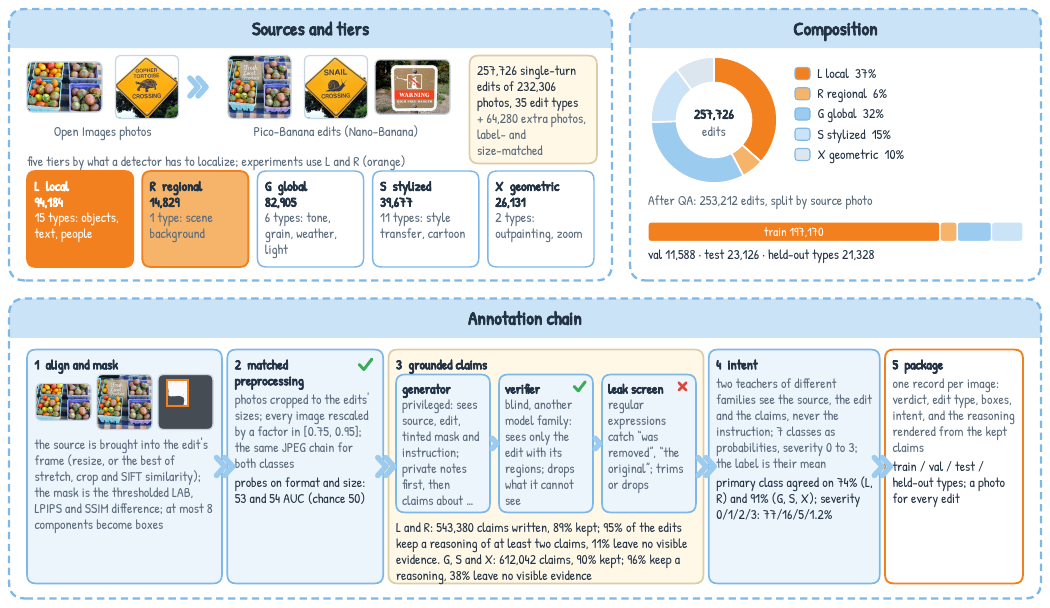}
    \caption{\textbf{Data pipeline and composition.} Top left: the edits come from Pico-Banana-400K, which edits Open Images photos with Nano-Banana, and the authentic photos are the edits' own sources plus extra Open Images photos; the 35 edit types fall into five tiers by what a detector has to localize, of which the local (L) and regional (R) tiers are used in the experiments. Top right: the edits by tier, and the edits that pass the quality checks, by split. Bottom: the annotation chain of \cref{sec:automatic}: the source is aligned to the edit and their difference becomes the mask and the boxes; both classes get the same processing so that no size or format cue remains; a privileged generator writes claims that a blind verifier and a leak screen filter; two teachers label the intent; the records are packaged per split with one photo per edit.}
    \label{fig:pipeline}
    \end{figure*} The pipeline (\cref{fig:pipeline}) turns an editing dataset's triplets, a source photo, an instruction and the edited image, into supervision for the four outputs of \cref{sec:task-outputs} without a human in the loop; \cref{sec:annotation} audits its output. \Cref{sec:supp-pipeline} gives the details of each step.

\paragraph{Source and tiers.}
\label{sec:data-source} Our edits come from Pico-Banana-400K~\cite{qian2026pico}, which edits Open Images photos~\cite{Kuznetsova2020} with Nano-Banana from instructions of 35 edit types; we use its single-turn edits, 257,726 after de-duplication. The experiments use the local (L) and regional (R) tiers, whose edits have a region to localize (\cref{sec:task-outputs}); the pipeline gives the other tiers (\cref{fig:pipeline}) mostly full-image masks, and they enter only the tests of \cref{sec:supp-gsx}. Splits are grouped by source photo, so no scene appears in two splits, with 85\%, 5\% and 10\% of the edits in train, validation and test. Four edit types are held out: every photo with an edit of a held-out type goes to the test pool, its held-out edits form a separate test split and its other edits stay in test. Three of the four types (expression, relocate object and text font) are local, and they form the \emph{Unseen} set of \cref{sec:exp-setup}. Edits whose files, registration or mask failed are excluded (1.8\%), which leaves 253,212 edits, 106,274 of them in L and R (\cref{sec:supp-pipeline}, \cref{tab:dataset}).

\paragraph{Masks and boxes.}
\label{sec:data-masks} A mask marks where the edit changed the photo. The mask is the thresholded difference of the aligned pair in three views, LAB color, LPIPS~\cite{zhang2018lpips} and SSIM~\cite{wang2004ssim}, from a generator built for this purpose. Boxes come from the mask's connected components after a light opening and closing: at most eight are kept, largest first, each as a box on the $0$--$1000$ grid.

\paragraph{Matched preprocessing.}
\label{sec:data-preprocessing} Nano-Banana writes PNG files in a few fixed sizes, and Open Images photos are camera JPEGs in many sizes, so size, format and compression history would separate the classes before any edit is looked at~\cite{grommelt2024fakeorjpeg}. Each authentic photo therefore draws its output size from the edits' size distribution and is center-cropped to it; every image, edited or not, is resampled by a factor drawn from $U(0.75, 0.95)$, so neither class keeps its native pixel grid; the edits first receive a simulated capture JPEG at a quality drawn from $[85, 98]$; and every image then receives the same final JPEG chain. Before anything is trained, probes on the file format, on bytes per pixel and on width, height, aspect ratio and the stored quantization table must be near chance; after this processing they reach 53 and 54 AUC (the step requires less than 55). On the test sets the processing of the two classes still differs (\cref{sec:supp-limitations}); with identical processing, fine-tuned TruFor keeps 96.4 of its 99.1 AUC and VB 88.0 of 93.1 (\cref{sec:exp-generalization}).

\paragraph{Grounded claims.}
\label{sec:data-claims} The reasoning is a list of claims about visible evidence, each tied to a region, written by a generator with privileged inputs and kept only if a blind verifier agrees. The generator, Qwen3-VL-32B~\cite{bai2025qwen3vltechnicalreport}, sees the aligned source, the edit, the instruction and the edit type. The verifier, InternVL3.5-8B~\cite{wang2025internvl35}, from a different model family, sees only the edited image with its regions drawn, never the source or the instruction, so it can tell whether a claim is checkable from the edit alone; a claim it cannot see is dropped. Leaks of the privileged inputs are caught by a regular-expression screen: a leaking trailing clause is trimmed, otherwise the claim is dropped. Authentic photos get a rationale the same way, from the processed photo alone, with claims that never support an edit and a verifier that checks every factual detail (\cref{tab:dataset}).

\paragraph{Intent.}
\label{sec:data-intent} Two teachers from different families, Qwen3-VL-32B and Gemma-3-27B~\cite{gemmateam2025gemma3}, label each edit's intent. Each sees the source, the edit with its regions outlined, the generator's private notes and its claims, but not the instruction: Pico-Banana's instructions are uniformly polite, and including them pushed every label toward benign.

\section{Annotation quality}
\label{sec:annotation}

The automatic pipeline in \Cref{sec:automatic} provides localization and semantic supervision for edited images, but its checks may still retain annotations that are not supported by visual evidence. The generator has access to the source image and editing instruction, whereas the detector observes only the edited image: knowing what was changed does not establish that the change leaves identifiable evidence. We therefore conduct human audits to examine whether the annotations describe the actual edits and whether their explanations and intent judgments are supported. Masks, reasoning, and intent are assessed on three separate subsets of 200 samples each. The mask subset is stratified equally across the L, R, S, and X tiers (50 samples per tier); we additionally report the L/R subset that matches the localization scope of this paper.

\subsection{Human Audit Results}
\label{sec:audit-results}

\begin{figure*}[t]
    \centering
    \includegraphics[width=\textwidth]{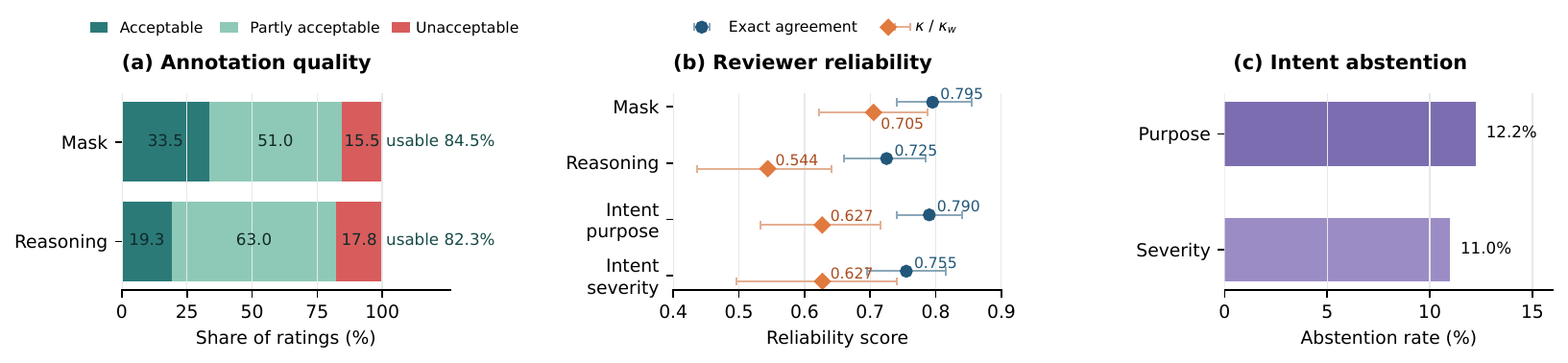}
    \caption{Human audit of masks, reasoning, and intent. (a)~Quality distributions, where usable combines acceptable and partly acceptable ratings. (b)~Exact and chance-corrected inter-reviewer agreement with 95\% bootstrap confidence intervals. (c)~Abstention rates for intent purpose and severity. Mask results aggregate the L, R, S, and X tiers; scope-matched L/R results are reported in the text.}
    \label{fig:human-audit-summary}
\end{figure*}

Figure~\ref{fig:human-audit-summary} separates annotation usability from reviewer reliability. For masks and reasoning, we regard both \emph{acceptable} and \emph{partly acceptable} annotations as usable: the latter retain the principal supervision signal, although with reduced spatial or evidential precision. Each of the three 200-sample subsets is independently evaluated by two reviewers; each intent evaluation comprises separate purpose and severity judgments.

Overall, 84.5\% of mask ratings are usable (33.5\% acceptable and 51.0\% partly acceptable; \cref{fig:human-audit-summary}a), while 15.5\% are unacceptable. Partly acceptable masks typically localize the correct target but contain limited spillover or small omissions. Reviewer agreement is 79.5\%, with $\kappa_w=0.705$ (\cref{fig:human-audit-summary}b). On the 100 L/R samples matching the localization scope of \Cref{sec:automatic}, the usable rate is 79.0\%, agreement is 80.0\%, and $\kappa_w=0.671$. The lower usable rate for L than R (69.0\% vs.\ 89.0\%) indicates that local edits pose the greater localization challenge.

Reasoning annotations are usable in 82.3\% of ratings, including 19.3\% rated fully acceptable, while 17.8\% are unacceptable. Errors arise primarily from over-specific claims about texture, illumination, perspective, or boundaries, rather than from misidentifying the edit itself. Reviewer agreement is 72.5\%, with $\kappa_w=0.544$.

For intent, reviewers agree on the primary purpose in 79.0\% of samples ($\kappa=0.627$) and on severity in 75.5\%. Among the 168 samples receiving numeric severity ratings from both reviewers, $\kappa_w=0.627$. Abstention rates of 12.3\% for purpose and 11.0\% for severity (\cref{fig:human-audit-summary}c) preserve cases that cannot be resolved from visual evidence alone.

Taken together, the audit shows that the generated annotations preserve the intended supervisory signal for most samples, while residual noise is concentrated in spatial precision and fine-grained evidential claims. Intent judgments remain reproducible despite their intrinsic ambiguity. We therefore treat these annotations as structured, noisy supervision rather than exhaustive ground truth, and isolate their downstream contribution through the ablations in Section~7. The sampling composition, rating criteria, and review protocol are provided in the supplementary material.

\section{Models and training}
\label{sec:models}

\subsection{Model}

\label{sec:models-model} Our model is Qwen2.5-VL-3B-Instruct~\cite{bai2025qwen25vl} fine-tuned with LoRA~\cite{hu2022lora}. Every variant trains on the same 27,846 images of the training split, 13,923 edits balanced over the 13 edit types and 13,923 authentic photos (\cref{sec:supp-training}). All reported runs use seed 0 (\cref{sec:supp-seeds}).

\subsection{Variants}
\label{sec:models-variants}

The variants differ in the text they are trained to write. \emph{VB} (verdict and boxes) writes the verdict and, for an edited image, the boxes of \cref{sec:task-outputs}, with no reasoning. \emph{SA} (structured answer) writes the structured answer of \cref{sec:task-outputs} without the reasoning: the verdict, the boxes, the edit type with its category, and the intent. \emph{RG} (grounded reasoning) writes the full answer: the reasoning, whose claims point to boxes, and then SA's structured answer. \emph{RN} (reasoning, no inline boxes) writes RG's reasoning with the box references removed from its claims, and then SA's structured answer. For an authentic image, VB and SA write only the verdict; RN and RG write their reasoning followed by the verdict. \emph{RU} (unverified reasoning) writes RN's answer but is trained on the claims as the generator wrote them, before the blind verification of \cref{sec:automatic}. \emph{AF} (answer first) writes RG's two parts in the opposite order; \emph{AF+W} is AF with a reweighted loss. \emph{RG+RL} (reinforcement learning) continues RG's adapter with group relative policy optimization (GRPO)~\cite{shao2024deepseekmath} on rewards that need no judge (\cref{sec:models-training}). The image score of each variant is the probability of ``edited'' at its verdict token, computed as for the zero-shot MLLMs (\cref{sec:supp-baselines}).

\section{Experiments}
\label{sec:experiments}

\subsection{Setup}
\label{sec:exp-setup}

\paragraph{Test sets.}

We build two test sets from the L and R edits of the two test splits (\cref{sec:supp-testsets}): \emph{Seen}, 1,950 edits of the 13 seen types and 1,950 photos, and \emph{Unseen}, 450 edits of the three held-out local types and the same photos; every method is scored on the same images.

\paragraph{Baselines and metrics.}
We evaluate three groups with their released weights and no training on our data. We measure detection with the image-level AUC, computed from each method's score, and with the accuracy and the true- and false-positive rates (TPR, FPR), computed from its verdict with ``edited'' as the positive class, and localization on the edited images only.

\subsection{In-domain detection and localization}
\label{sec:exp-indomain}

\begin{table*}[t]
\centering\footnotesize\setlength{\tabcolsep}{3.7pt}
\begin{tabular}{l ccccccc cc}
\toprule
 & \multicolumn{7}{c}{Seen: 13 edit types (1,950 edits + 1,950 photos)} & \multicolumn{2}{c}{Unseen (450 + 1,950)} \\
\cmidrule(lr){2-8}\cmidrule(lr){9-10}
Method & AUC [95\% CI] & Acc & TPR & FPR & pIoU & bF1 & Parse & AUC [95\% CI] & pIoU \\
\midrule
\multicolumn{10}{l}{\textit{IFDL methods}} \\
TruFor~\cite{Guillaro_2023_CVPR} & 47.3 {\scriptsize[45.9, 48.9]} & 48.6 & 14.9 & 17.6 & 4.6 & 4.2 & -- & 47.3 {\scriptsize[44.5, 50.2]} & 5.2 \\
MVSS-Net~\cite{Chen2021MVSS} & 43.4 {\scriptsize[42.0, 44.8]} & 46.1 & 64.2 & 71.9 & 1.1 & 0.7 & -- & 41.2 {\scriptsize[38.5, 44.2]} & 1.3 \\
PIM~\cite{kong2025pim} & 57.4 {\scriptsize[56.0, 58.7]} & 52.3 & 93.7 & 89.1 & 10.4 & 8.1 & -- & 57.0 {\scriptsize[54.1, 59.8]} & 9.1 \\
HiFi-Net~\cite{Guo_2023_CVPR} & 51.7 {\scriptsize[50.7, 52.8]} & 50.3 & 3.1 & 2.5 & 0.5 & 1.0 & -- & 52.9 {\scriptsize[50.2, 55.6]} & 0.6 \\
\midrule
\multicolumn{10}{l}{\textit{Forensic MLLMs}} \\
SIDA-7B~\cite{Huang_2025_CVPR} & 58.0 {\scriptsize[57.0, 59.0]} & 51.0 & 3.4 & 1.4 & 0.4 & 0.7 & 100.0 & 56.8 {\scriptsize[53.8, 59.8]} & 0.2 \\
FakeShield$^\dagger$~\cite{xu2025fakeshield} & 48.8 {\scriptsize[47.7, 49.8]} & 48.8 & 16.3 & 18.8 & 0.9 & 1.3 & 100.0 & 48.8 {\scriptsize[47.0, 50.9]} & 0.9 \\
So-Fake-R1$^\dagger$~\cite{huang2026sofakebenchmarkingexplainingsocial} & 50.5 {\scriptsize[49.3, 51.6]} & 50.5 & 26.1 & 25.1 & 2.7 & 3.8 & 99.7 & 47.9 {\scriptsize[45.8, 50.2]} & 1.9 \\
\midrule
\multicolumn{10}{l}{\textit{Zero-shot MLLMs}} \\
Qwen2.5-VL-3B$^\ddagger$~\cite{bai2025qwen25vl} & 6.2 {\scriptsize[5.6, 6.9]} & 17.5 & 34.9 & 100.0 & 4.7 & 4.4 & 35.3 & 6.8 {\scriptsize[5.7, 7.9]} & 5.2 \\
Qwen2.5-VL-7B$^\ddagger$~\cite{bai2025qwen25vl} & 49.1 {\scriptsize[48.2, 50.0]} & 49.2 & 89.7 & 91.4 & 9.5 & 7.4 & 97.4 & 46.3 {\scriptsize[43.3, 49.2]} & 9.5 \\
Qwen2.5-VL-32B (4-bit)~\cite{bai2025qwen25vl} & 51.0 {\scriptsize[50.2, 51.9]} & 51.0 & 12.9 & 10.9 & 1.2 & 1.7 & 100.0 & 50.5 {\scriptsize[47.5, 53.4]} & 0.9 \\
Qwen3-VL-8B~\cite{bai2025qwen3vltechnicalreport} & 55.6 {\scriptsize[54.7, 56.4]} & 53.9 & 52.1 & 44.2 & 9.0 & 8.9 & 99.9 & 54.3 {\scriptsize[51.5, 57.0]} & 9.0 \\
InternVL3-8B~\cite{zhu2025internvl3} & 53.3 {\scriptsize[52.5, 54.1]} & 51.2 & 21.6 & 19.3 & 2.2 & 2.3 & 99.4 & 53.4 {\scriptsize[50.7, 56.3]} & 1.8 \\
\midrule
\multicolumn{10}{l}{\textit{IFDL methods fine-tuned on our data}} \\
TruFor~\cite{Guillaro_2023_CVPR} & 99.1 {\scriptsize[98.9, 99.3]} & 95.1 & 93.8 & 3.6 & 30.8 & 31.5 & -- & 99.1 {\scriptsize[98.8, 99.4]} & 27.3 \\
MVSS-Net~\cite{Chen2021MVSS} & 100.0 {\scriptsize[100.0, 100.0]} & 100.0 & 100.0 & 0.0 & 35.2 & 33.2 & -- & 100.0 {\scriptsize[100.0, 100.0]} & 28.7 \\
PIM~\cite{kong2025pim} & 98.7 {\scriptsize[98.4, 99.0]} & 95.9 & 96.6 & 4.7 & 29.0 & 25.4 & -- & 98.8 {\scriptsize[98.4, 99.1]} & 22.0 \\
\midrule
\multicolumn{10}{l}{\textit{Ours: Qwen2.5-VL-3B~\cite{bai2025qwen25vl} with LoRA~\cite{hu2022lora} (\cref{sec:models})}} \\
VB: verdict and boxes & 93.1 {\scriptsize[92.4, 93.8]} & 78.1 & 60.1 & 3.9 & 16.4 & 26.0 & 100.0 & 91.3 {\scriptsize[90.0, 92.6]} & 11.0 \\
SA: VB + type and intent & 92.7 {\scriptsize[91.9, 93.4]} & 73.1 & 48.4 & 2.2 & 13.9 & 21.7 & 100.0 & 90.1 {\scriptsize[88.8, 91.4]} & 7.3 \\
RU: RN on unverified claims & 55.2 {\scriptsize[53.8, 56.6]} & 54.5 & 25.6 & 16.5 & 5.9 & 10.4 & 100.0 & 56.9 {\scriptsize[54.0, 59.9]} & 4.7 \\
RN: RG, no inline boxes & 57.7 {\scriptsize[56.4, 59.1]} & 55.4 & 27.7 & 16.9 & 6.4 & 11.0 & 99.9 & 55.9 {\scriptsize[52.9, 58.9]} & 4.2 \\
RG: grounded reasoning & 53.1 {\scriptsize[51.7, 54.5]} & 53.6 & 20.3 & 13.1 & 4.4 & 9.6 & 100.0 & 54.9 {\scriptsize[52.0, 58.0]} & 4.3 \\
RG+RL: RG + GRPO & 57.1 {\scriptsize[55.7, 58.5]} & 56.3 & 47.2 & 34.7 & 9.5 & 18.6 & 100.0 & 56.1 {\scriptsize[53.1, 59.0]} & 9.6 \\
AF: RG, answer first & 65.3 {\scriptsize[64.3, 66.2]} & 52.0 & 6.6 & 2.6 & 2.0 & 4.2 & 100.0 & 65.1 {\scriptsize[62.5, 67.7]} & 1.2 \\
AF+W: AF, weighted loss & 87.2 {\scriptsize[86.2, 88.1]} & 63.2 & 28.0 & 1.7 & 7.5 & 14.2 & 100.0 & 86.3 {\scriptsize[84.6, 87.9]} & 4.4 \\
\bottomrule
\end{tabular}
\caption{\textbf{Detectors on our test sets.} The first three groups use their released weights with no training on our data; the last two are trained on our data. \emph{Unseen}: 3 held-out edit types, same editor. Percentages; AUC with its 95\% bootstrap interval over source photos. FPR is on the authentic photos; pIoU and bF1 (box F1 at IoU 0.5) are on the edited images; Parse is the share of answers that parse. $^\dagger$No probability output: AUC ranks by the verdict. $^\ddagger$A parse failure counts as a wrong verdict with the worst-case score (\cref{sec:supp-indomain}). }
\label{tab:e1-baselines}
\end{table*}

No existing detector reaches 60 image AUC on the \emph{Seen} set (\cref{tab:e1-baselines}). The best are SIDA (58.0), PIM (57.4) and Qwen3-VL-8B (55.6), and pixel IoU stays at or below 10.4\%.

We also fine-tune TruFor, MVSS-Net and PIM on 10,000 edits of the training split, balanced over the 13 edit types with one edit per source photo, each paired with its own source photo as the authentic image. Each starts from its released weights; \cref{sec:supp-ifdl-ft} gives the training recipes.

Trained on our data, VB reaches 93.1 AUC on the \emph{Seen} set, 35 points above the best existing detector, with TPR 60.1 at FPR 3.9, pixel IoU 16.4 and box F1 26.0 (\cref{tab:e1-baselines}, \cref{sec:supp-per-type}).   The 253 edits marked as leaving no visible evidence (\cref{sec:supp-pipeline}) are detected almost as often as the others (TPR 58.5 against 60.3), and for 65\% of the edits it calls edited VB's box is the whole image (\cref{sec:supp-per-type}): two signs that VB responds to something other than the visible edit. The fine-tuned IFDL methods score higher still (TruFor 99.1, MVSS-Net 100.0 and PIM 98.7 AUC; pixel IoU 30.8, 35.2 and 29.0), but they, and in part VB, rely on a trace of the editor (\cref{sec:exp-generalization}).

RG stays near chance: 53.1 AUC, TPR 20.3 at FPR 13.1, box F1 9.6. It calls 80\% of the edits authentic. Its score is read after the reasoning has fixed the verdict, so its AUC is the verdicts' balanced accuracy (\cref{sec:supp-e-score}). The verdict is decided much earlier, by whether the first claim of the reasoning cites a box, as in the training targets, where every edited answer cites boxes and no authentic one does: a first claim that cites a box ends in ``edited'' 96.5\% of the time, and RG's first claim cites a box for only 19.8\% of the edits. The full answer, as we train it, thus costs about 40 AUC points against VB. VB and RG also differ in the edit type and intent that RG writes, but SA, which adds these fields to VB without reasoning, reaches 92.7 AUC, within 0.4 of VB, and RN, which keeps RG's reasoning but removes the box references from its claims, reaches 57.7 (\cref{tab:e1-baselines}). The cost therefore comes from the reasoning, not from the extra fields or from missing evidence: VB separates the images RG calls authentic at 92.7 AUC. The inline box references account for about 5 of the 40 points and the order for about 12 (AF, 65.3), and a reweighted loss recovers most of the rest: AF with half of each example's loss on the answer (AF+W) reaches 87.2 with all four outputs, 5.5 below SA (\cref{sec:supp-indomain}). RU, trained on the unverified claims, reaches 55.2, close to RN, so verification changes little for detection. RG+RL raises RG's TPR and box F1 but its AUC only to 57.1: the tuning moved the threshold, not the ranking.

\subsection{Generalization to held-out edit types and to other editors}
\label{sec:exp-generalization}

\begin{table}[t]
\centering\footnotesize\setlength{\tabcolsep}{3pt}
\begin{tabular}{l ccc cccc}
\toprule
 & \multicolumn{3}{c}{Editor} & \multicolumn{4}{c}{Trace test, against S} \\
\cmidrule(lr){2-4}\cmidrule(lr){5-8}
Method & NB & Qwen & DALL-E 2 & T & U & P & K \\
\midrule
\multicolumn{8}{l}{\textit{Trained on our data}} \\
TruFor & 94.1 & 41.2 & 47.4 & 96.4 & 82.4 & 71.9 & 49.0 \\
MVSS-Net & 100.0 & 69.0 & 55.5 & 100.0 & 98.5 & 92.1 & 50.7 \\
PIM & 93.2 & 70.3 & 54.6 & -- & -- & -- & -- \\
VB (ours) & 78.8 & 66.9 & 48.0 & 88.0 & 69.9 & 63.8 & 53.2 \\
SA (ours) & 79.5 & 69.3 & 48.0 & -- & -- & -- & -- \\
\midrule
\multicolumn{8}{l}{\textit{Released weights}} \\
TruFor & 46.3 & 48.0 & 60.3 & 48.6 & 52.1 & 50.4 & 49.7 \\
Best method & 57.6 & 61.1 & 66.4 & -- & -- & -- & -- \\
\bottomrule
\end{tabular}
\caption{\textbf{Other editors, and what the detectors detect.} Image AUC. Left: source photos against their Nano-Banana (NB) and Qwen-Image-Edit edits, and MagicBrush's DALL-E 2 edits. Right: versions of the edits against the source photo S (\cref{sec:exp-generalization}). Best method: the highest released one per column; --: not run.}
\label{tab:transfer}
\end{table}

\paragraph{Held-out edit types.}

The models trained on our data keep their detection on the \emph{Unseen} set, each within 2.6 AUC of its \emph{Seen} score (\cref{tab:e1-baselines}). Localization drops more: the pixel IoU of the fine-tuned IFDL detectors, VB and SA falls by 3.5 to 7.0 points. The held-out types share the training data's editor, so the editor's trace alone could explain this (\cref{sec:supp-gsx}); it is not transfer to new editors.

\paragraph{Another editor.} On MagicBrush's 528 DALL-E 2 edits and 263 photos, training on our data does not carry over (\cref{tab:transfer}, \cref{sec:supp-generalization}): fine-tuned TruFor falls below its released weights, VB is at chance, and at the fixed threshold both call almost every edit authentic. The highest AUCs come from released zero-shot MLLMs (up to 66.4), all higher than on our sets (50--56); this suggests that DALL-E 2's edits show more visible artifacts than Nano-Banana's, not that these models detect edits in general.

\paragraph{What the fine-tuned detectors detect.}

All our edits come from one editor, Nano-Banana~\cite{qian2026pico}, which regenerates the whole image, so a detector can learn a trace of that editor instead of the edit. We take 630 edits from the two test sets (40 per \emph{Seen} type except scene background, 50 per \emph{Unseen} type) and make five versions of each, processed identically (one resampling onto a common grid, the same JPEG compression). \emph{S} is the source photo and \emph{T} the edit. With $C$ the reference mask plus every pixel that changed visibly, \emph{P} is the source with only $C$ taken from the edit, \ie, the edited patch on an otherwise untouched photo; \emph{U} is the edit with $C$ put back from the source, \ie, the editor's output with the edit undone; \emph{K} is the source with $C$ taken from a copy of the source that went through the edit's resampling and compression, a control for the seam. We score each version against S (\cref{tab:transfer}; \cref{sec:supp-trace}). With identical processing, fine-tuned TruFor still separates T from S at 96.4 AUC (99.1 on the test sets) and VB at 88.0 (93.1; 78.8 on the second editor's Nano-Banana edits), so processing differences explain little of TruFor's score but may explain some of VB's. Both fine-tuned detectors find U, which holds no edit, more easily than P, which holds all of it, and far more easily than the seam control K. Our VB behaves the same way, if less strongly. The fine-tuned detectors, and to a lesser degree VB, thus respond mainly to a trace the editor leaves over the whole image, not to the edited content, which is consistent with their failure on MagicBrush. Training VB against this trace does not remove it (\cref{sec:exp-ablations}).

\paragraph{Second editor.}

We score every method on 390 source photos against their Nano-Banana edits and against their Qwen-Image-Edit~\cite{wu2025qwenimage} edits of the same instructions (\cref{tab:transfer}, \cref{sec:supp-second-editor}). The fine-tuned IFDL detectors keep 93--100 AUC on the Nano-Banana edits and fall to 41--70 on the Qwen edits, where at their threshold they flag 2--43\% of the edits, so the trace they learned is Nano-Banana's. VB and SA lose less and stay ahead of every released method (at most 61.1, SIDA-7B); RU, RN, RG and RG+RL stay near chance on both editors (53--59). On 4,893 edits by 18 further MLLM editors from MLLM-Edit's public release~\cite{yu2026mllmedit} (\cref{sec:supp-mllm-edit}), the fine-tuned detectors, VB and SA reach 44.8--59.6 AUC over all editors and at most 72.5 on any one (\cref{tab:supp-mllm-edit}), except fine-tuned MVSS-Net on Nano Banana Pro~\cite{google2025nanobananapro}, a later editor of Nano-Banana's family (84.6).

\subsection{Intent prediction}
\label{sec:exp-intent}

SA answers an intent for each image it calls edited (943 of the 1,950 \emph{Seen} edits). Against the pipeline's labels (\cref{sec:automatic}), its primary intent agrees on 64.2\% of those edits, where always answering the majority class, benign aesthetic, would give 58.6\%, and its severity correlates with the labeled severity at Spearman 0.50. The agreement comes from the frequent classes: recall is 81\% for benign aesthetic, 41--53\% for misinformation, identity, document and commercial edits, and 5\% for humor, and the binary question, harmful or not (\cref{sec:supp-pipeline}), is answered correctly for 76.8\% of the answered edits against 69.9\% for always ``benign''. On the 168 \emph{Unseen} edits it answers, the agreement falls to 51.2\% against a 79.2\% majority, and harmful edits are recognized at chance (35.7\% recall, 38.6\% false alarms), so intent transfers to new edit types even less than detection does.

\subsection{Grounding quality and errors}
\label{sec:exp-grounding}

Does a verdict rest on the region the model cites? For every \emph{Seen} image a model calls edited whose cited region covers at most half of it, we remove one region by inpainting, re-encode the image with its own JPEG tables and ask the model again (\cref{tab:faithfulness}, \cref{sec:supp-masking}). Re-encoding alone flips 16.7\% of RG's verdicts on edits. Against that baseline, removing the region RG cites flips its verdict on 76.6\% of the edits, a random region of the same shape on 36.9\% (paired difference $+39.7$ [31.9, 47.2]; RG+RL $+19.3$, VB $+35.5$), and on its false alarms the two figures are 67.6\% and 36.6\%. The verdict thus follows the region the model points at, whether or not an edit is there: when the evidence is hallucinated, removing it removes the verdict with it. Given another image's reasoning, RG's verdict follows it on every image (\cref{sec:supp-swap}).

\subsection{Ablations}

\label{sec:exp-ablations}

We retrain VB with one change at a time (\cref{sec:supp-ablations}, \cref{tab:ablations}). VO, trained to write the verdict alone, reaches 93.2 AUC on the \emph{Seen} set against VB's 93.1 but 62.3 against 66.9 on the Qwen edits. Half of the edits cost 5.6 AUC points on the \emph{Seen} set and a quarter 21.5, and all 69,754 L and R edits for training add 3.5 (96.6; 72.1 on the Qwen edits, 43.9 on MagicBrush), so detection at 13,923 edits is limited by the data, not by the supervision. VB+U adds 3,000 pairs in which \emph{U} is labeled authentic; it raises AUC on the Qwen edits from 66.9 to 70.1 (one seed), but the trace stays: VB+U separates \emph{U} from \emph{S} at 69.1 AUC against VB's 69.9.

\section{Limitations and ethics}
\label{sec:limitations}

\paragraph{Limitations.} All our edits come from one editor, so much of what the fine-tuned detectors learn is that editor's trace (\cref{sec:exp-generalization}). Nano-Banana's outputs also carry an invisible SynthID watermark~\cite{google2025gemini25flashimage,gowal2025synthidimage}; whether the trace is this watermark, the regeneration or both is not settled (\cref{sec:supp-trace-wm}). The pipeline needs only an editing dataset's triplets, so other editors can be added. \Cref{sec:supp-limitations} discusses the other limitations.

\paragraph{Release and ethics.} Pico-Banana is licensed CC BY-NC-ND 4.0, so we release our code and pipeline, not the images; to verify our numbers, the data must be rebuilt from Pico-Banana with our pipeline. The source photos come from Open Images and show real people; edits of age, gender or expression change how they appear, which is one more reason not to redistribute the images. Intent labels describe what an edit could be used for, not the people shown. Explanations of what looks edited could also guide a forger in hiding an edit.

\section{Conclusion}
\label{sec:conclusion}

We built an automatic pipeline for grounded, verified edit annotations and instruction-free intent labels, a dataset of 253K edits in five tiers with held-out types and a human audit, and a benchmark of forensic detectors, forensic MLLMs and zero-shot MLLMs. Existing detectors do not detect these edits. A 3B model trained on our data does, on seen and held-out types, but partly through the editor's whole-image trace: fine-tuned detectors learn it almost entirely, training against it does not remove it, and none of them carries to an unrelated editor. Grounded reasoning written before the verdict costs detection, most of which an answer-first model with a reweighted loss recovers, and a reasoning model's verdict rests on the regions it cites and on its reasoning. Intent is predicted above the majority class for the frequent purposes on seen types only. Next steps follow: edits from several editors in training and test, the global, stylized and geometric tiers, and learning reasoning and detection together.

\section*{LLM Usage Statement}
Large language models assisted this work at three stages. Claude Fable 5 and the Claude Opus 5 series (versions 5 through 5.5) helped us refine our own initial research ideas and assisted with the design and implementation of the experiments. GPT-6-Astra, GPT-5.6-Sol, and Qwen-32B were used for the initial automated annotation that preceded the human evaluation. For the writing, Claude Opus 5.5 and GPT-6-Astra were used to polish the language and improve the fluency of the text. We carefully reviewed and revised every output of these models, and we take full responsibility for the final content of this paper, including all text, claims, and artifacts produced with their help.

{
    \small
    \bibliographystyle{ieeenat_fullname}
    \bibliography{main}
}

\clearpage
\setcounter{page}{1}
\maketitlesupplementary
\appendix

\section{Extended related work}
\label{sec:supp-related}

\Cref{tab:supp-datasets} compares our data with datasets for forgery detection and localization, for explainable forgery detection and for instruction-based editing. Among them, only ours ties its explanations to boxes and checks each claim with a blind verifier, and only ours labels intent and severity; So-Fake-Set has free-text rationales for 6K evaluation images, and the public release of MLLM-Edit covers 18 instruction-based editors but has no explanations.

\begin{table*}[t]
\centering\footnotesize\setlength{\tabcolsep}{3pt}
\resizebox{\ifdim\width>\textwidth\textwidth\else\width\fi}{!}{%
\begin{tabular}{l rr l l l l l}
\toprule
Dataset & Edited/fake & Authentic & Made by & Location & Explanation & Intent & Held out \\
\midrule
CASIA v2.0~\cite{dong2013casia} & 5.1K & 7.5K & by hand & -- & -- & -- & -- \\
IMD2020~\cite{novozamsky2020imd2020} & 37K & 35K & classic, GAN & masks & -- & -- & -- \\
DEFACTO~\cite{mahfoudi2019defacto} & 229K & n/r & classic, automatic & masks & -- & -- & -- \\
AutoSplice~\cite{jia2023autosplice} & 3.6K & 2.3K & DALL-E 2 inpainting & masks & -- & -- & -- \\
MagicBrush~\cite{zhang2023magicbrush} & 10.4K & 5.3K & DALL-E 2, instruction & masks (by hand) & -- & -- & -- \\
MMTD-Set~\cite{xu2025fakeshield} & 52K & 54K & classic, GAN, SD inpainting & masks & free text & -- & -- \\
SID-Set~\cite{Huang_2025_CVPR} & 200K & 100K & LDM inpainting, FLUX & masks & free text (3K) & -- & -- \\
SynthScars~\cite{kang2025legion} & 12.2K & 0 & fully generated & masks (by hand) & free text & -- & -- \\
So-Fake-Set~\cite{huang2026sofakebenchmarkingexplainingsocial} & 1.3M & 650K & 30 generators & masks, boxes & free text (6K) & -- & 15 generators \\
FakeClue~\cite{wen2025spot} & \multicolumn{2}{c}{over 100K in all} & mostly generated & -- & free text & -- & -- \\
MLLM-Edit~\cite{yu2026mllmedit} & 242K & n/r & 18 MLLM editors & masks & -- & -- & -- \\
\midrule
Ours & 253K & 171K & Nano-Banana & masks, boxes & verified claims, boxes & 7 classes, severity & 4 types, 20 editors \\
\bottomrule
\end{tabular}}
\caption{\textbf{Comparison with prior datasets.} Edited/fake: manipulated, edited or generated images; authentic: real photos, as each dataset's paper counts them, or for MLLM-Edit its public release page (MMTD-Set: its training part; SID-Set and So-Fake-Set: tampered and fully generated images together, So-Fake-Set's counts approximate; FakeClue reports only its total of real and fake images). Location: pixel masks or boxes. Explanation: text that says why an image is judged edited, with the number of images that have it when only some do. Held out: test edit types or generators absent from training. n/r: not reported.}
\label{tab:supp-datasets}
\end{table*}

\subsection{Image manipulation detection and localization}

Classic benchmarks pair spliced, copied or inpainted photos with binary masks of the tampered regions~\cite{dong2013casia,hsu2006detecting,wen2016coverage}. Image forgery detection and localization (IFDL) methods trained on them look for low-level traces that a local manipulation leaves behind: MVSS-Net supervises noise and edge views at several scales~\cite{Chen2021MVSS}, TruFor fuses RGB with a learned noise fingerprint and outputs an anomaly map, a confidence map and an integrity score~\cite{Guillaro_2023_CVPR}, HiFi-Net classifies a forgery at several levels, from real or forged down to the method that made it, and localizes it with a metric-learning loss that gathers authentic pixels around a centre~\cite{Guo_2023_CVPR}, and PIM models the pixel correlations that camera demosaicing leaves and splicing breaks~\cite{kong2025pim}. Detectors of fully generated images instead look for traces of the generator, such as diffusion reconstruction error~\cite{wang2023dire}. Such detectors also pick up biases in their benchmarks: Grommelt~\etal show that JPEG compression and image size separate classes in common generated-image datasets and that detectors learn them~\cite{grommelt2024fakeorjpeg}, which is why we process our two classes the same way (\cref{sec:data-preprocessing}). An instruction-based editor sits between the two cases: it starts from a real photo but regenerates the whole image. On our test sets, these detectors are close to chance with their released weights (\cref{sec:exp-indomain}), and once fine-tuned on our data they respond mainly to a trace that the editor leaves over the whole image, not to the edited region (\cref{sec:exp-generalization}).

Later datasets widen the kinds of manipulation: IMD2020 collects real-life manipulated images from the Internet with their masks, next to a large set of camera images~\cite{novozamsky2020imd2020}, DEFACTO generates splicing, copy-move, object-removal and face forgeries automatically from public photo collections~\cite{mahfoudi2019defacto}, and AutoSplice edits photos with DALL-E~2 from text prompts and keeps the edited regions as masks~\cite{jia2023autosplice}. For fully generated images, a classifier trained on one GAN with suitable augmentation already transfers to other CNN generators~\cite{wang2020cnnspot}, a linear probe on the features of a frozen vision--language model transfers further~\cite{ojha2023univfd}, and GenImage benchmarks such detectors on about a million images from several generators~\cite{zhu2023genimage}.

\subsection{Instruction-based editors and editing data}

Since MagicBrush and the datasets of \cref{sec:related-editing}, editing data have grown in size and in the number of edit types: OmniEdit trains one editor on pairs made by several specialist models~\cite{wei2024omniedit}, AnyEdit unifies more than twenty edit types in one dataset~\cite{yu2025anyedit}, and ImgEdit pairs a large dataset of single- and multi-turn edits with a benchmark~\cite{ye2025imgedit}. Open editors such as FLUX.1 Kontext~\cite{labs2025kontext}, Step1X-Edit~\cite{liu2025step1xedit} and Qwen-Image-Edit~\cite{wu2025qwenimage}, our second editor (\cref{sec:supp-second-editor}), are candidates for the training and test sets from several editors that our results call for (\cref{sec:conclusion}).

\subsection{Faithful and grounded explanations}

An explanation is faithful when it reflects the process that produced the prediction, a different property from being plausible to a reader~\cite{jacovi2020faithfulness}. Chain-of-thought explanations can be unfaithful: a model's answer can follow a feature that its explanation never mentions~\cite{turpin2023unfaithful}, and interventions on the reasoning, such as truncating it or inserting a mistake, measure how much the answer depends on it~\cite{lanham2023faithfulness}. Our region-masking and rationale-swap tests (\cref{sec:supp-masking,sec:supp-swap}) are interventions of this kind, on the region a model cites and on the reasoning it reads before the verdict. Grounded MLLMs tie text to image regions, Kosmos-2 with discrete location tokens~\cite{peng2023kosmos2} and Shikra with coordinates written as numbers in the text~\cite{chen2023shikra}; our reasoning writes each region's box as numbers on a 0--1000 grid, as Shikra does.

\subsection{Harm and intent in synthetic media}

\label{sec:related-intent}

Misinformation research asks how images mislead in context. NewsCLIPpings pairs real images with real but mismatched captions~\cite{luo2021newsclippings}, DGM4 detects and grounds manipulations in image--text pairs~\cite{shao2023dgm4}, and MMFakeBench tests MLLMs on misinformation that mixes textual, visual and cross-modal distortions~\cite{liu2025mmfakebench}. Marchal \etal derive a taxonomy of generative-AI misuse tactics and goals from about 200 reported incidents~\cite{marchal2024misuse}. These works judge a post or a claim, usually together with its text. We instead predict the plausible purpose of an edit, and its potential harm if the image were believed real, from the edited image alone (\cref{sec:task-formulation}).

\section{Intent and severity}
\label{sec:supp-intent}

Severity is the potential harm if the edited image were believed real: 0 when nobody is misled about anything that matters (\eg a dessert swapped for another), 1 for a mild misrepresentation someone might rely on, 2 when it could mislead about people, products, places or events (\eg a warning sign removed), and 3 when it could cause real harm (\eg fabricated evidence of an accident or an altered document).

\subsection{Intent is not instruction following}

\label{sec:task-intent}

Intent prediction asks what an edit is plausibly for and how much harm it could do. Instruction following asks whether the editor did what it was asked, which is a question about editing quality. The two are independent: removing a warning sign exactly as requested is faithful editing and a moderate-severity edit. Our intent labels are produced without the instruction, because polite requests bias the labels toward benign (\cref{sec:automatic}). Both the labels and the model therefore judge an edit from the image alone.

\section{Data pipeline}
\label{sec:supp-pipeline}

The pipeline (\cref{fig:pipeline}) turns an editing dataset's triplets, a source photo, an instruction and the edited image, into supervision for the four outputs of \cref{sec:task-outputs} without a human in the loop; \cref{sec:annotation} audits its output. It runs as one script per step, each step records its configuration, and every random draw is seeded by the image, so a rerun is identical.

\subsection{Source and tiers}

Our edits come from Pico-Banana-400K~\cite{qian2026pico}, which edits Open Images photos~\cite{Kuznetsova2020} with Nano-Banana from instructions of 35 edit types; we use its single-turn edits, 257,726 after de-duplication. We map the 35 types to five tiers by what a detector has to localize: \emph{local} (L) edits of objects, text or people (15 types, 94,184 edits), \emph{regional} (R) edits of the scene background (14,829), \emph{global} (G) changes of tone, grain, season, weather or lighting (82,905), \emph{stylized} (S) edits (39,677) and \emph{geometric} (X) edits such as outpainting and zooming (26,131). The experiments use the L and R tiers, whose edits have a region to localize (\cref{sec:task-outputs}); the pipeline gives the other tiers a full-image mask, and they enter a detection test only (\cref{sec:supp-gsx}). Splits are grouped by source photo, so no scene appears in two splits, with 85\%, 5\% and 10\% of the edits in train, validation and test. Four edit types are held out: every photo with an edit of a held-out type goes to the test pool, its held-out edits form a separate test split and its other edits stay in test. Three of the four types (expression, relocate object and text font) are local, and they form the \emph{Unseen} set of \cref{sec:exp-setup}. Edits whose files, registration or mask failed are excluded (4,514 edits: 2.5\% of L and R, 1.2\% of G, S and X), which leaves 253,212 edits, 106,274 of them in L and R (\cref{tab:dataset}). Authentic images are the edits' own source photos, each in the same split as its edits, plus extra untouched Open Images photos chosen to match the source photos' distribution of image-level labels and sizes; each split pairs its edits with as many photos, and the held-out split reuses the test split's photos.

\begin{table}[t]
\centering\footnotesize\setlength{\tabcolsep}{2pt}
\begin{tabular}{l rrrr r}
\toprule
 & \multicolumn{4}{c}{Edits} & \\
\cmidrule(lr){2-5}
Split & all tiers & L and R & with reasoning & for training & Photos \\
\midrule
train & 197,170 & 82,206 & 187,712 & 139,779 & 139,779 \\
val & 11,588 & 4,789 & 11,037 & 8,237 & 8,237 \\
test & 23,126 & 9,943 & 22,000 & -- & 23,126 \\
held-out types & 21,328 & 9,336 & 20,932 & -- & test's \\
\midrule
total & 253,212 & 106,274 & 241,681 & 148,016 & 171,142 \\
\bottomrule
\end{tabular}
\caption{The five tiers per split, after the QA exclusions of \cref{sec:data-source}: all edits, those of the L and R tiers that the experiments use, edits with verified reasoning, and the edits for fine-tuning (train and val drop the edits whose reasoning found no visible evidence; the test splits keep every edit). Photos are the authentic images paired with each split (source photos plus extra Open Images photos); the held-out split (test, held-out types) reuses the test split's photos.}
\label{tab:dataset}
\end{table}

\subsection{Masks and boxes}

A mask marks where the edit changed the photo. The source is first brought into the edit's frame, by a plain resize when the aspect ratios agree and otherwise by the best of a stretch, a center crop and a similarity transform from SIFT matches, judged by the normalized cross-correlation of the gradients, so that the mask aligns with the image the detector will see. The mask is the thresholded difference of the aligned pair in three views, LAB color, LPIPS~\cite{zhang2018lpips} and SSIM~\cite{wang2004ssim}, from a generator built for this purpose; since the L and R edits are local by definition, the generator's routing of a large overall difference to a whole-image mask is off, and such a difference becomes a quality flag instead. Boxes come from the mask's connected components after a light opening and closing: components under 0.1\% of the image are dropped, components closer than 2\% of the diagonal are merged, and at most eight are kept, largest first, each as a box on the $0$--$1000$ grid. \Cref{sec:audit-mask} audits the masks, and \cref{sec:supp-boxes} says how the reference boxes of the test sets derive from them.

\subsection{Matched preprocessing}

Nano-Banana writes PNG files in a few fixed sizes, and Open Images photos are camera JPEGs in many sizes, so size, format and compression history would separate the classes before any edit is looked at~\cite{grommelt2024fakeorjpeg}. Each authentic photo therefore draws its output size from the edits' size distribution and is center-cropped to it (a draw that would discard more than half of the photo is redrawn); every image, edited or not, is resampled by a factor drawn from $U(0.75, 0.95)$, so neither class keeps its native pixel grid; the edits first receive a simulated capture JPEG at a quality drawn from $[85, 98]$; and every image then receives the same final JPEG chain, one pass at a quality from $[75, 95]$ and, with probability 0.3, a second. Edits are rescaled but never cropped, so a box on the $0$--$1000$ grid is the same in the raw edit, where the teachers annotate, and in the training image. Before anything is trained, probes on the file format, on bytes per pixel and on width, height, aspect ratio and the stored quantization table must be near chance; after this processing they reach 0.53 and 0.54 AUC (the step requires less than 0.55). A second probe, on the noise residuals of patches away from the edited region, is not expected to be at chance for an editor that regenerates the whole image; \cref{sec:exp-generalization} measures what this leaves for a detector to learn. \Cref{sec:supp-limitations} states where the processing of the two classes still differs on the test sets.

\subsection{Grounded claims}

The reasoning is a list of claims about visible evidence, each tied to a region, written by a generator with privileged inputs and kept only if a blind verifier agrees. The generator, Qwen3-VL-32B~\cite{bai2025qwen3vltechnicalreport}, sees the aligned source, the edit, the edit with its mask tinted and its regions numbered, the instruction and the edit type. It writes private notes first, a sanctioned place for what changed, and then claims that must describe the edited image as it looks now, each pointing to a region and optionally to a tighter box, and each marked as evidence of editing or as a note that a part looks consistent. The verifier, InternVL3.5-8B~\cite{wang2025internvl35}, from a different model family, sees only the edited image with its regions drawn, never the source or the instruction, so it can tell whether a claim is checkable from the edit alone; it returns, per claim, whether the claim is visible and grounded, and a claim it cannot see is dropped. Leaks of the privileged inputs are caught by a regular-expression screen (``was removed'', ``instead of'', ``the original photo'', the region labels of the overlay): a leaking trailing clause is trimmed, otherwise the claim is dropped. A claim's box is snapped to its region's box when the two agree and dropped when it lies outside the region. An edit keeps its reasoning when at least two claims survive; consistency notes are listed apart from the evidence, and an edit whose claims are all consistency notes is marked as leaving \emph{no visible evidence}, kept in the test splits, where \cref{sec:supp-testsets} uses it, but not in the training data. On the L and R tiers, 95.1\% and 97.3\% of the edits keep their reasoning, 88.1\% and 91.8\% of the claims survive (5.3\% and 3.1\% dropped as leaks, 5.3\% and 4.3\% as not visible), 11.7\% and 6.4\% of the edits are marked as leaving no visible evidence, and a kept claim has a median of 31 words. On the G, S and X tiers, 94.3\%, 96.5\% and 98.0\% of the edits keep their reasoning, 89.7\%, 92.9\% and 83.9\% of the claims survive, and 40.4\%, 30.6\% and 43.8\% of the edits are marked as leaving no visible evidence; a blind judge confirms the generator's evidence for 50\%, 67\% and 57\% of their claims. Authentic photos get a rationale the same way, from the processed photo alone, with claims that never support an edit and a verifier that checks every factual detail. \Cref{sec:audit-reasoning} audits whether the kept claims are evidence of editing.

\subsection{Intent}

Two teachers from different families, Qwen3-VL-32B and Gemma-3-27B~\cite{gemmateam2025gemma3}, label each edit's intent. Each sees the source, the edit with its regions outlined, the generator's private notes and its claims, but not the instruction: Pico-Banana's instructions are uniformly polite, and including them pushed every label toward benign. A teacher gives a probability per class of \cref{sec:task-outputs}, read as how plausible it is that someone made this edit for that purpose, with the most plausible one at 0.5 or more, and a severity on the anchored scale of \cref{sec:supp-intent}. The label is the mean over the teachers: a soft label, the classes above 0.5 as hard labels, the argmax as the primary class and the mean severity, from which the coarse labels \emph{harmful} (misinformation, document, identity or privacy) and \emph{moderate or worse} (severity at least 2) follow. On L and R, the teachers agree on the primary class of 74\% of the edits and on severity at Spearman 0.58 (quadratic-weighted $\kappa$ 0.59); 59\%, 27\%, 11\% and 2.6\% of the edits are at severity 0 to 3, highest on average for age and gender (1.66) and replaced text (1.59). On G, S and X, whose edits are almost all benign, they agree on the primary class of 91\%, and 91\%, 8.7\%, 0.5\% and 0.1\% of the edits are at severity 0 to 3; over all tiers, agreement is 84\% and severity Spearman 0.55 ($\kappa$ 0.57). Agreement is lowest on humor and commercial intent, which is why \cref{sec:exp-intent} reports the coarse labels alongside the classes. The intent labels of the L and R tiers were made from the generator's claims before verification; verification removes about 8\% of the claims, so the effect on the labels is small. \Cref{sec:audit-intent} audits the labels.

\section{Training and inference details}
\label{sec:supp-training}

Our model is Qwen2.5-VL-3B-Instruct~\cite{bai2025qwen25vl} fine-tuned with LoRA~\cite{hu2022lora}. The adapters (rank 64, $\alpha = 128$, dropout 0.05) sit on the attention and MLP projections of all 36 language-model layers and on the two linear layers of the merger that passes visual tokens to the language model. The vision encoder stays frozen. This trains 120.8M parameters, 3\% of the total. Each image is encoded at its own resolution, rounded to multiples of 28 pixels, instead of being resized to a fixed square, so small edits keep their size. The experiments are conducted on one NVIDIA RTX A6000 GPU with 48\,GB VRAM.

\subsection{Training data}

\label{sec:models-data}

We fine-tune on a subset of the training split that is balanced over edit types. It holds 1,071 edits of each of the 13 edit types, the number of edits of the rarest type, so 13,923 edits in total; this is the largest subset in which every type has the same count without any image being repeated, and the rest of the training split is used only by the full-data ablation (\cref{sec:supp-ablations}). It also holds 13,923 authentic photos: the source photos of 13,560 of these edits, and 363 other photos of the training split for the edits whose source photo is shared with another drawn edit or missing from the split. Every variant of our model trains on the same 27,846 images, so the variants differ only in their targets. Balancing gives every edit type the same weight, as in our test sets (\cref{sec:exp-setup}).

\subsection{Training}

\label{sec:models-training}

Each example is the image, the question of \cref{sec:supp-prompt-ours} and the target text. The loss is the cross-entropy on the target tokens and the end-of-turn token; the image and the question are masked. No example is truncated, because a cut target would teach a broken answer; the longest one has 1,854 tokens with the image (an RG example; VB's longest has 1,484). We train for one epoch (1,741 steps) with AdamW, learning rate $10^{-4}$, a cosine schedule with 3\% warm-up, and an effective batch of 16 (one example per step, gradients accumulated over 16 steps), in bf16 with gradient checkpointing on one RTX A6000. A validation set of 260 examples from the validation split (10 edits of each edit type and 130 authentic photos) tracks the loss every 500 steps. It does not select a checkpoint: we evaluate the weights at the end of the epoch. All reported runs use seed 0; VB and RG trained again with seeds 1 and 2 land within 0.5 (VB) and 1.2 (RG) AUC points of seed 0 on the \emph{Seen} set, so the gap between them is not a property of one run (\cref{sec:supp-seeds}).

\emph{RG+RL} (reinforcement learning) continues RG's adapter with group relative policy optimization (GRPO)~\cite{shao2024deepseekmath} on rewards that need no judge: the answer parses, the verdict matches the label, and for an edited image the box F1 between its regions and the reference boxes (weights 0.25, 1 and 0.5). It samples four answers per prompt for 2,000 prompts drawn from RG's own training examples (1,000 edits balanced over the edit types and 1,000 photos), 500 steps of four prompts, learning rate $10^{-5}$, no Kullback--Leibler term.

\emph{AF} (answer first) trains on RG's images and settings with the two parts of RG's target in the opposite order: SA's structured answer and then the reasoning, and for an authentic photo the verdict and then the reasoning. Its targets therefore have RG's length, 288 tokens on average with the end-of-turn token, against 62 for SA and 41 for VB. \emph{AF+W} (AF, weighted loss) trains on AF's examples and settings with a different loss: for each example, half of the loss is the mean cross-entropy over the tokens of its structured answer (for an authentic photo, the verdict alone) and half is the mean over its other target tokens, the reasoning and the end-of-turn token, and the 16 examples of an effective batch are averaged. Every other variant averages the cross-entropy over all target tokens of those 16 examples.

\subsection{Inference}

\label{sec:models-inference}

At test time the model receives the training question and decodes greedily, with at most 768 new tokens. Its image score is the probability of ``edited'' at the verdict token, computed as for the zero-shot MLLMs (\cref{sec:supp-baselines}). For VB and SA the verdict is the first token of the answer; for RN and RG it comes after the reasoning, which by then has largely fixed it (\cref{sec:exp-indomain}). All its boxes are kept and mapped from the $0$--$1000$ grid to image pixels. An answer without a readable verdict is a parse failure and is scored as in \cref{sec:exp-setup}.

\subsection{IFDL methods fine-tuned on our data}
\label{sec:supp-ifdl-ft}

\paragraph{Data.} We draw 10,000 edits from the training split, 769 or 770 of each of the 13 edit types and one per source photo, and pair each with its own source photo as the authentic image, so the training set holds 20,000 images; 500 edits from the validation split and their source photos, drawn the same way, form the validation set. No source photo of these sets appears in the test sets.

\paragraph{Training.} Each method starts from its released weights, trains for 30 epochs on one RTX A6000 with a batch of 18 (TruFor), 8 (MVSS-Net) or 14 (PIM) images, and keeps the checkpoint that its own validation rule selects. TruFor follows its official two-phase recipe on 512-pixel crops: phase 2 trains the encoder and the localization head and keeps the checkpoint with the best validation localization F1; phase 3 trains the confidence and detection heads and keeps the one with the best validation detection balanced accuracy (89.9\%). MVSS-Net's release has no training code, so, departing from its paper's recipe, we train its official model definition with the multi-view loss of IMDL-BenCo~\cite{ma2024imdlbenco} (Dice losses on the mask and on the edge band, and a cross-entropy on the image score), AdamW with learning rate $2\times10^{-5}$, two warm-up epochs and cosine decay, and IMDL-BenCo's augmentations without its synthetic copy-move and inpainting; it keeps the epoch with the best validation pixel F1 (epoch 27). PIM uses its official training code, validated after every epoch instead of every 1,600 steps, and keeps the epoch with the best validation F1 (epoch 15). At test time all three use the inference code and decision rules of their released weights (\cref{sec:supp-baselines}).

\section{Test sets}
\label{sec:supp-testsets}

The \emph{Seen} set asks how well a method detects the kinds of edits in our training data. It holds 150 edits of each of the 13 L and R edit types of that data (1,950 edits in total). Each edit is paired with its own source photo as the authentic image, so an edited image and its authentic counterpart show the same scene and differ only by the edit. A source photo can carry several edits with different instructions; that photo enters the authentic class only once, so the 1,950 edits have 1,925 distinct source photos. To keep the classes balanced, we add 25 photos from the same test split that were never edited; they are the counterpart of no edit. The \emph{Seen} set is therefore balanced, with 1,950 edited and 1,950 authentic images.

The \emph{Unseen} set asks whether detection carries over to edit operations that are absent from our training data. It holds 150 edits of each of the three held-out L types (\textit{expression}, \textit{relocate object} and \textit{text font}), made by the same editor and pipeline as the training data, and reuses the \emph{Seen} set's 1,950 authentic photos, so any difference between the two sets comes from the edited images alone. It thus holds 450 edited and 1,950 authentic images. Since it is unbalanced, we compare it with the \emph{Seen} set by AUC, which does not depend on the class ratio.

The claim pipeline (\cref{sec:supp-pipeline}) marks 253 of the \emph{Seen} set's edits as leaving no visible evidence. We keep them: they are still edited images, and removing them would make the test easier than the real task.

\subsection{Equal quotas per edit type}
\label{sec:supp-quota}
We draw the same number of edits per type instead of following the split's mix, in which the most common type is 5.7 times as frequent as the rarest.
Every type then weighs equally in the overall numbers and has enough samples for a per-type result.

\subsection{Manual check}
\label{sec:supp-check}
Before any model was run, one author inspected 100 items (96 edits and 4 photos): 91 of the 96 edits show a visible edit of the stated type, every mask was judged good (91) or partial (5), and none of the 4 photos looks edited.

\subsection{Reference boxes}
\label{sec:supp-boxes}
The reference boxes for box F1 come from the reference mask.
We take its 8-connected regions and drop those smaller than 0.25\% of the image area, keeping the largest region if all are smaller.
Two boxes are merged when the gap between them is at most 2\% of the image diagonal, until no such pair remains.
The boxes are sorted by area, and if more than three remain, the third absorbs the rest.
A method that outputs only a map gets its boxes from the map thresholded at 0.5 by the same rule.

\subsection{Second-editor set}
\label{sec:supp-second-editor}

\begin{table}[t]
\centering\footnotesize\setlength{\tabcolsep}{3pt}
\resizebox{\linewidth}{!}{%
\begin{tabular}{l cc ccc}
\toprule
 & \multicolumn{2}{c}{Image AUC [95\% CI]} & \multicolumn{3}{c}{Qwen edits} \\
\cmidrule(lr){2-3}\cmidrule(lr){4-6}
Method & Nano-Banana & Qwen & TPR & FPR & pIoU \\
\midrule
\multicolumn{6}{l}{\textit{IFDL methods}} \\
TruFor~\cite{Guillaro_2023_CVPR} & 46.3 {\scriptsize[43.6, 48.9]} & 48.0 {\scriptsize[45.5, 50.4]} & 6.7 & 7.7 & 5.0 \\
MVSS-Net~\cite{Chen2021MVSS} & 41.3 {\scriptsize[38.5, 44.0]} & 44.7 {\scriptsize[41.9, 47.6]} & 58.2 & 65.9 & 1.1 \\
PIM~\cite{kong2025pim} & 54.0 {\scriptsize[51.5, 56.6]} & 53.5 {\scriptsize[51.0, 56.2]} & 90.3 & 89.0 & 9.5 \\
HiFi-Net~\cite{Guo_2023_CVPR} & 50.8 {\scriptsize[48.8, 52.8]} & 54.6 {\scriptsize[52.6, 56.6]} & 3.8 & 2.3 & 0.5 \\
IML-ViT~\cite{ma2023imlvit} & 47.2 {\scriptsize[44.3, 50.1]} & 52.2 {\scriptsize[49.1, 55.4]} & 99.5 & 99.5 & 3.8 \\
\midrule
\multicolumn{6}{l}{\textit{Forensic MLLMs}} \\
SIDA-7B~\cite{Huang_2025_CVPR} & 57.6 {\scriptsize[55.6, 59.7]} & 61.1 {\scriptsize[58.7, 63.5]} & 9.5 & 3.3 & 1.5 \\
FakeShield$^\dagger$~\cite{xu2025fakeshield} & 53.6 {\scriptsize[51.8, 55.4]} & 50.3 {\scriptsize[48.6, 52.1]} & 17.4 & 16.9 & 1.7 \\
So-Fake-R1$^\dagger$~\cite{huang2026sofakebenchmarkingexplainingsocial} & 50.6 {\scriptsize[48.5, 52.8]} & 51.5 {\scriptsize[49.4, 53.8]} & 31.3 & 28.2 & 2.8 \\
\midrule
\multicolumn{6}{l}{\textit{Zero-shot MLLMs}} \\
Qwen2.5-VL-3B$^\ddagger$~\cite{bai2025qwen25vl} & 7.5 {\scriptsize[5.8, 9.3]} & 8.4 {\scriptsize[6.5, 10.3]} & 42.1 & 100.0 & 7.2 \\
Qwen2.5-VL-7B$^\ddagger$~\cite{bai2025qwen25vl} & 50.0 {\scriptsize[48.0, 51.9]} & 51.4 {\scriptsize[49.3, 53.5]} & 89.2 & 93.3 & 11.1 \\
Qwen2.5-VL-32B (4-bit)~\cite{bai2025qwen25vl} & 52.1 {\scriptsize[50.3, 54.0]} & 56.0 {\scriptsize[53.9, 58.2]} & 17.7 & 11.0 & 1.4 \\
Qwen3-VL-8B~\cite{bai2025qwen3vltechnicalreport} & 55.2 {\scriptsize[53.4, 57.0]} & 59.7 {\scriptsize[57.5, 61.8]} & 61.8 & 48.7 & 13.8 \\
InternVL3-8B~\cite{zhu2025internvl3} & 52.8 {\scriptsize[51.3, 54.3]} & 55.1 {\scriptsize[53.4, 56.9]} & 25.6 & 19.2 & 2.8 \\
\midrule
\multicolumn{6}{l}{\textit{IFDL methods fine-tuned on our data}} \\
TruFor~\cite{Guillaro_2023_CVPR} & 94.1 {\scriptsize[92.7, 95.4]} & 41.2 {\scriptsize[38.2, 44.2]} & 7.9 & 7.2 & 3.5 \\
MVSS-Net~\cite{Chen2021MVSS} & 100.0 {\scriptsize[100.0, 100.0]} & 69.0 {\scriptsize[66.2, 71.9]} & 1.5 & 0.3 & 0.6 \\
PIM~\cite{kong2025pim} & 93.2 {\scriptsize[91.6, 94.8]} & 70.3 {\scriptsize[67.6, 73.0]} & 43.1 & 14.6 & 9.2 \\
\midrule
\multicolumn{6}{l}{\textit{Ours: Qwen2.5-VL-3B~\cite{bai2025qwen25vl} with LoRA~\cite{hu2022lora} (\cref{sec:models})}} \\
VB: verdict and boxes & 78.8 {\scriptsize[76.5, 81.2]} & 66.9 {\scriptsize[64.2, 69.6]} & 39.5 & 18.7 & 15.0 \\
SA: VB + type and intent & 79.5 {\scriptsize[77.3, 81.9]} & 69.3 {\scriptsize[66.7, 72.1]} & 33.1 & 9.5 & 12.7 \\
RU: RN on unverified claims & 54.3 {\scriptsize[51.5, 57.1]} & 56.4 {\scriptsize[53.3, 59.5]} & 29.5 & 22.1 & 9.6 \\
RN: RG, no inline boxes & 53.7 {\scriptsize[50.8, 56.7]} & 54.5 {\scriptsize[51.8, 57.3]} & 27.7 & 21.8 & 8.5 \\
RG: grounded reasoning & 53.4 {\scriptsize[50.2, 56.6]} & 53.4 {\scriptsize[50.5, 56.6]} & 24.1 & 17.9 & 8.1 \\
RG+RL: RG + GRPO & 59.4 {\scriptsize[56.2, 62.7]} & 58.1 {\scriptsize[55.2, 61.3]} & 49.7 & 36.9 & 17.0 \\
AF: RG, answer first & 58.4 {\scriptsize[56.6, 60.3]} & 61.1 {\scriptsize[59.0, 63.3]} & 6.4 & 3.3 & 2.5 \\
AF+W: AF, weighted loss & 72.4 {\scriptsize[70.1, 74.8]} & 64.3 {\scriptsize[61.8, 67.2]} & 11.3 & 5.1 & 4.4 \\
\bottomrule
\end{tabular}}
\caption{\textbf{A second editor on the same photos and instructions.} The 390 source photos against their Nano-Banana edits (the same edits as in \emph{Seen}, under this processing) and against their Qwen-Image-Edit edits (\cref{sec:exp-setup}); percentages, AUC with its 95\% bootstrap interval over source photos. TPR and pIoU are on the Qwen edits at the fixed threshold; FPR is on the shared photos. $^\dagger$No probability output: AUC ranks by the verdict. $^\ddagger$A parse failure counts as a wrong verdict with the worst-case score.}
\label{tab:second-editor}
\end{table}

To separate the editor from the edit, we re-edit 390 \emph{Seen} sources (30 per edit type, one per photo) with Qwen-Image-Edit-2509~\cite{wu2025qwenimage} from the same Pico-Banana instructions (40 steps, guidance 4.0; examples in the supplement). The set holds each source photo, its Nano-Banana edit and its Qwen edit, 1,170 images standardized together as in \cref{sec:task-input} (864 px long side for 354 of the 390 groups, 640 for the rest), with masks from one difference-based generator for both editors. The Nano-Banana edits are a control: the same photos and edits as in \emph{Seen}, under this processing. The Nano-Banana numbers are below those of \emph{Seen} for the same edits (VB 78.8 against 93.1, fine-tuned TruFor 94.1 against 99.1): here each photo is standardized together with its edits, so the two classes share one processing path, which the test sets of \cref{sec:exp-setup} do not guarantee (\cref{sec:supp-limitations}).

\subsubsection{Generation} 

Qwen-Image-Edit-2509~\cite{wu2025qwenimage} re-edits each source photo from its Pico-Banana instruction with 40 denoising steps, a true classifier-free guidance scale of 4.0 and a seed fixed per edit, in bf16 on one RTX A6000 with model offloading: a median of 190 s and 39 GB per edit. The 390 sources are 30 per edit type, one per photo, drawn by a seeded round-robin over the \emph{Seen} edits. The editor keeps the aspect ratio of the source and returns about 1,250 px on the long side, as Nano-Banana does.

\subsubsection{Set} 

The photo, its Nano-Banana edit and its Qwen edit form one group and go through the standardization of \cref{sec:task-input} together, so the three classes share one size tier and one processing history. The reference masks of both editors come from the same difference-based generator as the training masks, run with its whole-image routing off; on the Nano-Banana edits they agree with the training masks at a median IoU of 0.42 (70\% above 0.3), the difference being how the pair is aligned, so pixel metrics on this set are comparable between the editors but not with \cref{tab:e1-baselines}.

\subsubsection{Examples} 

\begin{figure}[t]
\centering
\includegraphics[width=\linewidth]{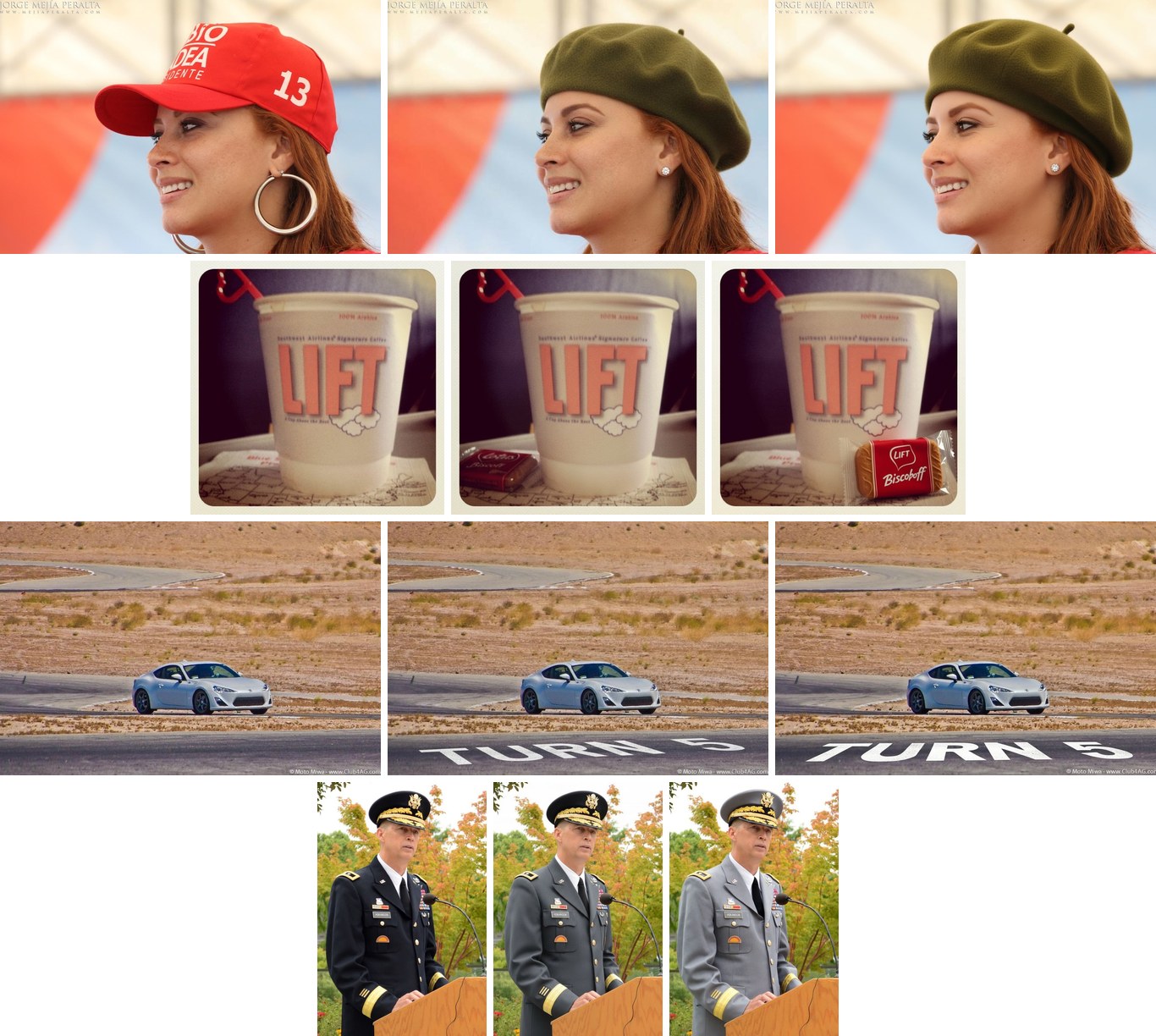}
\caption{\textbf{Second editor.} Source photo (left), its Nano-Banana edit (middle) and its Qwen-Image-Edit edit (right), all standardized together, for four instructions: replace the cap with an olive beret and the hoops with stud earrings; add a pack of Biscoff cookies next to the cup; paint ``TURN 5'' on the asphalt; change the jacket to charcoal gray.}
\label{fig:second-editor}
\end{figure}

\Cref{fig:second-editor} shows four groups. Qwen follows the instruction as closely as Nano-Banana and sometimes more literally; for a few prompts it regenerates the scene instead of editing it, and the difference mask then covers most of the image.

\subsection{MagicBrush set}
\label{sec:supp-magicbrush}

To test another editor, we use the MagicBrush dev split~\cite{zhang2023magicbrush}: 528 DALL-E 2 edits made from human instructions, with human-drawn masks, and its 263 first-turn photos as authentic images (later turns start from an edit; 3 first-turn photos are themselves DALL-E 2 outputs and are left out). Every edit carries the DALL-E 2 watermark in its bottom-right corner, so we crop the bottom 2.5\% of every image and mask before standardizing as in \cref{sec:automatic}.

\subsection{MLLM-Edit set}
\label{sec:supp-mllm-edit}

MLLM-Edit~\cite{yu2026mllmedit} edits photos from ten public datasets with 18 MLLM-based editors, one insertion, removal or replacement per photo. We use the natural edits of its public release, which holds each editor's photos and edited images; its masks and instructions are not released yet, so we score detection only. The photos are small (a median long side of 600 px), and most editors return a larger image, up to about four times the photo's size. For each editor we draw up to 30 pairs per source dataset, with a fixed seed, among the pairs that our standardization can keep without enlarging either image; LSUN Church, whose photos are 256 px, gives about one pair per editor. This gives 270--275 pairs per editor, 4,893 in all, from 4,834 distinct photos (59 photos appear with two to four editors). Each photo and its edit are standardized together as in \cref{sec:supp-second-editor}, to a long side of 416 px for 52\% of the images, 512 for 17\%, 640 for 8\% and 864 for 22\%; masks from the same difference-based generator serve only to register and crop the pair. Since most editors enlarge the image, the edit is usually reduced more than its photo.

\section{Baselines, metrics and prompts}
\label{sec:supp-baselines}
\label{sec:supp-metrics}

\paragraph{Baselines.} We evaluate three groups with their released weights and no training on our data.

The first group is IFDL methods (\cref{sec:related-imdl}): TruFor~\cite{Guillaro_2023_CVPR}, MVSS-Net~\cite{Chen2021MVSS}, PIM~\cite{kong2025pim}, HiFi-Net~\cite{Guo_2023_CVPR} and IML-ViT~\cite{ma2023imlvit}. TruFor's image score comes from its detection head and HiFi-Net's is $1-P(\text{real})$ from its classifier; MVSS-Net, PIM and IML-ViT have no image-level head, so their score is the maximum of the predicted probability map. IML-ViT, used with its released checkpoint trained under the protocol that TruFor also follows, reads a $1024 \times 1024$ window, so a larger image is covered by evenly spaced windows whose probabilities are averaged where they overlap. HiFi-Net's map is a distance $d$ from the centre of authentic pixels, which we map to $1-2^{-d/2.3}$, so that 0.5 falls at its official threshold of 2.3. An image is predicted as edited when its score is at least 0.5, and the pixels whose probability is at least 0.5 form the predicted edited region.

The second group is forensic MLLMs, which are trained to detect, localize and explain manipulations (\cref{sec:related-mllm}): SIDA-7B~\cite{Huang_2025_CVPR}, FakeShield~\cite{xu2025fakeshield} and So-Fake-R1~\cite{huang2026sofakebenchmarkingexplainingsocial}. Each uses its own prompt and decision rule. SIDA's score is $1-P(\text{real})$ from its class head, with its mask when it predicts a tampered image. FakeShield gives a text verdict and a mask; So-Fake-R1 gives a final label and a box. Neither gives a probability, so their AUC ranks by the verdict.

The third group is general MLLMs using zero-shot: Qwen2.5-VL at 3B, 7B and 32B (4-bit)~\cite{bai2025qwen25vl}, Qwen3-VL-8B~\cite{bai2025qwen3vltechnicalreport} and InternVL3-8B~\cite{zhu2025internvl3}. Each is asked one question (\cref{sec:supp-prompt-zeroshot}) and must answer with a JSON object whose first key is the verdict, followed by the scope and up to three boxes. The score is the probability of ``edited'' at the verdict token, renormalized over ``edited'' and ``authentic'' (each with its single-token synonyms).

\paragraph{Metrics.} Pixel IoU compares the predicted region with the reference mask and is averaged over images; the predicted region is the thresholded map for methods that output a map and the union of the predicted boxes for the others.

Box F1 compares the predicted boxes with up to three reference boxes derived from the reference mask (\cref{sec:supp-boxes}). The reference merges nearby regions and keeps at most three boxes, so a prediction that splits or merges the regions differently from the reference is penalized even when it covers the same pixels.

Most methods give no confidence for each box, so average precision is not defined for them, and we report F1 instead: after Hungarian matching, a predicted box is correct when its IoU with the matched reference box is at least 0.5. For methods that output only a map, we derive boxes from the thresholded map in the same way as for the reference. An answer that cannot be parsed counts as a wrong verdict with the worst possible score and an empty prediction, so that failing to answer never helps a method, and we report the parse rate alongside. We report 95\% confidence intervals from a bootstrap with 2,000 resamples. An edit and its source photo show the same scene and are not independent, so the bootstrap resamples source photos, each together with its edits. The thresholds of 0.5 on scores and maps are fixed in advance rather than tuned on the test sets, so accuracy, TPR and FPR show each method as released; AUC does not depend on a threshold.

\subsection{Prompt of our model}
\label{sec:supp-prompts}
\label{sec:supp-prompt-ours}
Every training example and every test query of our model gives the image followed by the question of our training data:
\begin{quote}\small\ttfamily\raggedright
Is this image edited? If it is, explain the visual evidence, give the edited regions as [x0, y0, x1, y1] boxes in 0-1000 coordinates, the edit type, and the likely intent with a severity from 0 to 3.
\end{quote}

The question is the same for every variant of \cref{sec:models-variants}, so VB and SA are trained to answer less than it asks; we keep one question so that the variants differ only in their targets.

\subsection{Prompt of the zero-shot MLLMs}
\label{sec:supp-prompt-zeroshot}
The zero-shot MLLMs receive the image followed by this prompt (line breaks as sent):
\begin{quote}\small\ttfamily\raggedright
Is this photo authentic, or has it been edited, for example with an AI image editor or photo-editing software? Look closely for objects, text or people that were added, removed or changed, and for changes to lighting, colour or style.\\[0.5em]
Answer with this JSON object inside <answer></answer> tags and nothing else:\\
<answer>\{"verdict": "edited" | "authentic", "scope": "local" | "global" | "none", "boxes": [[x0, y0, x1, y1], ...]\}</answer>\\
- "verdict": "edited" if any part of the image was changed, "authentic" if it is an unmodified photo.\\
- "scope": "local" if the edit changed particular regions or objects, "global" if it changed the whole image (a filter, lighting, colour tone, season, weather or style); "none" for an authentic photo.\\
- "boxes": up to 3 boxes around the edited regions, each [x0, y0, x1, y1] in pixel coordinates of the image (x to the right, y down); [] for an authentic photo or a global edit.
\end{quote}
Qwen3-VL-8B and InternVL3-8B give boxes on a $0$--$1000$ grid, so for them the last line reads ``coordinates scaled to 0-1000 across the image's width and height'' instead of ``pixel coordinates of the image''.
SIDA-7B, FakeShield and So-Fake-R1 use the prompts of their releases.

\section{In-domain results}
\label{sec:supp-indomain}

No existing detector reaches 60\% image AUC on the \emph{Seen} set (\cref{tab:e1-baselines}). IML-ViT, which \cref{tab:e1-baselines} leaves out for space, reaches 47.1 AUC [45.8, 48.6] on the \emph{Seen} set and 46.8 [43.9, 49.9] on the \emph{Unseen} set, with pixel IoU 3.9 and box F1 2.8. The best are SIDA (58.0), PIM (57.4) and Qwen3-VL-8B (55.6), and pixel IoU stays at or below 10.4\%. The detectors fail in opposite directions. Qwen2.5-VL-7B, PIM and IML-ViT call most images edited (TPR 89.7, 93.7 and 99.2 at FPR 91.4, 89.1 and 99.4). SIDA, HiFi-Net, Qwen2.5-VL-32B and TruFor call most images authentic (TPR at most 14.9). Accuracy stays near 50\% either way. MVSS-Net, IML-ViT and TruFor rank the authentic photos above the edits (AUC 43.4, 47.1 and 47.3, all three intervals below 50). Differences in processing do not explain this: in the trace test of \cref{sec:exp-generalization}, where an edit and its source photo are processed identically, released TruFor still ranks the source above the edit (AUC 48.6 [47.2, 49.9]), and so do MVSS-Net (39.9 [37.9, 41.9]) and IML-ViT (44.0 [42.0, 46.1]; \cref{tab:supp-matched}), so the reason lies in the images themselves. Of Qwen2.5-VL-3B's answers, 65\% (2,522) do not parse: 1,641 of them write \texttt{()} after a box, 818 repeat a box until the token limit and 63 have other JSON errors. The pre-specified rule scores these answers as failures (AUC 6.2). Every one of them states a complete verdict, always ``edited''; rescoring them with that verdict gives 49.8, because the model then calls every image edited.

Rescoring the failed answers that state a complete verdict with that verdict gives image AUC Qwen2.5-VL-3B 49.8 on Seen (2,522 answers repaired) and 49.2 on Unseen; Qwen2.5-VL-7B 51.9 on Seen (102 answers repaired) and 50.0 on Unseen (sensitivity analysis; the pre-specified numbers are those in the table).

Its FPR is 3.8 on the 1,925 photos that are paired with an edit and 12.0 on the 25 unpaired ones (3 photos).

RG, trained on the same images to write its grounded reasoning before the verdict, stays near chance: 53.1 AUC [51.7, 54.5], TPR 20.3 at FPR 13.1, box F1 9.6. All its answers parse, but it calls 80\% of the edits authentic. Two properties of its answers explain this. First, its score is read at the verdict token, which comes after the reasoning, and by then the reasoning has fixed the verdict: every answer whose reasoning closes with ``Together these cues indicate\ldots'' says ``edited'' (651 of 651), and every one that closes with ``I find no evidence\ldots'' says ``authentic'' (3,249 of 3,249). Every score is therefore below 0.01 or above 0.99, and ranking by the score is ranking by the verdict: the AUC (53.1) is the verdicts' balanced accuracy (53.6) up to noise (\cref{sec:supp-e-score}). Second, the verdict is decided much earlier, by whether the first claim of the reasoning cites a box, as in the training targets, where every edited answer cites boxes and no authentic one does: on the \emph{Seen} set, a first claim that cites a box ends in ``edited'' 96.5\% of the time and one that does not ends in ``authentic'' 98.9\% of the time, and RG's first claim cites a box for only 19.8\% of the edits (and 12.9\% of the photos). Taking the probability of a box at that point as the score, a diagnostic we designed after seeing these results, makes the scores graded but raises AUC only to 57.1 [55.6, 58.5] on \emph{Seen} ($+4.0$, paired bootstrap [2.4, 5.6]) and 56.9 [53.8, 59.8] on \emph{Unseen} ($+1.9$, [$-1.5$, 5.1]); the pre-specified numbers are those in \cref{tab:e1-baselines}. The evidence is in the images: among the 3,249 images that RG calls authentic, VB's score separates edits from photos at 92.7 AUC, RG's box probability at 55.2. The full answer, as we train it, thus costs about 40 AUC points against VB. VB and RG also differ in the edit type and intent that RG writes, but SA, which adds these fields to VB without reasoning, reaches 92.7 AUC [91.9, 93.4], within 0.4 of VB, and RN, which keeps RG's reasoning but removes the box references from its claims, reaches 57.7 [56.4, 59.1] (\cref{tab:e1-baselines}). The cost therefore comes from the reasoning, not from the extra fields, and the inline box references account for about 5 of its 40 points; the order accounts for about 12 more and a weighted loss for a further 22, as AF and AF+W show below. RU, trained on the unverified claims, reaches 55.2 [53.8, 56.6], close to RN, so verification changes little for detection. RG+RL, RG's adapter tuned with GRPO on the verdict and box rewards (\cref{sec:models-variants}), answers ``edited'' for 41\% of the images where RG does for 17\%, which raises its TPR to 47.2 at FPR 34.7 and its box F1 to 18.6, but its AUC only to 57.1 [55.7, 58.5]: the verdict reward stayed between 0.46 and 0.56 (chance is 0.5) in every 25-step block of the 500 steps, so the tuning moved the threshold, not the ranking.

AF tests how much of this cost is the order. It writes SA's structured answer first and RG's reasoning after it, so its verdict token is predicted from the image alone, as SA's is. It reaches 65.3 AUC [64.3, 66.2] on the \emph{Seen} set and 65.1 [62.5, 67.7] on the \emph{Unseen} set: 12.2 points above RG on \emph{Seen} (paired bootstrap [10.4, 13.9]) and 27.4 below SA ([26.4, 28.5]). All its answers parse, and its reasoning always agrees with its verdict: every answer that says ``edited'' closes with ``Together these cues indicate\ldots'' (180 of 180) and all but one of the others with ``I find no evidence\ldots'' (3,719 of 3,720). But it calls 93\% of the edits authentic (TPR 6.6 at FPR 2.6). Unlike RG's, its scores are graded, yet they hardly separate the classes: their median is 0.32 on the edits and 0.27 on the photos, against 0.50 and 0.02 for SA. The order thus accounts for about 12 of the 40 points, and most of the cost remains when the verdict is written first. AF resembles VB trained on a quarter of the edits (71.6 AUC, 11.6\% of the edits called edited; \cref{sec:supp-ablations}), a model whose verdict has had little training. AF+W tests one reading of this, that AF's verdict receives a small share of the loss: it is one token of a target of 288 on average, against 62 for SA. AF+W trains on AF's targets with half of each example's loss on the structured answer and half on the rest of the target (\cref{sec:models-training}). It reaches 87.2 AUC [86.2, 88.1] on the \emph{Seen} set and 86.3 [84.6, 87.9] on the \emph{Unseen} set: 22.0 points above AF on \emph{Seen} (paired bootstrap [20.9, 23.0]) and 5.5 below SA ([4.9, 6.1]). Its scores separate the classes again, with a median of 0.29 on the edits and 0.04 on the photos, and its reasoning agrees with its verdict in all 3,900 answers. It still calls 72\% of the edits authentic (TPR 28.0 at FPR 1.7), and its boxes stay below SA's (box F1 14.2 against 21.7). The weighted loss thus recovers most of what the order does not: a model that writes its reasoning after a weighted answer detects 5.5 points below SA, not 27. This loss also averages over the examples where the standard loss averages over their tokens, so a short target counts as much as a long one, and one run does not separate the two changes. On the other sets AF reaches 54.9 AUC on MagicBrush and 58.4 and 61.1 on the Nano-Banana and Qwen edits of the second-editor set, and AF+W 50.9, 72.4 and 64.3 (\cref{tab:cross-editor,tab:second-editor}). AF answers an intent for the 129 \emph{Seen} edits it detects and agrees with the pipeline's primary intent on 59.7\% of them, and AF+W for 546 and 58.6\%, against 943 and 64.2\% for SA.

\subsection{RG's image score}
\label{sec:supp-e-score}

For a score that takes only two values, one per verdict, the AUC equals the balanced accuracy of the verdicts. Write $t$ for the TPR and $f$ for the FPR. A random pair of one edit and one photo is ranked correctly when the edit is called edited and the photo authentic (probability $t(1-f)$), ranked wrongly in the opposite case, and tied otherwise; a tie counts one half. The AUC is then $t(1-f) + \tfrac12 tf + \tfrac12 (1-t)(1-f) = \tfrac12 (t + 1 - f)$, the balanced accuracy. On the \emph{Seen} set, RG calls 396 of the 1,950 edits and 255 of the 1,950 photos edited, so $t = 20.3\%$, $f = 13.1\%$ and the balanced accuracy is 53.6\%.

RG's scores are not exactly two values: every ``authentic'' answer scores below $10^{-7}$ and every ``edited'' one above $0.999999$, but within each verdict the scores still differ slightly, and these differences rank pairs with the same verdict. With $w_1$ and $w_0$ the AUC among the images called edited and among those called authentic,
\begin{equation}
\text{AUC} = t(1-f) + t f\, w_1 + (1-t)(1-f)\, w_0 .
\end{equation}
For RG, $w_1 = 50.7\%$ and $w_0 = 49.2\%$, both at chance, so $\text{AUC} = 17.65 + 1.35 + 34.08 = 53.1\%$, the value in \cref{tab:e1-baselines}. The half point below the balanced accuracy comes from the images RG calls authentic, where these tiny differences rank the photos slightly above the edits.

\subsection{Seeds}
\label{sec:supp-seeds}

\begin{table}[t]
\centering\footnotesize\setlength{\tabcolsep}{3pt}
\resizebox{\linewidth}{!}{%
\begin{tabular}{l c ccccc}
\toprule
Variant & Seed & \emph{Seen} & \emph{Unseen} & MagicBrush & Nano-Banana & Qwen \\
\midrule
VB: verdict and boxes & 0 & 93.1 & 91.3 & 48.0 & 78.8 & 66.9 \\
 & 1 & 92.8 & 90.8 & 45.7 & 78.1 & 66.7 \\
 & 2 & 92.9 & 91.0 & 46.6 & 79.1 & 67.8 \\
\midrule
RG: grounded reasoning & 0 & 53.1 & 54.9 & 52.6 & 53.4 & 53.4 \\
 & 1 & 54.3 & 55.0 & 53.8 & 55.0 & 54.2 \\
 & 2 & 53.2 & 52.2 & 46.0 & 54.0 & 54.7 \\
\bottomrule
\end{tabular}}
\caption{\textbf{Seeds.} Image AUC of VB and RG trained with seeds 0, 1 and 2 on the \emph{Seen}, \emph{Unseen} and MagicBrush sets (\cref{sec:exp-setup}) and on the second-editor set per editor (\cref{tab:second-editor}); seed 0 is the run reported everywhere else.}
\label{tab:supp-seeds}
\end{table}

VB and RG were trained twice more with seeds 1 and 2, with everything else as in \cref{sec:models-training}, and scored on every test set (\cref{tab:supp-seeds}). VB's AUC varies by at most 0.5 points across seeds on the \emph{Seen} and \emph{Unseen} sets and by 2.3 on MagicBrush, where it is near chance; RG's by at most 2.8 points on our test sets and by 7.8 on MagicBrush, where every seed is near chance. Every number in the main text is from seed 0.

\subsection{Ablation runs}
\label{sec:supp-ablations}

Each run of \cref{sec:exp-ablations} is VB retrained with one change, with the settings and seed 0 of \cref{sec:models-training} and, except for VB-7B, its 3B model; the number of steps follows the number of examples (one epoch, an effective batch of 16). \textbf{VB-7B} trains VB's recipe on Qwen2.5-VL-7B-Instruct with the same data, settings and seed (1,741 steps). \textbf{VO} trains on VB's 27,846 images with the verdict as its only target, \texttt{\{"verdict": "edited"\}}, so it writes no boxes (1,741 steps). \textbf{VB-50} and \textbf{VB-25} train on nested subsets of VB's edits, balanced over the 13 edit types (535 or 536 and 267 or 268 edits per type), each with its photos: 6,962 edits and 6,962 photos (871 steps) and 3,481 and 3,481 (436 steps). \textbf{VB-full} trains on every local and regional edit of the training split, 69,754 edits that are not balanced over the edit types, and as many authentic photos (139,508 examples, 8,720 steps; 57.8 hours on one RTX A6000 against VB's 12.9). \textbf{VB+U} trains on VB's 27,846 examples and 6,000 more, built from 3,000 of VB's training edits with the processing of the trace test (\cref{sec:supp-trace}): T, the edit, labelled edited with its boxes mapped to the new grid, and U, the edit with the region $C$ put back from the source, labelled authentic. T and U of a pair go through the same processing, so within the added pairs the label is balanced and the processing carries no label. VB+U thus sees no new edit, only 3,000 of VB's edits a second time; it has 33,846 examples and 2,116 steps. The trace test (630 pairs) uses test edits only.

\begin{table}[t]
\centering\footnotesize\setlength{\tabcolsep}{3pt}
\resizebox{\linewidth}{!}{%
\begin{tabular}{l r c cc cc c cc}
\toprule
 & & & \multicolumn{2}{c}{\emph{Seen}} & \multicolumn{2}{c}{\emph{Unseen}} & MagicBrush & \multicolumn{2}{c}{Second editor} \\
\cmidrule(lr){4-5}\cmidrule(lr){6-7}\cmidrule(lr){8-8}\cmidrule(lr){9-10}
Variant & Edits & Boxes & AUC & pIoU & AUC & pIoU & AUC & Nano-Banana & Qwen \\
\midrule
VB (\cref{sec:models-variants}) & 13,923 & yes & 93.1 & 16.4 & 91.3 & 11.0 & 48.0 & 78.8 & 66.9 \\
VB on Qwen2.5-VL-7B & 13,923 & yes & 95.6 & 20.0 & 94.3 & 13.8 & 41.3 & 85.7 & 67.3 \\
VO: verdict only & 13,923 & no & 93.2 & -- & 92.6 & -- & 44.3 & 78.4 & 62.3 \\
VB, all L and R edits & 69,754 & yes & 96.6 & 22.8 & 94.9 & 14.9 & 43.9 & 87.1 & 72.1 \\
VB, half the edits & 6,962 & yes & 87.5 & 11.6 & 86.4 & 8.8 & 48.4 & 72.1 & 61.6 \\
VB, a quarter of the edits & 3,481 & yes & 71.6 & 3.2 & 70.9 & 1.3 & 51.6 & 64.8 & 62.1 \\
VB $+$ 3,000 editor-trace pairs & 13,923 & yes & 93.4 & 15.7 & 90.7 & 9.7 & 52.2 & 81.4 & 70.1 \\
\bottomrule
\end{tabular}}
\caption{\textbf{Ablations of VB.} Image AUC and pixel IoU on the \emph{Seen}, \emph{Unseen} and MagicBrush sets (\cref{sec:exp-setup}) and image AUC on the second-editor set per editor (\cref{tab:second-editor}). Each row retrains VB (seed 0) with one change (\cref{sec:exp-ablations}). Edits: training edits, each paired with its source photo (the 3,000 trace pairs re-process edits that VB already trains on, so VB+U has no new edits); a cell is -- where the variant predicts no boxes.}
\label{tab:ablations}
\end{table}

We retrain VB with one change at a time and score the result as before (\cref{tab:ablations}). Removing the box targets changes detection little on our sets: VO, trained to write the verdict alone, reaches 93.2 AUC on the \emph{Seen} set against VB's 93.1 and 92.6 against 91.3 on the \emph{Unseen} set (paired bootstrap $+1.2$ [0.7, 1.8]), and the same 78.4 on the second editor's Nano-Banana photos, but 62.3 against 66.9 on the Qwen edits ($-4.6$ [$-6.0$, $-3.3$]). It answers ``edited'' more readily (TPR 84.5 at FPR 14.8, against VB's 60.1 at 3.9), so on our sets the box targets move VB's threshold, not its ranking; what they add is the boxes, and a few points on the other editor. The amount of training data matters more: half of the edits cost 5.6 AUC points on the \emph{Seen} set (87.5) and 4.9 on the \emph{Unseen} set (86.4), and a third of the localization (pixel IoU 11.6 against 16.4); a quarter costs 21.5 and 20.4 points (71.6 and 70.9) and almost all of the localization (pixel IoU 3.2), and the model then calls only 11.6\% of the edits edited. Detection at 13,923 edits is thus limited by the data, not by the supervision, and VB-full shows it from the other side: with all 69,754 edits of the training split it reaches 96.6 AUC on the \emph{Seen} set (paired bootstrap against VB $+3.5$ [3.0, 4.0]; TPR 75.9 at FPR 3.8; pixel IoU 22.8) and 94.9 on the \emph{Unseen} set ($+3.6$ [2.7, 4.5]), at or above VB on the 7B model (95.6 and 94.3) and still below the fine-tuned IFDL detectors (98.7 to 100.0). On the other editor the data size matters less: 62.1, 61.6, 66.9 and 72.1 AUC on the Qwen edits at 3,481, 6,962, 13,923 and 69,754 edits ($+5.2$ [3.3, 7.3] for VB-full against VB), against 64.8, 72.1, 78.8 and 87.1 on the Nano-Banana edits of the same photos ($+8.3$ [6.4, 10.1]). On MagicBrush VB-full falls to 43.9 ($-4.1$ [$-6.9$, $-1.3$]), so five times the edits do not carry the detector to an unrelated editor. VB+U, which adds 3,000 editor-trace pairs, leaves detection on our sets where it was and raises it on the other editor. We take 3,000 of VB's training edits and make two versions of each, processed alike: the edit, labelled edited, and the edit with its region put back from the source, labelled authentic (the \emph{U} version of \cref{sec:exp-generalization}); no new edit is added. VB+U reaches 93.4 AUC on the \emph{Seen} set (paired bootstrap against VB $+0.3$ [$-0.0$, 0.6]; TPR 55.4 at FPR 3.1) and 90.7 on the \emph{Unseen} set ($-0.6$ [$-1.1$, $-0.0$]), and 70.1 on the Qwen edits ($+3.2$ [1.9, 4.6]) and 81.4 on the Nano-Banana photos ($+2.6$ [1.5, 3.8]), gains larger than the spread of VB's three seeds on these sets (at most 1.1 points, \cref{tab:supp-seeds}), though VB+U is a single seed. The trace stays: in the trace test VB+U separates \emph{U} from \emph{S} at 69.1 AUC against VB's 69.9 (paired $-0.8$ [$-1.6$, $-0.1$]) and \emph{T} from \emph{S} at 88.1 against 88.0 ($+0.1$ [$-0.6$, 0.9]). It calls fewer \emph{U} images edited (6.7\% against 18.3\%), but also fewer edits (27.6\% against 46.3\%) and fewer source photos (2.4\% against 5.1\%), so the threshold moves and the ranking hardly does. On MagicBrush VB+U reaches 52.2 AUC ($+4.2$ [2.4, 5.8]), above VB's three seeds (45.7--48.0) and the highest of the ablations (44--52), but still near chance. A larger backbone learns the same trace. Trained with VB's data and settings, Qwen2.5-VL-7B reaches 95.6 AUC on the \emph{Seen} set and 94.3 on the \emph{Unseen} set (TPR 70.5 at FPR 4.3; pixel IoU 20.0), and 85.7 on the second editor's Nano-Banana photos, but 67.3 on its Qwen edits against VB's 66.9, 41.3 on MagicBrush, and in the trace test separates \emph{U} from \emph{S} at 70.5 AUC against VB's 69.9 (paired $+0.6$ [$-1.1$, 2.2]) and \emph{T} from \emph{S} at 89.9 against 88.0 ($+1.9$ [0.4, 3.4]). Two more variants are in \cref{sec:supp-ablations}: RG trained with one fixed sentence in place of every photo's rationale stays at 53.1 AUC, and SA trained on one teacher's intent alone matches the two-teacher SA on every set.

Two variants test the supervision of the reasoning models. RG without photo rationales replaces the rationale of every authentic training photo by one fixed sentence that keeps the reasoning block and its format; it reaches 53.1 AUC on the \emph{Seen} set [51.8, 54.2], as RG does, and calls 3.4\% of the edits and 0.2\% of the photos edited, so the photos' rationales are not what holds RG near chance. SA on one teacher trains on the intent labels of the Qwen teacher alone instead of the two-teacher mean; it reaches 92.6 AUC on the \emph{Seen} set against SA's 92.7 and 90.3 against 90.1 on the \emph{Unseen} set, answers an intent for 45.8\% of the \emph{Seen} edits against 48.4\%, and agrees with the pipeline's primary intent on 29.7\% of them against 31.0\% (severity Spearman 0.48 against 0.50).

\subsection{Per edit type}
\label{sec:supp-per-type}

\begin{table}[t]
\centering\footnotesize\setlength{\tabcolsep}{3pt}
\begin{tabular}{l r ccccc}
\toprule
Edit type & $n$ & TPR & pIoU & box F1 & box P & box R \\
\midrule
\multicolumn{7}{l}{\emph{Seen}} \\
add object & 150 & 70.0 & 16.4 & 22.3 & 37.8 & 15.8 \\
scene background & 150 & 70.0 & 35.5 & 60.3 & 83.0 & 47.3 \\
clothing & 150 & 68.7 & 19.5 & 33.1 & 53.6 & 23.9 \\
replace object & 150 & 68.0 & 24.3 & 35.4 & 51.7 & 26.9 \\
replace text & 150 & 65.3 & 13.5 & 17.7 & 26.7 & 13.3 \\
add text & 150 & 64.0 & 15.9 & 20.7 & 32.0 & 15.3 \\
age gender & 150 & 62.7 & 16.5 & 22.9 & 39.8 & 16.0 \\
pose & 150 & 60.7 & 18.3 & 36.4 & 63.4 & 25.5 \\
object attribute & 150 & 60.0 & 17.5 & 34.3 & 64.1 & 23.4 \\
accessories & 150 & 57.3 & 10.1 & 19.1 & 43.0 & 12.3 \\
object geometry & 150 & 48.0 & 11.4 & 23.4 & 48.8 & 15.4 \\
translate text & 150 & 47.3 & 8.4 & 17.2 & 45.6 & 10.6 \\
remove object & 150 & 38.7 & 5.5 & 9.2 & 23.8 & 5.7 \\
\midrule
\multicolumn{7}{l}{\emph{Unseen}} \\
expression & 150 & 56.7 & 12.0 & 26.5 & 54.3 & 17.5 \\
relocate object & 150 & 52.0 & 11.7 & 20.0 & 43.9 & 12.9 \\
text font & 150 & 48.7 & 9.2 & 15.7 & 36.7 & 10.0 \\
\bottomrule
\end{tabular}
\caption{\textbf{VB per edit type.} Detection and localization of VB (seed 0) on each edit type of the \emph{Seen} and \emph{Unseen} sets (\cref{sec:exp-setup}); percentages at the fixed threshold. Box P and R: precision and recall of the predicted boxes against the reference boxes at IoU 0.5.}
\label{tab:supp-per-type}
\end{table}

\Cref{tab:supp-per-type} gives VB's detection and localization on each edit type (\cref{sec:exp-grounding}).

Its detection depends on the edit type: TPR is 70.0 for added objects and for scene backgrounds, but 38.7 for removed objects, 47.3 for translated text and 48.0 for changed object geometry, and removed objects also have the lowest pixel IoU (5.5), so the edits that leave no new content behind are the hardest.

VB's boxes are precise but few: on the \emph{Seen} set its box precision is 2--4 times its box recall in every edit type (\cref{tab:supp-per-type}), because the reference has one box per connected component of the mask and VB usually draws one. For 765 of the 1,171 \emph{Seen} edits it calls edited (65\%), that one box is the whole image, in every edit type (51--84\%) and even for 47\% of the edits whose reference covers under a tenth of the image (73\% of those over half); VB then says that something was edited but not where. Localization is best where new content fills a large region (scene backgrounds 60.3 box F1, replaced objects 35.4) and worst where content disappears or changes subtly (removed objects 9.2, translated text 17.2). Detection, in contrast, does not depend on the edit's visibility: VB detects the 253 edits annotated as leaving no visible evidence as often as the rest (TPR 58.5 against 60.3), and edits under 10\% of the image as often as those over 25\% (62.5--66.7 against 58.6), which is what a whole-image trace predicts (\cref{sec:exp-generalization}).

\section{Generalization and robustness}
\label{sec:supp-generalization}

\begin{table}[t]
\centering\footnotesize\setlength{\tabcolsep}{2pt}
\begin{tabular}{l cccc}
\toprule
\multicolumn{5}{l}{\textbf{(a)} MagicBrush dev: 528 DALL-E 2 edits + 263 photos} \\
\midrule
Method & AUC [95\% CI] & TPR & FPR & pIoU \\
\midrule
\multicolumn{5}{l}{\textit{IFDL methods}} \\
TruFor~\cite{Guillaro_2023_CVPR} & 60.3 {\scriptsize[57.0, 63.6]} & 14.0 & 5.3 & 7.4 \\
MVSS-Net~\cite{Chen2021MVSS} & 60.2 {\scriptsize[56.7, 63.8]} & 39.2 & 27.4 & 1.5 \\
PIM~\cite{kong2025pim} & 54.8 {\scriptsize[51.1, 58.1]} & 76.9 & 68.8 & 7.1 \\
HiFi-Net~\cite{Guo_2023_CVPR} & 55.7 {\scriptsize[52.8, 58.5]} & 5.9 & 6.5 & 0.3 \\
IML-ViT~\cite{ma2023imlvit} & 59.6 {\scriptsize[55.8, 63.4]} & 100.0 & 100.0 & 5.6 \\
\midrule
\multicolumn{5}{l}{\textit{Forensic MLLMs}} \\
SIDA-7B~\cite{Huang_2025_CVPR} & 59.7 {\scriptsize[56.9, 62.7]} & 35.8 & 23.6 & 5.1 \\
FakeShield$^\dagger$~\cite{xu2025fakeshield} & 50.5 {\scriptsize[49.9, 51.2]} & 99.4 & 98.5 & 11.4 \\
So-Fake-R1$^\dagger$~\cite{huang2026sofakebenchmarkingexplainingsocial} & 46.3 {\scriptsize[43.6, 49.0]} & 74.4 & 81.7 & 6.3 \\
\midrule
\multicolumn{5}{l}{\textit{Zero-shot MLLMs}} \\
Qwen2.5-VL-3B$^\ddagger$~\cite{bai2025qwen25vl} & 26.4 {\scriptsize[22.5, 30.6]} & 65.0 & 100.0 & 11.2 \\
Qwen2.5-VL-7B$^\ddagger$~\cite{bai2025qwen25vl} & 57.5 {\scriptsize[54.0, 61.3]} & 93.4 & 95.4 & 18.3 \\
Qwen2.5-VL-32B (4-bit)~\cite{bai2025qwen25vl} & 65.0 {\scriptsize[62.1, 67.9]} & 23.5 & 11.8 & 5.4 \\
Qwen3-VL-8B~\cite{bai2025qwen3vltechnicalreport} & 66.3 {\scriptsize[63.4, 69.2]} & 83.0 & 59.7 & \textbf{18.7} \\
InternVL3-8B~\cite{zhu2025internvl3} & \textbf{66.4 {\scriptsize[63.5, 69.4]}} & 48.3 & 27.4 & 10.7 \\
\midrule
\multicolumn{5}{l}{\textit{IFDL methods fine-tuned on our data}} \\
TruFor~\cite{Guillaro_2023_CVPR} & 47.4 {\scriptsize[44.6, 50.3]} & 4.4 & 5.7 & 0.5 \\
MVSS-Net~\cite{Chen2021MVSS} & 55.5 {\scriptsize[52.1, 58.7]} & 0.2 & 0.8 & 0.0 \\
PIM~\cite{kong2025pim} & 54.6 {\scriptsize[51.6, 57.7]} & 0.6 & 0.0 & 0.0 \\
\midrule
\multicolumn{5}{l}{\textit{Ours: Qwen2.5-VL-3B~\cite{bai2025qwen25vl} with LoRA~\cite{hu2022lora} (\cref{sec:models})}} \\
VB: verdict and boxes & 48.0 {\scriptsize[44.9, 51.1]} & 0.4 & 0.4 & 0.1 \\
SA: VB + type and intent & 48.0 {\scriptsize[44.9, 51.2]} & 0.4 & 0.4 & 0.1 \\
RU: RN on unverified claims & 52.9 {\scriptsize[49.2, 56.8]} & 28.4 & 26.6 & 6.7 \\
RN: RG, no inline boxes & 53.5 {\scriptsize[49.9, 57.1]} & 28.0 & 25.9 & 6.4 \\
RG: grounded reasoning & 52.6 {\scriptsize[48.9, 56.0]} & 25.9 & 25.5 & 6.6 \\
RG+RL: RG + GRPO & 54.7 {\scriptsize[51.3, 58.3]} & 46.2 & 41.4 & 11.2 \\
AF: RG, answer first & 54.9 {\scriptsize[52.0, 57.8]} & 3.8 & 0.4 & 1.4 \\
AF+W: AF, weighted loss & 50.9 {\scriptsize[47.8, 54.0]} & 0.4 & 0.4 & 0.1 \\
\midrule
\multicolumn{5}{l}{\textbf{(b)} Trace test: AUC against source S (630 pairs)} \\
\midrule
Detector & T & P & U & K \\
\midrule
TruFor~\cite{Guillaro_2023_CVPR}, fine-tuned & 96.4 & 71.9 & 82.4 & 49.0 \\
MVSS-Net~\cite{Chen2021MVSS}, fine-tuned & 100.0 & 92.1 & 98.5 & 50.7 \\
TruFor~\cite{Guillaro_2023_CVPR}, released & 48.6 & 50.4 & 52.1 & 49.7 \\
\bottomrule
\end{tabular}
\caption{\textbf{Another editor, and what the fine-tuned detectors detect.} (a) MagicBrush dev (\cref{sec:exp-generalization}); percentages, AUC with its 95\% bootstrap interval over source photos. $^\dagger$No probability output: AUC ranks by the verdict. $^\ddagger$A parse failure counts as a wrong verdict with the worst-case score. (b) Each version is scored against its source photo S, all five processed identically: T is the edit, P the source with the edited region taken from the edit, U the edit with that region put back from the source, K the source with that region from a copy of the source sent along the edit's processing (seam control). All 95\% intervals are within $\pm$1.6 points.}
\label{tab:cross-editor}
\end{table}

On the \emph{Unseen} set, every existing detector stays within 3 points of its AUC on the \emph{Seen} set (differences from $-2.8$ to $+1.1$; \cref{tab:e1-baselines}). None of them was trained on our edit types, so for them the distinction between seen and held-out types means little; the \emph{Unseen} set tests the models fine-tuned on our data. The two sets share their authentic photos, so these differences come from the edits alone. For FakeShield and So-Fake-R1, the \emph{Unseen} set's authentic half reuses their predictions on the \emph{Seen} set instead of a second run. The other methods were run on the \emph{Unseen} set in full: SIDA and the IFDL methods gave the same scores on the shared photos, and the zero-shot MLLMs the same verdict on 95--100\% of them.

Standardizing MagicBrush (\cref{sec:supp-magicbrush}) removes the differences in size and format, but not DALL-E 2's own trace: outside the edited region an edit is smoother than its own photo in 99.2\% of the 265 first-turn pairs, and the share of high-frequency energy alone separates MagicBrush's edits from its photos at 0.70 AUC (0.53 on our \emph{Seen} set); sending each photo along the edit's resampling path leaves this unchanged. Training on our data does not carry over to this editor (\cref{tab:cross-editor}a), so the detectors trained on it evidently do not pick up this trace either. Fine-tuned TruFor falls from 60.3 AUC with its released weights to 47.4, 12.9 points lower (paired bootstrap [8.5, 17.2]); fine-tuned MVSS-Net falls from 60.2 to 55.5, 4.8 points lower, within noise ([$-0.2$, 9.7]); fine-tuned PIM stays at 54.6 against 54.8 ([$-4.3$, 4.4]). At the fixed threshold all three call almost every MagicBrush edit authentic (TPR 4.4, 0.2 and 0.6). Our VB fails in the same way: it calls 788 of the 791 images authentic and reaches 48.0 AUC, against 93.1 on the \emph{Seen} set, and SA does the same (48.0). RU, RN, RG and RG+RL stay near chance here as well (52.9, 53.5, 52.6 and 54.7 AUC). The highest AUCs on MagicBrush come from released zero-shot MLLMs, InternVL3-8B 66.4, Qwen3-VL-8B 66.3 and Qwen2.5-VL-32B 65.0, all higher than on our sets (50--56); this suggests that DALL-E 2's edits show more visible artifacts than Nano-Banana's, not that these models detect edits in general.

\begin{table*}[t]
\centering\footnotesize\setlength{\tabcolsep}{4pt}
\begin{tabular}{l r cc ccc ccc}
\toprule
 & & \multicolumn{2}{c}{Ours} & \multicolumn{3}{c}{IFDL fine-tuned on our data} & \multicolumn{3}{c}{IFDL, released weights} \\
\cmidrule(lr){3-4}\cmidrule(lr){5-7}\cmidrule(lr){8-10}
Editor & Edits & VB & SA & TruFor & MVSS-Net & PIM & TruFor & MVSS-Net & PIM \\
\midrule
BAGEL & 271 & 45.1 & 45.6 & 32.9 & 58.2 & 48.5 & 58.2 & 57.7 & 47.1 \\
FLUX.2-klein-Base9B & 272 & 47.5 & 48.9 & 46.4 & 53.2 & 51.6 & 53.0 & 52.3 & 47.9 \\
GLM-Image & 272 & 57.1 & 58.4 & 45.9 & 50.4 & 62.6 & 46.3 & 46.8 & 45.5 \\
GPTImage-2-1.5 & 271 & 49.4 & 49.9 & 46.9 & 60.1 & 50.2 & 48.4 & 54.6 & 47.4 \\
GrokImagine & 272 & 51.7 & 54.1 & 45.1 & 57.2 & 57.1 & 49.1 & 53.7 & 55.4 \\
HunyuanImage3.0 & 272 & 58.1 & 58.3 & 48.3 & 59.1 & 59.3 & 44.4 & 52.9 & 53.0 \\
Jimeng4.6 & 271 & 53.6 & 53.5 & 43.7 & 56.2 & 52.7 & 50.1 & 50.4 & 46.2 \\
JoyAI-Image-Edit & 271 & 63.5 & 63.5 & 61.2 & 59.0 & 71.1 & 52.3 & 50.1 & 50.9 \\
KlingImage3.0 & 271 & 52.1 & 52.2 & 27.8 & 65.1 & 58.8 & 44.0 & 58.3 & 54.0 \\
LongCatImage & 271 & 54.8 & 55.8 & 59.7 & 59.8 & 69.5 & 51.7 & 53.9 & 44.3 \\
NanoBananaPro-Gemini3Pro & 275 & 58.0 & 59.3 & 50.9 & 84.6 & 55.9 & 51.0 & 50.5 & 49.9 \\
OmniGenv2 & 270 & 45.2 & 47.5 & 29.7 & 57.7 & 51.5 & 59.3 & 53.7 & 49.0 \\
QwenImageEditPlus & 273 & 63.5 & 64.3 & 50.3 & 59.4 & 71.9 & 48.9 & 47.1 & 56.5 \\
ReveV1 & 271 & 48.3 & 49.6 & 45.7 & 58.7 & 61.3 & 45.7 & 47.4 & 50.4 \\
Seedream4.5 & 273 & 55.8 & 57.9 & 31.8 & 58.1 & 48.9 & 50.1 & 54.1 & 50.9 \\
Step1XEditv1.2 & 272 & 55.1 & 55.4 & 58.8 & 56.8 & 72.5 & 50.3 & 51.1 & 48.2 \\
ViduQ2 & 273 & 56.3 & 56.6 & 30.5 & 58.7 & 53.6 & 47.8 & 52.2 & 52.4 \\
Wan2.6Image & 272 & 62.7 & 62.3 & 52.2 & 60.5 & 56.0 & 48.5 & 44.9 & 59.8 \\
\midrule
All editors & 4,893 & 54.4 & 55.2 & 44.8 & 59.6 & 58.4 & 50.0 & 51.7 & 50.5 \\
\quad 95\% CI & & {\scriptsize[53.7, 55.0]} & {\scriptsize[54.5, 55.9]} & {\scriptsize[44.0, 45.6]} & {\scriptsize[58.8, 60.4]} & {\scriptsize[57.6, 59.1]} & {\scriptsize[49.2, 50.8]} & {\scriptsize[51.1, 52.4]} & {\scriptsize[49.6, 51.4]} \\
\quad TPR / FPR & & 9.5 / 6.7 & 5.9 / 2.9 & 8.9 / 10.6 & 2.2 / 0.3 & 5.8 / 1.7 & 6.3 / 5.5 & 32.2 / 31.1 & 79.3 / 80.0 \\
\bottomrule
\end{tabular}
\caption{\textbf{Eighteen MLLM editors.} Image AUC on the natural edits of MLLM-Edit's public release~\cite{yu2026mllmedit} (\cref{sec:supp-mllm-edit}), each editor's edits against their own photos, editors as named in the release; percentages. Per editor, the 95\% bootstrap interval over photos spans 4.6--8.1 points. Last rows: all editors pooled, with its interval, and TPR / FPR at each method's fixed threshold.}
\label{tab:supp-mllm-edit}
\end{table*}

\Cref{tab:supp-mllm-edit} scores the methods trained on our data and the released IFDL detectors on the 18 editors of MLLM-Edit. Over all editors, the methods trained on our data reach 44.8 (fine-tuned TruFor) to 59.6 AUC (fine-tuned MVSS-Net) and the released detectors 50.0--51.7, so no detector we run on this set reaches 60 AUC over all editors, whether trained on our data or not. At the fixed threshold, the methods trained on our data flag 2--10\% of the edits (TPR 2.2--9.5), and the released detectors flag about as many photos as edits (TPR 6.3, 32.2 and 79.3 at FPR 5.5, 31.1 and 80.0). Fine-tuned TruFor ranks the photos above the edits for 12 of the 18 editors (down to 27.8 AUC on KlingImage3.0). The best scores of the methods trained on our data come from fine-tuned PIM, 69.5--72.5 AUC on LongCatImage, JoyAI-Image-Edit, QwenImageEditPlus and Step1XEditv1.2. On QwenImageEditPlus, VB, SA and fine-tuned PIM come close to their AUCs on our Qwen-Image-Edit edits (63.5, 64.3 and 71.9 against 66.9, 69.3 and 70.3; \cref{tab:second-editor}), although these photos are smaller (\cref{sec:supp-mllm-edit}). The one exception is fine-tuned MVSS-Net on Nano Banana Pro, 84.6 AUC [81.8, 87.2], where the other methods trained on our data reach 50.9--59.3; even there it flags only 18.2\% of the edits at its threshold, with no false positive (\cref{sec:supp-limitations}). Without released masks we cannot test whether any method finds the edited region on this set.

\subsection{Trace test}
\label{sec:supp-trace}

\begin{table*}[t]
\centering\footnotesize\setlength{\tabcolsep}{4pt}
\begin{tabular}{l l cccc}
\toprule
Detector & Test set & T & P & U & K \\
\midrule
TruFor~\cite{Guillaro_2023_CVPR}, fine-tuned & Seen (480) & 96.7 {\scriptsize[95.8, 97.6]} & 72.2 {\scriptsize[70.5, 74.2]} & 83.1 {\scriptsize[81.4, 84.9]} & 48.9 {\scriptsize[48.5, 49.3]} \\
 & Unseen (150) & 95.0 {\scriptsize[92.8, 97.1]} & 71.0 {\scriptsize[68.2, 74.5]} & 80.0 {\scriptsize[76.8, 83.5]} & 49.2 {\scriptsize[48.3, 49.9]} \\
\midrule
MVSS-Net~\cite{Chen2021MVSS}, fine-tuned & Seen (480) & 100.0 {\scriptsize[100.0, 100.0]} & 92.8 {\scriptsize[91.2, 94.3]} & 98.6 {\scriptsize[97.9, 99.2]} & 50.5 {\scriptsize[49.7, 51.3]} \\
 & Unseen (150) & 100.0 {\scriptsize[99.9, 100.0]} & 89.7 {\scriptsize[86.3, 92.8]} & 98.1 {\scriptsize[96.6, 99.3]} & 51.4 {\scriptsize[50.1, 52.8]} \\
\midrule
TruFor~\cite{Guillaro_2023_CVPR}, released & Seen (480) & 48.7 {\scriptsize[47.2, 50.2]} & 50.1 {\scriptsize[49.0, 51.3]} & 52.0 {\scriptsize[51.0, 53.2]} & 49.5 {\scriptsize[48.9, 50.1]} \\
 & Unseen (150) & 48.1 {\scriptsize[45.2, 50.8]} & 51.5 {\scriptsize[49.3, 53.9]} & 52.2 {\scriptsize[50.3, 54.1]} & 50.6 {\scriptsize[49.0, 52.3]} \\
\bottomrule
\end{tabular}
\caption{\textbf{Trace test by test set.} Image AUC of each version against its source photo S, with the 95\% interval from 2,000 bootstrap resamples of pairs; the number of pairs is in parentheses. Versions as in \cref{sec:exp-generalization} and \cref{sec:supp-trace}.}
\label{tab:supp-trace}
\end{table*}

\begin{table}[t]
\centering\footnotesize\setlength{\tabcolsep}{3pt}
\resizebox{\linewidth}{!}{%
\begin{tabular}{l cccc}
\toprule
 & In & \multicolumn{3}{c}{Outside the mask} \\
\cmidrule(lr){3-5}
 & mask & Changed & \multicolumn{2}{c}{Unchanged: near, far} \\
\midrule
Share of pixels & 20.9 & 2.3 & 16.2 & 60.5 \\
\midrule
TruFor~\cite{Guillaro_2023_CVPR}, fine-tuned & 56.0 & 42.7 & 34.4 & 17.0 \\
MVSS-Net~\cite{Chen2021MVSS}, fine-tuned & 56.2 & 35.4 & 31.6 & 13.9 \\
TruFor~\cite{Guillaro_2023_CVPR}, released & 8.1 & 9.7 & 6.9 & 7.4 \\
\bottomrule
\end{tabular}}
\caption{\textbf{Where the detectors flag pixels.} Percentage of pixels flagged ($p \geq 0.5$) in each group, averaged over the 1,653 \emph{Seen} edits that are registered to their source; first row: each group's share of all pixels. Changed: CIE76 difference to the warped source above 10. Near: within 2\% of the image diagonal of the mask.}
\label{tab:supp-flags}
\end{table}

We take 630 edits from the two test sets (40 per \emph{Seen} type except scene background, 50 per \emph{Unseen} type) and make five versions of each, processed identically (one resampling onto a common grid, the same JPEG compression). \emph{S} is the source photo and \emph{T} the edit. With $C$ the reference mask plus every pixel that changed visibly, \emph{P} is the source with only $C$ taken from the edit, \ie, the edited patch on an otherwise untouched photo; \emph{U} is the edit with $C$ put back from the source, \ie, the editor's output with the edit undone; \emph{K} is the source with $C$ taken from a copy of the source that went through the edit's resampling and compression, a control for the seam. We score each version against S (\cref{tab:cross-editor}b). With identical processing, fine-tuned TruFor still separates T from S at 96.4 AUC (99.1 on the test sets) and VB at 88.0 (93.1; 78.8 on the second editor's Nano-Banana edits), so processing differences explain little of TruFor's score but may explain some of VB's. Both fine-tuned detectors find U more easily than P (TruFor 82.4 \vs 71.9, MVSS-Net 98.5 \vs 92.1) and far more easily than the seam control K, while released TruFor stays at chance on every version. Our VB behaves the same way, if less strongly: it separates T from S at 88.0 AUC and finds U (69.9) more easily than P (63.8), both above K (53.2). The fine-tuned detectors, and to a lesser degree VB, thus respond mainly to a trace the editor leaves over the whole image, not to the edited content, which is consistent with their failure on MagicBrush. Training VB against this trace does not remove it (\cref{sec:exp-ablations}).

\paragraph{Pairs.} We take edits from both test sets whose reference mask was computed against the registered source photo, with a registration rotation below $0.5^\circ$, and leave out scene background, whose edited region covers most of the image. After a shuffle with a fixed seed we keep 40 pairs per \emph{Seen} type and 50 per \emph{Unseen} type, skipping a pair whose region $C$ (below) covers more than 60\% of the image; 630 of the 849 pairs built are used.

\paragraph{Processing.} Both images of a pair go through the same steps. The edit is first compressed once as JPEG at the source photo's own quality, estimated from its quantization table, so that both carry one earlier compression. Both are then cropped to the area they share, using the affine map that registered them, and resampled once with a Lanczos filter onto the same grid, at a scale drawn from $U(0.75, 0.95)$ of the source's size, so both are reduced. All five versions are finally saved as JPEG at a quality drawn from $\{75, \dots, 95\}$; with probability 0.3 they are compressed twice, at two such qualities.

\paragraph{Versions.} $C$ is the reference mask, dilated by 1\% of the image diagonal, together with the pixels where edit and source differ visibly: a CIE76 color difference above 10 after a Gaussian blur of 1.5 pixels, cleaned by a morphological opening, keeping components of at least 0.05\% of the image, and dilated by 3 pixels. The versions are blended with $C$ blurred by 2 pixels: P takes $C$ from the edit into the source, U takes $C$ from the source into the edit, and K takes $C$ into the source from a copy of the source that went through the edit's path (warped onto the edit's grid, compressed at the same quality, resampled like the edit). We score all 3,150 images as in \cref{sec:exp-setup} and compute the AUC of each version against S, each pair contributing one image of each. \Cref{tab:supp-trace} gives the results per test set; the \emph{Seen} and \emph{Unseen} pairs agree.

\paragraph{Where the detectors flag pixels.} If the reference masks missed part of an edit, a detector would rightly flag pixels outside them. We test this on the saved maps of the 1,653 \emph{Seen} edits whose source is registered: we warp the source onto the processed edit and split the pixels outside the mask into those that changed visibly (CIE76 above 10) and those that did not, near the mask (within 2\% of the diagonal) or far from it. Only 2.3\% of all pixels lie outside the mask and changed, so the masks miss little. Yet the fine-tuned detectors flag 17.0\% (TruFor) and 13.9\% (MVSS-Net) of the unchanged pixels far from the mask, against 7.4\% for released TruFor, which flags every group at about the same rate (\cref{tab:supp-flags}). The fine-tuned detectors thus flag pixels the edit did not touch, as they would if they had learned a whole-image trace.

\subsection{Robustness to processing}
\label{sec:supp-robust}

We process every image of the \emph{Seen} set in five ways and score VB and the three fine-tuned IFDL methods on each version as in \cref{sec:exp-setup} (\cref{tab:supp-robust}). JPEG 75 and JPEG 50 re-encode the image at that quality. Down 75 and down 50 reduce each side to 75\% or 50\% with a Lanczos filter and keep the image's own JPEG tables, so only the resampling changes. Crop 80 keeps 80\% of each side, also with the image's own tables; for an edit the window is placed where it keeps the most edited pixels, for a photo at random. An edit wider than the window cannot be kept whole, so the crop keeps a median 88.8\% of an edit's pixels (at least 48.7\%). The reference masks are resized or cropped with the image.

Under every condition except half size, VB loses more AUC than the fine-tuned detectors: 5.1 points at JPEG 75, 12.8 at JPEG 50, 6.7 at down 75 and 8.8 at crop 80, against at most 6.7 for them. Halving each side costs every model except MVSS-Net the most: VB falls from 93.1 to 75.5 AUC and calls only 9.9\% of the edits edited (60.1\% on the original images), fine-tuned TruFor falls to 72.6 and PIM to 82.6. Fine-tuned MVSS-Net keeps 100.0 AUC under every condition.

\begin{table*}[t]
\centering\footnotesize\setlength{\tabcolsep}{4pt}
\resizebox{\linewidth}{!}{%
\begin{tabular}{l cccccc}
\toprule
Model & Original & JPEG 75 & JPEG 50 & Down 75 & Down 50 & Crop 80 \\
\midrule
\multicolumn{7}{l}{\textit{Image AUC [95\% CI]}} \\
VB & 93.1 {\scriptsize[92.4, 93.8]} & 88.0 {\scriptsize[87.1, 88.9]} & 80.3 {\scriptsize[79.2, 81.3]} & 86.4 {\scriptsize[85.5, 87.4]} & 75.5 {\scriptsize[74.3, 76.7]} & 84.3 {\scriptsize[83.3, 85.3]} \\
TruFor~\cite{Guillaro_2023_CVPR}, fine-tuned & 99.1 {\scriptsize[98.9, 99.3]} & 98.0 {\scriptsize[97.7, 98.3]} & 95.9 {\scriptsize[95.3, 96.4]} & 94.9 {\scriptsize[94.2, 95.5]} & 72.6 {\scriptsize[71.5, 73.6]} & 98.1 {\scriptsize[97.7, 98.4]} \\
MVSS-Net~\cite{Chen2021MVSS}, fine-tuned & 100.0 {\scriptsize[100.0, 100.0]} & 100.0 {\scriptsize[100.0, 100.0]} & 100.0 {\scriptsize[100.0, 100.0]} & 100.0 {\scriptsize[100.0, 100.0]} & 100.0 {\scriptsize[100.0, 100.0]} & 100.0 {\scriptsize[99.9, 100.0]} \\
PIM~\cite{kong2025pim}, fine-tuned & 98.7 {\scriptsize[98.4, 99.0]} & 97.9 {\scriptsize[97.5, 98.3]} & 92.0 {\scriptsize[91.3, 92.8]} & 95.8 {\scriptsize[95.3, 96.4]} & 82.6 {\scriptsize[81.5, 83.8]} & 94.1 {\scriptsize[93.4, 94.8]} \\
\midrule
\multicolumn{7}{l}{\textit{TPR / FPR at the fixed threshold}} \\
VB & 60.1 / 3.9 & 57.3 / 7.2 & 36.1 / 6.2 & 55.8 / 8.1 & 9.9 / 0.7 & 53.3 / 9.3 \\
\bottomrule
\end{tabular}}
\caption{\textbf{Robustness to processing.} The \emph{Seen} set (1,950 edits and 1,950 photos) as it is (Original) and after JPEG re-encoding at quality 75 or 50, downscaling each side to 75\% or 50\%, or cropping 80\% of each side (\cref{sec:supp-robust}); each version is scored on its own. Percentages; AUC with its 95\% bootstrap interval over source photos. The Original column repeats \cref{tab:e1-baselines}.}
\label{tab:supp-robust}
\end{table*}

\subsection{Whole-image tiers}
\label{sec:supp-gsx}

The global (G), stylized (S) and geometric (X) tiers of \cref{sec:data-source} change the whole image, so their edits have no region to localize. To test whether the detectors of \cref{tab:e1-baselines} still find such edits, we draw 50 test edits of each of the 18 trainable types of these tiers by the quota rule of \cref{sec:supp-quota} (G 250, S 550, X 100) and 50 edits of the held-out global type weather, and pair them with 900 photos, 896 of them source photos of the 900 trainable-type edits and 4 other untouched photos. The images come from the same processed release as the other test sets, and the edits are scored at the image level only. Of the 900 edits, 308 (34\%) are annotated as leaving no visible evidence, against 253 of the 1,950 \emph{Seen} edits (13\%).

With their released weights the IFDL methods stay between 42.2 and 56.8 AUC, near chance as on the \emph{Seen} set (\cref{tab:supp-gsx}). The models trained on our local and regional edits detect the whole-image edits nearly as well as those: VB reaches 90.0 AUC (93.1 on \emph{Seen}) and 86.4 on weather, and the fine-tuned IFDL methods 95.4 to 100.0. This is what the editor's trace of \cref{sec:exp-generalization} predicts: the edits come from the same editor, which regenerates the whole image whatever the instruction, so a trace learned on local edits is also present in whole-image edits. VB is weakest on the global tier (84.9 AUC), whose edits change tone, grain, season or lighting, and strongest on the stylized tier (92.4), and it detects the edits annotated as leaving no visible evidence about as often as the others (TPR 50.3 against 53.0), as on the \emph{Seen} set. Of the forensic MLLMs, SIDA reaches 71.3 AUC, above its 58.0 on \emph{Seen}, mostly on the stylized tier (79.9).

FakeShield stays at chance (47.6 AUC), while So-Fake-R1 (60.2), the zero-shot MLLMs other than Qwen2.5-VL-3B (69.8 to 77.9) and our reasoning variants RG and RG+RL (64.8 and 63.3) score above their 49.1 to 57.1 on \emph{Seen}. Like SIDA, they gain on the stylized tier (67.1 to 90.9) but stay between 50.0 and 62.5 on the global tier, between 34.5 and 52.4 on the geometric tier and between 40.9 and 57.2 on weather. A stylized edit re-renders the photo, or a person in it, as a painting, a cartoon, a toy figure or a sketch, which a model can recognize from what the image shows, without the editor's trace that VB and the fine-tuned IFDL methods detect on every tier.

Trained on all five tiers, VB detects the whole-image edits better still. This variant keeps VB's 27,846 training images but spreads its 13,923 edits evenly over all 31 trainable types, about 460 per type, so it sees 5,973 local and regional edits instead of 13,923. It reaches 96.0 AUC on the 18 types, 93.7 on the global tier and 94.6 on weather, a type it was not trained on either (last row of \cref{tab:supp-gsx}). On the other test sets it loses 1.1 to 3.9 points: 90.9 AUC on \emph{Seen}, 90.2 on \emph{Unseen}, 44.1 on MagicBrush, and 77.0 and 65.1 on the Nano-Banana and Qwen edits of the second-editor set. VB-full (\cref{sec:supp-ablations}), which sees five times as many local and regional edits as VB and none of these tiers, reaches 92.7 AUC on the 18 types and 93.7 on weather.

\begin{table*}[t]
\centering\footnotesize\setlength{\tabcolsep}{3.7pt}
\resizebox{\ifdim\width>\textwidth\textwidth\else\width\fi}{!}{%
\begin{tabular}{l cccccccc c}
\toprule
 & \multicolumn{8}{c}{18 edit types: 900 edits (G 250, S 550, X 100) + 900 photos} & Held out: weather (50 + 900) \\
\cmidrule(lr){2-9}\cmidrule(lr){10-10}
Method & AUC [95\% CI] & G & S & X & TPR & FPR & No ev. & Ev. & AUC [95\% CI] \\
\midrule
\multicolumn{10}{l}{\textit{IFDL methods}} \\
TruFor~\cite{Guillaro_2023_CVPR} & 42.2 {\scriptsize[39.8, 44.7]} & 45.5 & 41.1 & 40.5 & 11.3 & 16.3 & 8.8 & 12.7 & 43.5 {\scriptsize[35.4, 52.0]} \\
MVSS-Net~\cite{Chen2021MVSS} & 47.6 {\scriptsize[45.4, 49.8]} & 48.9 & 48.7 & 38.5 & 70.3 & 70.9 & 68.5 & 71.3 & 43.1 {\scriptsize[35.1, 51.0]} \\
PIM~\cite{kong2025pim} & 56.8 {\scriptsize[54.7, 58.8]} & 54.7 & 58.1 & 55.0 & 92.9 & 89.8 & 92.5 & 93.1 & 60.2 {\scriptsize[51.9, 68.2]} \\
HiFi-Net~\cite{Guo_2023_CVPR} & 53.1 {\scriptsize[51.1, 54.9]} & 46.4 & 55.7 & 55.1 & 3.9 & 3.2 & 3.6 & 4.1 & 46.0 {\scriptsize[37.3, 54.1]} \\
IML-ViT~\cite{ma2023imlvit} & 51.7 {\scriptsize[49.4, 54.0]} & 42.0 & 56.4 & 49.7 & 99.4 & 99.6 & 99.4 & 99.5 & 50.2 {\scriptsize[42.4, 57.3]} \\
\midrule
\multicolumn{10}{l}{\textit{Forensic MLLMs}} \\
SIDA-7B~\cite{Huang_2025_CVPR} & 71.3 {\scriptsize[69.4, 73.1]} & 58.0 & 79.9 & 57.3 & 9.3 & 1.4 & 6.8 & 10.6 & 62.0 {\scriptsize[53.9, 69.9]} \\
FakeShield$^\dagger$~\cite{xu2025fakeshield} & 47.6 {\scriptsize[46.0, 49.2]} & 43.8 & 49.9 & 44.3 & 15.6 & 20.3 & 12.3 & 17.2 & 42.8 {\scriptsize[39.6, 46.7]} \\
So-Fake-R1$^\dagger$~\cite{huang2026sofakebenchmarkingexplainingsocial} & 60.2 {\scriptsize[58.3, 62.1]} & 50.7 & 67.1 & 46.0 & 47.4 & 27.0 & 41.2 & 50.7 & 48.5 {\scriptsize[42.2, 55.2]} \\
\midrule
\multicolumn{10}{l}{\textit{Zero-shot MLLMs}} \\
Qwen2.5-VL-3B$^\ddagger$~\cite{bai2025qwen25vl} & 8.9 {\scriptsize[7.6, 10.2]} & 5.0 & 11.4 & 4.9 & 35.0 & 100.0 & 35.1 & 35.0 & 4.8 {\scriptsize[2.4, 7.7]} \\
Qwen2.5-VL-7B$^\ddagger$~\cite{bai2025qwen25vl} & 69.8 {\scriptsize[67.8, 71.9]} & 53.3 & 83.7 & 34.5 & 93.4 & 91.6 & 92.2 & 94.1 & 40.9 {\scriptsize[33.0, 49.0]} \\
Qwen2.5-VL-32B (4-bit)~\cite{bai2025qwen25vl} & 76.2 {\scriptsize[74.5, 78.1]} & 60.7 & 89.6 & 41.9 & 53.0 & 8.6 & 40.3 & 59.6 & 49.2 {\scriptsize[39.2, 59.1]} \\
Qwen3-VL-8B~\cite{bai2025qwen3vltechnicalreport} & 77.9 {\scriptsize[76.0, 79.6]} & 62.5 & 90.9 & 44.2 & 76.7 & 38.7 & 67.9 & 81.2 & 55.3 {\scriptsize[46.5, 64.4]} \\
InternVL3-8B~\cite{zhu2025internvl3} & 76.0 {\scriptsize[74.3, 77.7]} & 58.4 & 88.4 & 52.4 & 56.6 & 15.2 & 45.1 & 62.5 & 57.2 {\scriptsize[49.5, 64.6]} \\
\midrule
\multicolumn{10}{l}{\textit{IFDL methods fine-tuned on our data (L and R)}} \\
TruFor~\cite{Guillaro_2023_CVPR} & 96.7 {\scriptsize[96.0, 97.3]} & 99.2 & 95.1 & 99.4 & 75.3 & 1.9 & 77.3 & 74.3 & 99.5 {\scriptsize[98.9, 99.8]} \\
MVSS-Net~\cite{Chen2021MVSS} & 100.0 {\scriptsize[100.0, 100.0]} & 100.0 & 100.0 & 100.0 & 99.9 & 0.1 & 100.0 & 99.8 & 100.0 {\scriptsize[100.0, 100.0]} \\
PIM~\cite{kong2025pim} & 95.4 {\scriptsize[94.5, 96.2]} & 98.3 & 93.5 & 98.8 & 77.9 & 3.3 & 79.5 & 77.0 & 98.4 {\scriptsize[96.8, 99.5]} \\
\midrule
\multicolumn{10}{l}{\textit{Ours, trained on L and R: Qwen2.5-VL-3B~\cite{bai2025qwen25vl} with LoRA~\cite{hu2022lora}}} \\
VB: verdict and boxes & 90.0 {\scriptsize[88.7, 91.3]} & 84.9 & 92.4 & 89.6 & 52.1 & 3.3 & 50.3 & 53.0 & 86.4 {\scriptsize[80.9, 91.4]} \\
SA: VB + type and intent & 89.4 {\scriptsize[88.0, 90.6]} & 83.5 & 92.3 & 87.9 & 38.4 & 1.7 & 38.6 & 38.3 & 85.6 {\scriptsize[80.1, 90.5]} \\
RG: grounded reasoning & 64.8 {\scriptsize[62.5, 66.9]} & 50.0 & 75.3 & 44.1 & 23.0 & 16.2 & 19.5 & 24.8 & 57.0 {\scriptsize[47.7, 65.6]} \\
RG+RL: RG + GRPO & 63.3 {\scriptsize[61.2, 65.4]} & 51.6 & 71.2 & 49.2 & 50.7 & 37.8 & 45.5 & 53.4 & 52.5 {\scriptsize[43.7, 60.7]} \\
VB, all L and R edits & 92.7 {\scriptsize[91.6, 93.7]} & 92.6 & 92.4 & 94.6 & 53.0 & 3.3 & 54.2 & 52.4 & 93.7 {\scriptsize[90.1, 96.6]} \\
\midrule
\multicolumn{10}{l}{\textit{Ours, trained on all five tiers}} \\
VB on all tiers & 96.0 {\scriptsize[95.3, 96.7]} & 93.7 & 98.1 & 90.5 & 79.4 & 3.4 & 72.1 & 83.3 & 94.6 {\scriptsize[91.7, 97.0]} \\
\bottomrule
\end{tabular}}
\caption{\textbf{Detectors on whole-image edits.} Test edits of the global (G), stylized (S) and geometric (X) tiers, 50 of each trainable type, and of the held-out global type weather, against the same photos (\cref{sec:supp-gsx}). Only the last row was trained on these tiers. Percentages; AUC with its 95\% bootstrap interval over source photos; G, S and X give the AUC of that tier's edits against all photos. TPR and FPR at the fixed threshold; No ev.\ and Ev.\ give the TPR on the 308 edits annotated as leaving no visible evidence and on the others. $^\dagger$No probability output: AUC ranks by the verdict. $^\ddagger$A parse failure counts as a wrong verdict with the worst-case score.}
\label{tab:supp-gsx}
\end{table*}

\subsection{Trace test under processing}
\label{sec:supp-trace-wm}

Nano-Banana's outputs carry an invisible SynthID watermark (\cref{sec:supp-limitations}), and such a watermark is built to be robust to common image perturbations~\cite{gowal2025synthidimage}. To see whether the trace of \cref{sec:supp-trace} is as robust, we process the versions S, T, P and U of all 630 pairs in five ways and score each version against S again (\cref{tab:supp-wm}): JPEG re-encoding at quality 50; reducing each side to 50\% or to 25\% with a Lanczos filter; a Gaussian blur of radius 2 pixels; and a crop of 50\% of each side at a random position. All versions of a pair receive the same operation, with the same window for the crop, and all but the JPEG re-encoding keep the image's own JPEG tables.

The trace that fine-tuned TruFor and VB find fades with the processing. TruFor separates U from S at 82.4 AUC without it, at 73.1 after JPEG 50, 61.4 after the blur, 57.3 at half size and 50.9 at a quarter; VB falls from 69.9 to 59.1, 55.2, 57.6 and 51.4. Fine-tuned MVSS-Net keeps the trace: U stays at 97.7 to 98.5 after JPEG 50, the blur and half size and at 91.4 at a quarter, and only the crop lowers it, to 65.6, together with its score on the edit itself (T, 73.7). A cue that fades under perturbations of the kind a watermark is built to resist is not read the way the watermark's decoder reads it, so TruFor and VB respond to a fine-grained property of the editor's output that such processing removes; MVSS-Net's cue survives it, as a watermark would, and fits either explanation. We can neither decode the watermark nor check that it survives these operations at these strengths, so the test narrows the question and does not settle it.

\section{Grounding tests}
\label{sec:supp-grounding}

\subsection{Region-masking test}
\label{sec:supp-masking}

\begin{table}[t]
\centering\footnotesize\setlength{\tabcolsep}{3pt}
\resizebox{\linewidth}{!}{%
\begin{tabular}{l l r cccc l}
\toprule
 & & & \multicolumn{4}{c}{Verdict flips to authentic (\%) after removing} & \\
\cmidrule(lr){4-7}
Model & Images & $n$ & nothing & random & cited & reference & cited $-$ random \\
\midrule
RG & edits & 252 & 16.7 & 36.9 & 76.6 & 68.6 & $+$39.7 [31.9, 47.2] \\
RG & photos & 145 & 13.1 & 36.6 & 67.6 & -- & $+$31.0 [20.0, 42.1] \\
RG+RL & edits & 347 & 9.5 & 25.1 & 44.4 & 52.8 & $+$19.3 [12.7, 25.9] \\
RG+RL & photos & 258 & 15.1 & 27.9 & 55.4 & -- & $+$27.5 [19.8, 35.3] \\
VB & edits & 307 & 1.3 & 23.1 & 58.6 & 54.7 & $+$35.5 [29.4, 41.3] \\
VB & photos & 14 & 0.0 & 42.9 & 85.7 & -- & $+$42.9 [14.3, 64.3] \\
\bottomrule
\end{tabular}}
\caption{Region-masking test. For every \emph{Seen} image a model calls edited (and whose cited region covers at most half of the image), one region is removed by Telea inpainting and the image re-encoded with its own JPEG tables: nothing (re-encoding only), a random region of the same shape, the region the model itself cited (RG, RG+RL: the boxes in its reasoning; VB: its predicted boxes) or the edit's reference region. Cells are the share of verdicts that change to authentic; the last column is the paired difference between the cited and the random region with a 95\% bootstrap interval over source photos.}
\label{tab:faithfulness}
\end{table}

\begin{table*}[t]
\centering\footnotesize\setlength{\tabcolsep}{5pt}
\begin{tabular}{l ccc c}
\toprule
 & \multicolumn{3}{c}{480 \emph{Seen} pairs} & All 630 pairs \\
\cmidrule(lr){2-4}\cmidrule(lr){5-5}
Method & As in the \emph{Seen} set & Identical processing & Change & Identical processing \\
\midrule
\multicolumn{5}{l}{\textit{IFDL methods}} \\
TruFor~\cite{Guillaro_2023_CVPR} & 44.8 {\scriptsize[41.7, 47.9]} & 48.7 {\scriptsize[47.1, 50.2]} & $+3.9$ & 48.6 {\scriptsize[47.2, 49.9]} \\
MVSS-Net~\cite{Chen2021MVSS} & 40.4 {\scriptsize[37.6, 43.0]} & 39.9 {\scriptsize[37.7, 42.3]} & $-0.5$ & 39.9 {\scriptsize[37.9, 41.9]} \\
PIM~\cite{kong2025pim} & 58.8 {\scriptsize[56.2, 61.6]} & 53.1 {\scriptsize[50.8, 55.5]} & $-5.7$ & 52.5 {\scriptsize[50.6, 54.3]} \\
HiFi-Net~\cite{Guo_2023_CVPR} & 52.2 {\scriptsize[50.1, 54.3]} & 52.3 {\scriptsize[50.6, 54.0]} & $+0.1$ & 52.2 {\scriptsize[50.8, 53.5]} \\
IML-ViT~\cite{ma2023imlvit} & 47.8 {\scriptsize[45.0, 50.8]} & 44.0 {\scriptsize[41.6, 46.4]} & $-3.8$ & 44.0 {\scriptsize[42.0, 46.1]} \\
\midrule
\multicolumn{5}{l}{\textit{Forensic MLLMs}} \\
SIDA-7B~\cite{Huang_2025_CVPR} & 58.1 {\scriptsize[56.2, 59.9]} & 56.0 {\scriptsize[54.6, 57.5]} & $-2.1$ & 55.1 {\scriptsize[53.8, 56.5]} \\
FakeShield$^\dagger$~\cite{xu2025fakeshield} & 50.7 {\scriptsize[48.4, 53.0]} & 52.7 {\scriptsize[51.4, 54.2]} & $+2.0$ & 52.6 {\scriptsize[51.3, 54.0]} \\
So-Fake-R1$^\dagger$~\cite{huang2026sofakebenchmarkingexplainingsocial} & 51.4 {\scriptsize[49.2, 53.5]} & 50.6 {\scriptsize[49.0, 52.4]} & $-0.8$ & 51.0 {\scriptsize[49.4, 52.5]} \\
\midrule
\multicolumn{5}{l}{\textit{Zero-shot MLLMs}} \\
Qwen2.5-VL-3B$^\ddagger$~\cite{bai2025qwen25vl} & 6.0 {\scriptsize[4.8, 7.4]} & 7.5 {\scriptsize[6.0, 9.1]} & $+1.5$ & 7.7 {\scriptsize[6.4, 9.1]} \\
Qwen2.5-VL-7B$^\ddagger$~\cite{bai2025qwen25vl} & 48.5 {\scriptsize[46.4, 50.4]} & 49.0 {\scriptsize[47.0, 51.1]} & $+0.5$ & 49.0 {\scriptsize[47.3, 50.5]} \\
Qwen2.5-VL-32B (4-bit)~\cite{bai2025qwen25vl} & 51.6 {\scriptsize[49.9, 53.4]} & 52.6 {\scriptsize[51.1, 54.1]} & $+1.0$ & 51.8 {\scriptsize[50.6, 53.1]} \\
Qwen3-VL-8B~\cite{bai2025qwen3vltechnicalreport} & 54.4 {\scriptsize[52.9, 56.0]} & 55.3 {\scriptsize[53.7, 57.0]} & $+0.9$ & 54.4 {\scriptsize[53.1, 55.7]} \\
InternVL3-8B~\cite{zhu2025internvl3} & 53.4 {\scriptsize[51.8, 54.9]} & 52.6 {\scriptsize[51.4, 53.9]} & $-0.8$ & 52.2 {\scriptsize[51.1, 53.3]} \\
\midrule
\multicolumn{5}{l}{\textit{IFDL methods fine-tuned on our data}} \\
TruFor~\cite{Guillaro_2023_CVPR} & 99.3 {\scriptsize[98.8, 99.6]} & 96.7 {\scriptsize[95.8, 97.6]} & $-2.6$ & 96.4 {\scriptsize[95.5, 97.2]} \\
MVSS-Net~\cite{Chen2021MVSS} & 100.0 {\scriptsize[100.0, 100.0]} & 100.0 {\scriptsize[100.0, 100.0]} & $0.0$ & 100.0 {\scriptsize[100.0, 100.0]} \\
PIM~\cite{kong2025pim} & 98.4 {\scriptsize[97.6, 99.1]} & 97.2 {\scriptsize[96.1, 98.1]} & $-1.2$ & 97.1 {\scriptsize[96.1, 97.9]} \\
\midrule
\multicolumn{5}{l}{\textit{Ours (\cref{sec:models,sec:supp-ablations})}} \\
VB: verdict and boxes & 92.9 {\scriptsize[91.4, 94.2]} & 88.2 {\scriptsize[86.4, 89.9]} & $-4.7$ & 88.0 {\scriptsize[86.5, 89.5]} \\
SA: VB + type and intent & 92.2 {\scriptsize[90.7, 93.6]} & 88.2 {\scriptsize[86.4, 89.9]} & $-4.0$ & 88.0 {\scriptsize[86.6, 89.5]} \\
VO: verdict only & 93.1 {\scriptsize[91.7, 94.3]} & 89.1 {\scriptsize[87.4, 90.7]} & $-4.0$ & 88.9 {\scriptsize[87.5, 90.4]} \\
RU: RN on unverified claims & 54.6 {\scriptsize[51.9, 57.4]} & 53.7 {\scriptsize[50.9, 56.5]} & $-0.9$ & 53.3 {\scriptsize[50.9, 55.8]} \\
RN: RG, no inline boxes & 58.6 {\scriptsize[55.9, 61.5]} & 54.4 {\scriptsize[51.9, 57.1]} & $-4.2$ & 54.1 {\scriptsize[51.9, 56.4]} \\
RG: grounded reasoning & 53.6 {\scriptsize[51.0, 56.4]} & 50.1 {\scriptsize[47.4, 52.8]} & $-3.5$ & 50.5 {\scriptsize[48.1, 52.9]} \\
RG+RL: RG + GRPO & 58.4 {\scriptsize[55.7, 61.2]} & 53.2 {\scriptsize[50.3, 55.8]} & $-5.2$ & 54.0 {\scriptsize[51.5, 56.5]} \\
AF: RG, answer first & 63.7 {\scriptsize[61.8, 65.7]} & 58.4 {\scriptsize[56.9, 60.0]} & $-5.3$ & 57.7 {\scriptsize[56.5, 58.9]} \\
AF+W: AF, weighted loss & 87.4 {\scriptsize[85.5, 89.1]} & 80.6 {\scriptsize[78.6, 82.7]} & $-6.8$ & 80.2 {\scriptsize[78.4, 82.0]} \\
\bottomrule
\end{tabular}
\caption{\textbf{Existing methods and ours under identical processing.} Image AUC with its 95\% bootstrap interval over source photos. The 480 \emph{Seen} pairs of the trace test (\cref{sec:supp-trace}) are scored twice: as the 480 edits and their 478 source photos appear in the \emph{Seen} set, and as versions T and S, processed identically; Change is the second AUC minus the first. The last column scores T against S on all 630 pairs. Marks as in \cref{tab:e1-baselines}.}
\label{tab:supp-matched}
\end{table*}

\begin{table*}[t]
\centering\footnotesize\setlength{\tabcolsep}{5pt}
\begin{tabular}{l ccc ccc ccc}
\toprule
 & \multicolumn{3}{c}{TruFor~\cite{Guillaro_2023_CVPR}, fine-tuned} & \multicolumn{3}{c}{MVSS-Net~\cite{Chen2021MVSS}, fine-tuned} & \multicolumn{3}{c}{VB} \\
\cmidrule(lr){2-4}\cmidrule(lr){5-7}\cmidrule(lr){8-10}
Processing & T & P & U & T & P & U & T & P & U \\
\midrule
None & 96.4 & 71.9 & 82.4 & 100.0 & 92.1 & 98.5 & 88.0 & 63.8 & 69.9 \\
\midrule
JPEG 50 & 87.5 & 66.5 & 73.1 & 100.0 & 89.4 & 97.8 & 69.8 & 60.5 & 59.1 \\
Down 50 & 64.6 & 54.9 & 57.3 & 100.0 & 88.8 & 97.7 & 65.3 & 56.9 & 57.6 \\
Down 25 & 53.6 & 51.6 & 50.9 & 99.4 & 75.4 & 91.4 & 53.9 & 52.3 & 51.4 \\
Blur & 71.3 & 57.5 & 61.4 & 100.0 & 86.3 & 98.5 & 60.6 & 55.5 & 55.2 \\
Crop 50 & 89.7 & 68.7 & 75.2 & 73.7 & 57.1 & 65.6 & 69.2 & 58.2 & 59.5 \\
\bottomrule
\end{tabular}
\caption{\textbf{Trace test under processing.} Image AUC of versions T, P and U against the source photo S on the 630 pairs of the trace test, without further processing (first row; \cref{sec:supp-trace}) and after one operation applied alike to all versions of a pair: JPEG re-encoding at quality 50, each side reduced to 50\% or to 25\%, a Gaussian blur of radius 2 pixels, or a crop of 50\% of each side (\cref{sec:supp-trace-wm}). All 95\% intervals are within $\pm$2.3 points.}
\label{tab:supp-wm}
\end{table*}

Does a verdict rest on the region the model cites? For every \emph{Seen} image a model calls edited, we remove one region by inpainting (OpenCV's Telea method, so no generative editor is involved), re-encode the image with its own JPEG tables and ask the model again (\cref{tab:faithfulness}). The region is the one the model cited (RG and RG+RL: the boxes in their reasoning; VB: its predicted boxes), the same shape moved to a random place, the edit's reference region, or nothing, which tests the re-encoding alone; images whose cited region covers more than half of the image are left out, which keeps 397 of RG's 651 positive calls, 605 of RG+RL's 1,596 and 321 of VB's 1,247. Re-encoding alone flips 16.7\% of RG's verdicts on edits and 9.5\% of RG+RL's, against 1.3\% of VB's, although it changes the pixels by 0.05 grey levels on average: the reasoning models' verdicts are brittle to noise far below what any edit leaves. Against that baseline, removing the cited region flips RG's verdict on 76.6\% of the edits, a random region of the same shape on 36.9\% (paired difference $+39.7$ [31.9, 47.2]) and the reference region on 68.6\%; for RG+RL the three figures are 44.4, 25.1 ($+19.3$ [12.7, 25.9]) and 52.8, for VB 58.6, 23.1 ($+35.5$ [29.4, 41.3]) and 54.7. The gap holds in every tercile of the cited area and, as a point estimate, in every edit type with 15 or more tested edits. The photos the models wrongly call edited behave the same way: removing the region RG cites flips 67.6\% of its false alarms, a random region 36.6\% ($+31.0$ [20.0, 42.1]); for RG+RL, 55.4 against 27.9 ($+27.5$ [19.8, 35.3]). The verdict thus follows the region the model points at, whether or not an edit is there: when the evidence is hallucinated, removing it removes the verdict with it.

\subsection{Rationale swap}
\label{sec:supp-swap}

Does the verdict also follow the reasoning? We give RG and RG+RL each \emph{Seen} image with a reasoning text in place of their own and read the verdict off as the probability of ``edited'' after it (\cref{sec:supp-swap}). With their own reasoning they reproduce their generated verdicts on 99.97\% and 93.7\% of the images. With the reasoning they wrote for an authentic photo, both call every image authentic; with the reasoning of an edit they called edited, RG calls every image edited and RG+RL 91.0\% of the edits and 87.8\% of the photos. Once a reasoning is written, the image hardly matters: AUC falls to 50.6 and 52.2 for RG and to 51.0 and 55.0 for RG+RL.

For each \emph{Seen} image we force the model's answer up to its verdict, \texttt{\textless think\textgreater}\emph{reasoning}\texttt{\textless /think\textgreater}\allowbreak\texttt{\textless answer\textgreater\{"verdict":}, and compare the summed log-probabilities of the two continuations \texttt{"edited"} and \texttt{"authentic"}; their normalized probability is the score, with 0.5 as the threshold. The reasoning is the model's own, or one drawn at random (seed 0) from the reasonings it generated for authentic photos it called authentic or for edits it called edited, always from another source photo. All 3,900 images have a readable reasoning for both models. The full run starts only if, on a 60-image canary, the model's own reasoning reproduces its generated verdict on at least 95\% of the images (RG 100\%, RG+RL 96.7\%).

\section{Rationale}
\label{sec:rationale}

The reasoning that RG, RN and RU are trained to write before their answer is rendered from the verified claims of \cref{sec:data-claims}, so it cannot disagree with the labels. Its form is fixed. It opens with ``I examine the image for signs of manipulation.'', then gives each kept evidence claim as one sentence followed by its box on the $0$--$1000$ grid, then, when the generator also wrote notes that a part looks consistent, one sentence ``Other parts look consistent: \ldots'' listing them with their boxes, then at most one sentence about the whole image (``Areas outside the suspicious region also look regenerated.'' when the verifier saw regeneration outside the regions, or ``There is no visible local blending seam, which is consistent with an editor that regenerates the whole image.'' when no low-level cue survived), and it closes with ``Together these cues indicate a localized edit (\emph{edit type}) in $n$ region(s).'' The structured answer of \cref{sec:task-outputs} follows in a separate \texttt{\textless answer\textgreater} block.

In RG's 13,923 training edits the generator wrote 5.0 claims per edit, of which the verifier and the leak screen kept 3.5 evidence claims (one to five; 1,398 edits keep one, 3,430 keep five) and 1.1 consistency notes; 28\% of the edits lost at least one claim, mostly to the leak screen (3,421 claims) and to the verifier not seeing the cue (793). A kept claim has 31 words on average, 4.6 boxes appear in a reasoning, and a reasoning has 192 words (62 to 327). The most frequent cue families among the evidence claims are texture (13,784 claims), shadow (6,038), lighting (5,125), text rendering (4,345), reflection (3,781), colour (3,229) and anatomy (3,106); 58.6\% of the reasonings carry the regeneration sentence and 10.3\% the no-seam sentence.

An authentic photo gets a reasoning of the same form from the same generator, which sees only the processed photo, is told that it is authentic, and writes two to four specific claims that each name a cue and may give a box; the verifier checks them as it checks the edits' claims. The text ends with ``I find no evidence that the image was edited.'' and the answer holds only the verdict. These reasonings have 123 words on average (39 to 238), and 16.7\% of them place at least one box. The ablation of \cref{sec:supp-ablations} that replaces every photo reasoning by one fixed sentence (``RG without photo rationales'') shows what this text contributes.

\begin{figure*}[t]
    \centering
    \includegraphics[width=\textwidth]{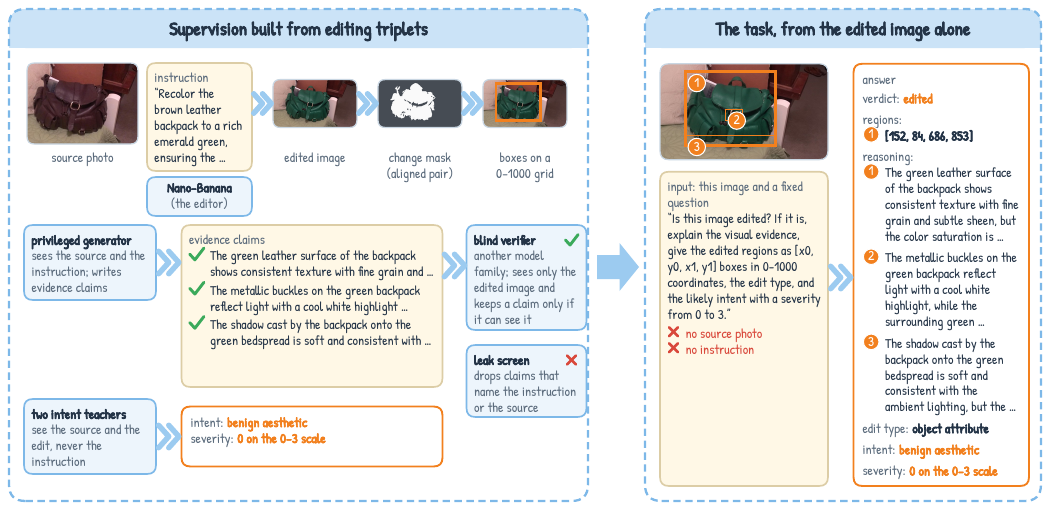}
    \caption{A test edit of the object-attribute type (a backpack recoloured) with its supervision (left) and the training target (right), in the layout of \cref{fig:teaser}. Intent benign aesthetic, severity 0. One generator claim was dropped because its box fell outside the edited region; the fourth kept claim is a consistency note.}
    \label{fig:supp-example-attribute}
\end{figure*}

\begin{figure*}[t]
    \centering
    \includegraphics[width=\textwidth]{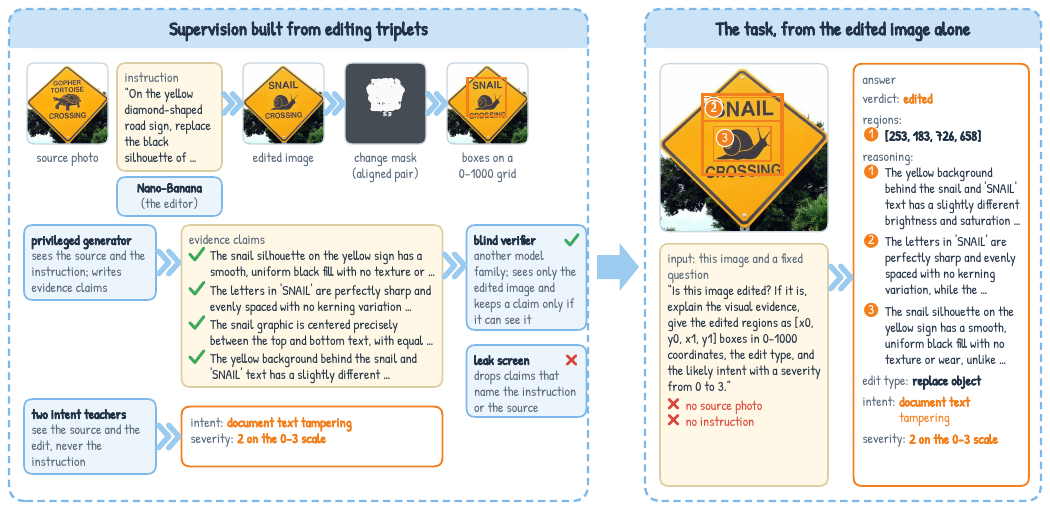}
    \caption{A replace-object edit: the tortoise on a road sign became a snail, with its text. Intent document text tampering, severity 2. Four evidence claims survived; they point at the symbol's flat fill, the lettering and a patch of yellow of different brightness.}
    \label{fig:supp-example-replace}
\end{figure*}

\begin{figure*}[t]
    \centering
    \includegraphics[width=\textwidth]{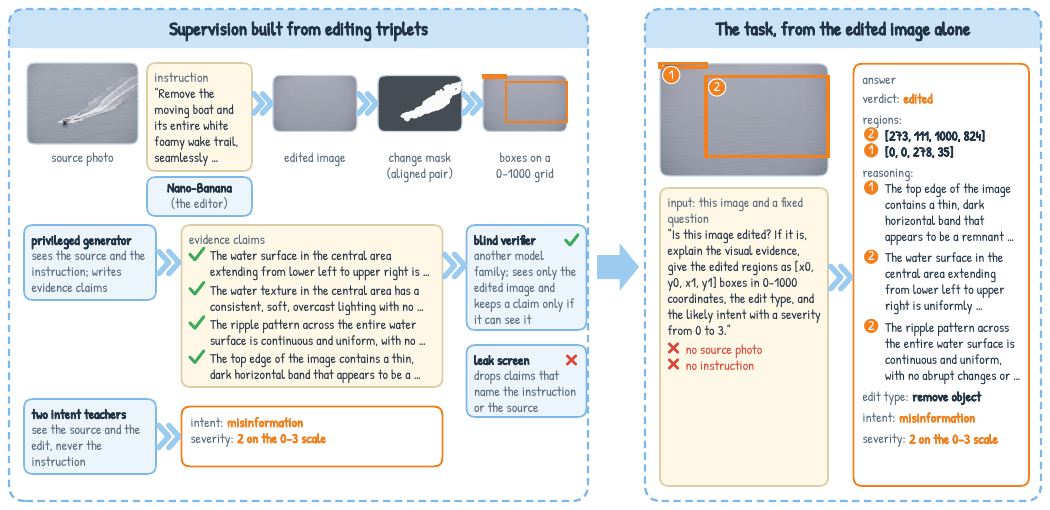}
    \caption{A remove-object edit: a speedboat and its wake removed from a lake. Intent misinformation, severity 2. The removed object leaves nothing to point at, so the claims describe the fill (the uniform water surface) and a band at the top edge; two regions.}
    \label{fig:supp-example-remove}
\end{figure*}

\begin{figure*}[t]
    \centering
    \includegraphics[width=\textwidth]{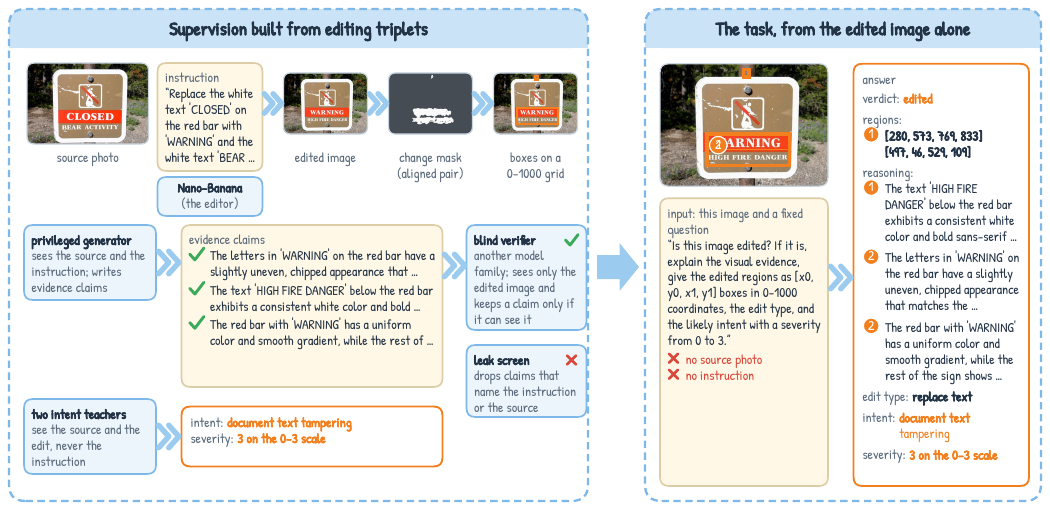}
    \caption{A replace-text edit: ``CLOSED / BEAR ACTIVITY'' on a trail sign rewritten as ``WARNING / HIGH FIRE DANGER''. Intent document text tampering, severity 3. The kept claims are about the lettering and the red bar; two regions.}
    \label{fig:supp-example-text}
\end{figure*}

\Cref{fig:supp-example-attribute,fig:supp-example-replace,fig:supp-example-remove,fig:supp-example-text} show four more test records in the layout of \cref{fig:teaser}, chosen so that the four severity levels of \cref{sec:supp-intent} all appear: a benign recolouring (severity 0), a road sign whose symbol and text were replaced (document text tampering, 2), a boat removed from a lake (misinformation, 2) and a safety notice whose text was rewritten (document text tampering, 3). The pipeline panel of each shows the kept claims and the intent label; the task panel shows the reasoning as the model is trained to write it, with each claim's box numbered on the image.

\section{Further limitations}
\label{sec:supp-limitations}

\paragraph{One editor.} All our training and test edits come from one editor, Nano-Banana. Detectors fine-tuned on them, our VB included, learn a trace that this editor leaves on the whole image and lose most of their accuracy on MagicBrush, whose edits come from DALL-E 2 (\cref{sec:exp-generalization}). Our \emph{Unseen} set holds out edit types, not editors, and on the other editors we test, MagicBrush's DALL-E 2, Qwen-Image-Edit and the 18 of MLLM-Edit (\cref{sec:exp-generalization}), the methods trained on our data reach at most 72.5 AUC, fine-tuned MVSS-Net on Nano Banana Pro aside (84.6), so we make no claim of detection across editors. Google states that Nano-Banana's outputs carry an invisible SynthID watermark~\cite{google2025gemini25flashimage,gowal2025synthidimage}; whether the trace of \cref{sec:exp-generalization} is this watermark, the regeneration itself or both is not settled: under common processing the trace fades for fine-tuned TruFor and VB and stays for fine-tuned MVSS-Net (\cref{sec:supp-trace-wm}). Fine-tuned MVSS-Net detects the edits of Nano Banana Pro, whose outputs Google also watermarks~\cite{google2025nanobananapro}, at 84.6 AUC and those of the other 17 MLLM-Edit editors at 50.4--65.1 (\cref{tab:supp-mllm-edit}); this fits either explanation, since the two Google editors share both the watermark and the model family. We tried to train the trace away by adding 3,000 pairs in which the edit with its region put back from the source is labelled authentic (VB+U, \cref{sec:exp-ablations}); VB+U separates this version from the source as well as VB does (69.1 against 69.9 AUC), so one such run does not remove the trace, although it improves detection on the second editor's edits (70.1 against 66.9 AUC on the Qwen edits, one seed).

\paragraph{Reasoning.} Trained to write their reasoning before the verdict, RU, RN and RG detect edits near chance (\cref{sec:exp-indomain}), and tuning RG with GRPO (RG+RL) raised its true-positive rate but not its AUC. Writing the answer first recovered about 12 of the 40 points (AF), and 34 with half of the training loss on the answer (AF+W, one run), which still leaves the model 5.5 points below SA (\cref{sec:supp-indomain}); in these two models the reasoning comes after the verdict, and we have not run the tests of \cref{sec:exp-grounding} on them. We have not yet found a training recipe in which a model reasons before its verdict and keeps its detection, and removing the photos' rationales does not change this (\cref{sec:exp-ablations}). The verdicts of RG and RG+RL do rest on the regions they cite and on the reasoning itself (\cref{sec:exp-grounding}), but re-encoding an image without any change flips one of RG's verdicts in six, so a verdict that follows its evidence is not yet a reliable one.

\paragraph{Labels and processing.} Masks, boxes, reasoning and intent are produced automatically and checked only on samples (\cref{sec:supp-check,sec:annotation}). Both classes are standardized to the same size tiers and format (\cref{sec:data-preprocessing}), but the authentic photos and the edits of the test sets were not resampled the same way: most photos (77\%) were enlarged into their size tier, every edit was reduced, and only the edits received an extra JPEG compression. Two results bound what this difference can explain. It does not help the released methods, none of which exceeds 58.0 AUC on the \emph{Seen} set (\cref{tab:e1-baselines}). The methods trained on our data could learn it, so we score the existing methods and ours on 480 \emph{Seen} pairs twice, as they appear in the \emph{Seen} set and with the edit and its source photo processed identically (\cref{tab:supp-matched}): the difference explains at most 3 points of the fine-tuned detectors' AUC and at most 7 of our models'. VB is also sensitive to common processing (\cref{sec:supp-robust}): halving each side of an image lowers its AUC on the \emph{Seen} set from 93.1 to 75.5.

\paragraph{Existing methods and ours under identical processing.} We score all methods of \cref{tab:e1-baselines}, IML-ViT and VO on versions S and T of the trace test (\cref{sec:supp-trace}; \cref{tab:supp-matched}). The edits and source photos of its 480 \emph{Seen} pairs are also images of the \emph{Seen} set, so for the same scenes we compare each method's AUC under the test set's processing with its AUC under identical processing. Among the released methods the change lies between $-5.7$ points (PIM, 58.8 to 53.1) and $+3.9$ (TruFor, 44.8 to 48.7). The fine-tuned detectors lose at most 2.6 points (TruFor, 99.3 to 96.7). VB, SA and VO lose 4.0 to 4.7 points and reach 88.2 to 89.1; AF+W loses 6.8 (87.4 to 80.6) and is then 7.6 points below SA, against 4.8 on the same pairs in the \emph{Seen} set. The models that reason before their verdict reach 50.1 to 54.4, and RG is at chance (50.1 [47.4, 52.8]). The unequal processing thus explains a few points of the scores, up to 6.8 for AF+W, and changes none of our conclusions: under identical processing no released method exceeds 56.0 (SIDA), the fine-tuned detectors stay above 96 and our models without reasoning near 88, and reasoning before the verdict still costs almost all of the detection.

\begin{table*}[t]
\centering\footnotesize\setlength{\tabcolsep}{5pt}
\begin{tabular}{l r ccc c}
\toprule
Test set & Edits & VO & VO+NB (control) & VO+ME & VO+ME $-$ VO+NB \\
\midrule
\multicolumn{6}{l}{\textit{MLLM-Edit, editors held out of VO+ME's training}} \\
Five editors, same-size pairs & 277 & 58.9 {\scriptsize[55.6, 62.3]} & 59.1 {\scriptsize[55.9, 62.5]} & 62.7 {\scriptsize[59.6, 65.9]} & $+3.6$ {\scriptsize[+1.2, +6.0]} \\
Five editors, all pairs & 1,362 & 55.6 {\scriptsize[54.3, 56.8]} & 55.6 {\scriptsize[54.4, 56.9]} & 60.2 {\scriptsize[58.8, 61.7]} & $+4.6$ {\scriptsize[+3.8, +5.6]} \\
Six editors without same-size pairs & 1,632 & 53.6 {\scriptsize[52.4, 54.8]} & 53.1 {\scriptsize[52.0, 54.3]} & 57.9 {\scriptsize[56.6, 59.3]} & $+4.8$ {\scriptsize[+3.9, +5.8]} \\
\midrule
\multicolumn{6}{l}{\textit{MLLM-Edit, the seven editors VO+ME trains on (pairs not used in training)}} \\
Same-size pairs & 621 & 52.8 {\scriptsize[50.8, 54.8]} & 53.2 {\scriptsize[51.1, 55.1]} & 58.6 {\scriptsize[56.5, 60.8]} & $+5.5$ {\scriptsize[+3.8, +7.1]} \\
All pairs & 1,899 & 49.4 {\scriptsize[48.4, 50.4]} & 49.6 {\scriptsize[48.5, 50.6]} & 54.9 {\scriptsize[53.7, 56.1]} & $+5.3$ {\scriptsize[+4.5, +6.2]} \\
\midrule
\multicolumn{6}{l}{\textit{Other test sets}} \\
MagicBrush & 528 & 44.3 {\scriptsize[41.4, 47.5]} & 44.3 {\scriptsize[41.5, 47.5]} & 55.5 {\scriptsize[52.0, 59.0]} & $+11.1$ {\scriptsize[+8.8, +13.6]} \\
Second editor, Qwen edits & 390 & 62.3 {\scriptsize[59.7, 64.9]} & 61.4 {\scriptsize[58.7, 64.1]} & 63.0 {\scriptsize[60.5, 65.5]} & $+1.6$ {\scriptsize[+0.7, +2.6]} \\
Second editor, Nano-Banana edits & 390 & 78.4 {\scriptsize[76.1, 80.8]} & 79.2 {\scriptsize[76.9, 81.5]} & 78.7 {\scriptsize[76.4, 81.0]} & $-0.5$ {\scriptsize[-1.2, +0.2]} \\
\emph{Seen} & 1,950 & 93.2 {\scriptsize[92.4, 93.8]} & 93.9 {\scriptsize[93.2, 94.5]} & 93.6 {\scriptsize[92.9, 94.3]} & $-0.2$ {\scriptsize[-0.4, -0.1]} \\
\emph{Unseen} & 450 & 92.6 {\scriptsize[91.4, 93.7]} & 93.2 {\scriptsize[92.1, 94.2]} & 92.8 {\scriptsize[91.5, 93.9]} & $-0.4$ {\scriptsize[-0.7, -0.1]} \\
\bottomrule
\end{tabular}
\caption{\textbf{Training on several editors.} Image AUC with its 95\% bootstrap interval over source photos. VO+ME adds 1,355 pairs of seven MLLM-Edit editors to VO's training set; VO+NB, the control, adds as many further Nano-Banana pairs. Last column: the difference on the same images, with its paired interval. In a same-size pair the edit has exactly its photo's size.}
\label{tab:supp-multi-editor}
\end{table*}

\paragraph{Training on several editors.} All our training edits come from one editor. As a first test of adding others, we train the verdict-only variant VO (\cref{sec:supp-ablations}), with its settings, on its own examples plus 1,355 pairs, an edit and its photo each, of seven editors of the MLLM-Edit release (\cref{sec:supp-mllm-edit}; FLUX.2-klein-Base9B, GPTImage-2-1.5, HunyuanImage3.0, LongCatImage, OmniGenv2, ReveV1 and Step1XEditv1.2, as the release names them). We call this run VO+ME. Its control, VO+NB, adds the same number of further Nano-Banana pairs instead, so the two runs differ only in where the extra pairs come from. We add only pairs whose edit has exactly its photo's size, so that the standardization treats both images alike, and none that shares a photo with our MLLM-Edit test pairs. The other eleven editors are held out: five with same-size pairs, among them Nano Banana Pro and QwenImageEditPlus, and six whose outputs almost never have the photo's size (\cref{tab:supp-multi-editor}).

On the held-out editors VO+ME is a few points above its control: 62.7 against 59.1 AUC on their 277 same-size pairs (paired difference $+3.6$ [1.2, 6.0]), 60.2 against 55.6 on all their 1,362 pairs ($+4.6$ [3.8, 5.6]) and 57.9 against 53.1 on the six editors without same-size pairs ($+4.8$ [3.9, 5.8]). It also gains 11.1 points on MagicBrush (55.5 against 44.3) and 1.6 on the second editor's Qwen edits (63.0 against 61.4). The control stays within 0.9 points of VO on all of these sets, so the gain comes from the other editors' pairs, not from more data, and it costs little on Nano-Banana's edits: 93.6 AUC on \emph{Seen} against the control's 93.9. But the gain is small. No group of other editors' edits exceeds 63.0 AUC; VB and SA reach 61.7 and 62.0 on the held-out same-size pairs with no other editor in training; and at the fixed threshold VO+ME flags 39\% of the photos of those pairs, against 6\% for VO. The 1,355 pairs of seven editors thus move the detector a few points toward other editors and do not make it detect them.

\begin{table*}[t]
\centering\footnotesize\setlength{\tabcolsep}{4pt}
\begin{tabular}{l l cccc c}
\toprule
Training edits & Test edits & T & U & P & K & U $-$ P \\
\midrule
\multicolumn{7}{l}{\textit{TruFor, fine-tuned on}} \\
10,000 Nano-Banana pairs & Nano-Banana & 97.0 {\scriptsize[95.9, 98.0]} & 84.7 {\scriptsize[82.6, 86.8]} & 71.9 {\scriptsize[69.7, 74.2]} & 49.4 & $+12.8$ {\scriptsize[+10.1, +15.6]} \\
 & Qwen & 49.1 {\scriptsize[46.4, 51.8]} & 47.4 {\scriptsize[45.7, 48.9]} & 51.9 {\scriptsize[50.2, 53.6]} & 50.1 & $-4.5$ {\scriptsize[-6.7, -2.5]} \\
450 Nano-Banana pairs & Nano-Banana & 75.6 {\scriptsize[72.9, 78.3]} & 65.4 {\scriptsize[63.4, 67.7]} & 58.3 {\scriptsize[56.6, 60.3]} & 53.2 & $+7.0$ {\scriptsize[+4.6, +9.7]} \\
 & Qwen & 66.1 {\scriptsize[63.6, 68.6]} & 56.4 {\scriptsize[55.1, 58.0]} & 57.8 {\scriptsize[55.7, 59.9]} & 53.7 & $-1.4$ {\scriptsize[-3.8, +1.1]} \\
450 Qwen pairs & Nano-Banana & 55.6 {\scriptsize[53.7, 57.7]} & 53.8 {\scriptsize[52.6, 55.1]} & 53.6 {\scriptsize[52.2, 55.3]} & 51.7 & $+0.2$ {\scriptsize[-1.7, +1.9]} \\
 & Qwen & 80.4 {\scriptsize[77.7, 83.0]} & 63.4 {\scriptsize[61.8, 65.4]} & 67.9 {\scriptsize[65.5, 70.4]} & 51.6 & $-4.5$ {\scriptsize[-6.9, -2.0]} \\
\midrule
\multicolumn{7}{l}{\textit{MVSS-Net, fine-tuned on}} \\
10,000 Nano-Banana pairs & Nano-Banana & 100.0 {\scriptsize[100.0, 100.0]} & 98.8 {\scriptsize[97.8, 99.5]} & 91.1 {\scriptsize[88.9, 93.2]} & 52.3 & $+7.6$ {\scriptsize[+5.3, +10.1]} \\
 & Qwen & 70.1 {\scriptsize[66.9, 73.2]} & 58.8 {\scriptsize[56.5, 61.2]} & 60.9 {\scriptsize[58.0, 64.1]} & 50.7 & $-2.1$ {\scriptsize[-5.7, +1.4]} \\
450 Nano-Banana pairs & Nano-Banana & 84.9 {\scriptsize[82.5, 87.2]} & 70.6 {\scriptsize[68.4, 73.2]} & 65.0 {\scriptsize[62.9, 67.3]} & 51.8 & $+5.7$ {\scriptsize[+3.0, +8.2]} \\
 & Qwen & 73.3 {\scriptsize[70.6, 76.1]} & 59.8 {\scriptsize[58.1, 61.6]} & 63.1 {\scriptsize[60.7, 65.4]} & 52.7 & $-3.3$ {\scriptsize[-5.9, -0.6]} \\
450 Qwen pairs & Nano-Banana & 60.9 {\scriptsize[58.4, 63.5]} & 52.2 {\scriptsize[50.9, 53.7]} & 58.2 {\scriptsize[56.2, 60.4]} & 50.0 & $-6.0$ {\scriptsize[-8.2, -3.7]} \\
 & Qwen & 80.4 {\scriptsize[77.8, 83.0]} & 61.2 {\scriptsize[59.6, 63.2]} & 72.1 {\scriptsize[69.5, 74.6]} & 49.7 & $-10.8$ {\scriptsize[-13.5, -8.0]} \\
\bottomrule
\end{tabular}
\caption{\textbf{TruFor and MVSS-Net trained on a second editor.} Image AUC of the trace test's versions against the source photo S, with its 95\% bootstrap interval over the 296 photos of the second-editor set kept for both editors (\cref{sec:supp-trace,sec:supp-second-editor}). The 10,000-pair detectors are those of \cref{sec:supp-ifdl-ft}; the 450-pair detectors are trained on the edits of one editor and the same photos. Last column: the paired difference on the same photos. All K intervals are within $\pm$1.3 points.}
\label{tab:supp-qwen-ft}
\end{table*}

\paragraph{Training on a second editor.} The whole-image trace of \cref{sec:supp-trace} could come from Nano-Banana alone, for instance from its watermark (\cref{sec:supp-trace-wm}), or from any editor. To tell these apart, we re-edit 500 photos of the forensic detectors' fine-tuning set (\cref{sec:supp-ifdl-ft}; 38 or 39 per edit type) with Qwen-Image-Edit-2509, from the same instructions and with the settings of the second-editor set (\cref{sec:supp-second-editor}); the code we run adds no watermark. We then fine-tune TruFor and MVSS-Net twice with the recipe of \cref{sec:supp-ifdl-ft}: on 450 Qwen edits with their photos, and on the Nano-Banana edits of the same 450 photos, keeping 50 further photos for validation. The photos and both editors' edits are standardized together, so the two runs differ only in the editor. We score the four detectors with the trace test built on the second-editor set, whose photos are not in the fine-tuning set: for every photo outside scene background we make the five versions of both editors' edits, with masks from the data pipeline's alignment and mask generator, and keep 296 photos whose pairs of both editors pass the filters of \cref{sec:supp-trace} (\cref{tab:supp-qwen-ft}).

Qwen's edits leave a trace the detectors learn: trained on 450 Qwen pairs, TruFor and MVSS-Net separate the Qwen edits from their photos at 80.4 AUC each, against 75.6 and 84.9 for the same detectors trained and tested on Nano-Banana. Part of this trace lies outside the edited region: the Qwen-trained detectors find U at 63.4 and 61.2, well above the seam control K (51.6 and 49.7). They find P more easily still (67.9 and 72.1), whereas on Nano-Banana's edits U is the easier version, but most of this contrast comes from the size of the region, which is larger for Qwen (median 32\% of the image against 23\%) and so gives P more edited pixels and U fewer untouched ones. On the 77 photos whose two regions differ by at most 5\% of the image, the Qwen-trained detectors find U about as well as the Nano-Banana-trained ones find Nano-Banana's (65.1 and 61.7 against 63.3 and 66.9) and P somewhat more easily (68.3 and 69.0 against 58.9 and 66.4). Each trace belongs to its editor: the Qwen-trained detectors score Nano-Banana's edits at 55.6 and 60.9, and the Nano-Banana-trained ones score Qwen's at 66.1 and 73.3. An editor whose output carries no watermark thus also leaves a whole-image trace that fine-tuned detectors learn, at this scale about as readily as Nano-Banana's, so a watermark is not needed to explain such a trace, though Nano-Banana's may add to it. Each detector was trained once, and on 450 pairs instead of the 10,000 of \cref{sec:supp-ifdl-ft}, with which the Nano-Banana trace is far easier to find (U at 84.7 and 98.8); how the two editors' traces compare at that scale remains open.

\section{Human Audit Protocol}
\label{app:human-audit}

This appendix specifies the judgments collected in the human audit
and the evidence available to reviewers. The three tasks address
different annotation errors: inaccurate spatial coverage, unsupported
forensic explanations, and intent labels that exceed what an image can
establish. Their criteria are defined separately so that reliability in
one field is not taken as evidence of reliability in another.

\subsection{Mask Localization Quality}
\label{sec:audit-mask}

Instruction-based editing can alter the appearance of both the target region and its surroundings, so differences between the image pair do not necessarily identify the intended edit. An automatic mask may cover the main target while also including background content outside the modification. Such errors enter both the localization supervision and the reference boxes derived from the mask. Reviewers compare the image pair and editing instruction to assess whether the mask covers the actual modification, distinguishing correct localization of the main change from accurate spatial coverage. We report the proportions rated acceptable, partly acceptable, and unacceptable, together with results by tier to expose coverage errors that an aggregate score may obscure.

\subsection{Evidence in Reasoning}
\label{sec:audit-reasoning}

A generated explanation may identify the edited object correctly while justifying its conclusion through nonexistent texture, shadow, or boundary anomalies. Agreement from another model does not establish that these claims are visually observable. Our human audit evaluates each explanation as a whole, asking whether its main judgment is supported by specific visual evidence. The source image helps establish what changed, but the forensic evidence cited in the explanation must be observable in the edited image itself. The audit thus distinguishes knowledge of the modification from a supported account of why it appears edited, rather than checking only whether the explanation matches the instruction. The effect of reasoning supervision on model performance is evaluated separately through the ablations in Section~7.

\subsection{Plausibility of Intent}
\label{sec:audit-intent}

Unlike an edited region, an edit's purpose is rarely determined uniquely by the image. Removing an object may simplify a composition or change how a viewer interprets an event; without the context of use, an automatic label may present one possibility as an established purpose. We therefore evaluate intent along two complementary dimensions: the most plausible \emph{purpose} of the edit and its potential \emph{severity} if the edited image is treated as authentic. Purpose is a nominal category describing the likely use of the modification, whereas severity is an ordinal assessment of its possible adverse impact. Neither is treated as a statement of the creator's actual motivation. Reviewers infer purpose from the discernible change and assign severity under the assumption that the image is believed to be real. They may abstain when the available evidence is insufficient, preserving uncertainty that the image alone cannot resolve. We report agreement on purpose and severity separately, along with the proportion of ratings for which reviewers abstain.

\subsection{Audit Samples and Review Setup}
\label{app:audit-setup}

The review interface contains separate pools of 200 masks, 200 reasoning
annotations, and 200 intent examples. These are review-pool sizes, not
counts of completed judgments. The pools need not share sample
identifiers, since each audit evaluates a different annotation field.
The current mask pool comprises 50 examples from each of four supplied
strata: local editing (L), background replacement (R), person
stylization (S), and outpainting (X). Results from this pool must be
distinguished from results restricted to the L/R scope of the main task.
Equal allocation across these strata does not estimate the source
dataset's natural mixture without weighting.

Reviewers consult task-specific criteria and learning examples before
rating the formal samples. The interface displays model-generated
assessments as reference material but does not expose another human
reviewer's answers. The resulting judgments are therefore collected
under model-assisted review conditions. An optional ambiguity flag
records cases requiring further inspection without replacing the
task-specific answer. The number of completed judgments and the
coverage of repeated human review will be reported with the audit
results.

\subsection{Mask Rating Criteria}
\label{app:audit-mask-criteria}

Reviewers read the editing instruction, compare the original and edited
images, and inspect the mask overlay. The instruction identifies the
requested operation; the image pair establishes what actually changed.
The rating concerns whether the mask covers that modification rather
than whether it merely overlaps the target object.

An \emph{acceptable} mask covers the main modified areas without
obvious excess or omissions. A \emph{partly acceptable} mask locates the
main edit but includes noticeable excess coverage or misses part of the
modified area. An \emph{unacceptable} mask misses the main edit or
predominantly covers unchanged content.

The expected coverage depends on the operation. For local edits,
reviewers inspect the modified object, text, or attribute; object removal
includes the filled area, and object movement requires attention to
both its old and new positions. For background replacement, the mask
should cover the replaced background while excluding retained foreground
subjects. For person stylization, its extent should follow the body
parts actually changed. For outpainting, reviewers locate the original
image within the expanded canvas and check coverage of the newly added
area. These judgments assess coverage qualitatively and do not create
pixel-level reference masks.

\subsection{Reasoning Rating Criteria}
\label{app:audit-reasoning-criteria}

Reviewers assess the complete generated explanation against the image
pair and the accompanying region information. They check whether the
described objects and changes are correct, whether the cited visual
details are observable, and whether those details support the editing
judgment. The original image establishes the change; forensic claims
about the edited image must be verifiable in that image itself.
Mask overlays and region coordinates assist inspection but are not
treated as independently validated evidence.

An \emph{acceptable} explanation correctly describes the main change
and supports its central judgment with observable evidence, without
obvious errors. A \emph{partly acceptable} explanation contains relevant
evidence but also secondary inaccuracies, exaggeration, or unsupported
claims. An \emph{unacceptable} explanation misidentifies the main change
or relies on central claims that contradict the images or lack visual
support. Each sample receives one overall rating, rather than separate
scores for every evidence statement.

Correct localization alone is insufficient: an explanation can name the
edited object while inventing its shadows or texture. Conversely, a
visible feature is not necessarily evidence of manipulation. For
example, smooth texture or the absence of a shadow must be interpreted
in the depicted scene rather than treated as an editing artifact by
default. Material that cannot be inspected is not automatically assigned
an unfavorable quality rating.

\subsection{Intent Assessment Criteria}
\label{app:audit-intent-criteria}

Reviewers select one plausible primary purpose and independently assess
potential negative impact. The judgment concerns a possible use of the
edit, not the creator's actual motivation. The purpose options are
defined as follows:

\begin{itemize}
    \item \emph{Aesthetic enhancement}: improving appearance, style,
    or composition.

    \item \emph{Commercial use}: an edit with identifiable advertising,
    product-presentation, or promotional cues.

    \item \emph{Privacy protection}: concealing identifying or sensitive
    information.

    \item \emph{Misinformation}: altering the depiction of an event or
    scene in a way that could mislead a viewer about the facts.

    \item \emph{Document/text tampering}: changing the content or meaning
    of records, documents, or written information.

    \item \emph{Identity manipulation}: changing who a person appears
    to be.

    \item \emph{Humor/memes}: an edit with discernible parody,
    exaggeration, or entertainment cues.
\end{itemize}

An additional \emph{uncertain/cannot determine} option applies when the
evidence is insufficient, competing purposes cannot be distinguished,
or an overlay obscures the relevant change. A depicted product alone
does not establish commercial purpose; decorative text is not
automatically document tampering; ordinary appearance changes do not
necessarily manipulate identity. This single-choice assessment concerns
the primary purpose, not the full set of potentially compatible intents.

Potential negative impact is rated under the assumption that the edited
image is believed to be real: 0 indicates no obvious negative impact;
1 indicates limited misunderstanding or inconvenience; 2 indicates a
potentially substantive effect on judgments about a person, event, or
record; and 3 requires a concrete basis for severe deception or harm.
Reviewers may also select \emph{cannot determine}. Purpose does not
determine severity, and a hypothetical harmful scenario alone does not
justify a high rating.

\subsection{Review Interface}
\label{app:audit-interface}

\begin{figure*}[t]
    \centering
    \includegraphics[width=\textwidth]
        {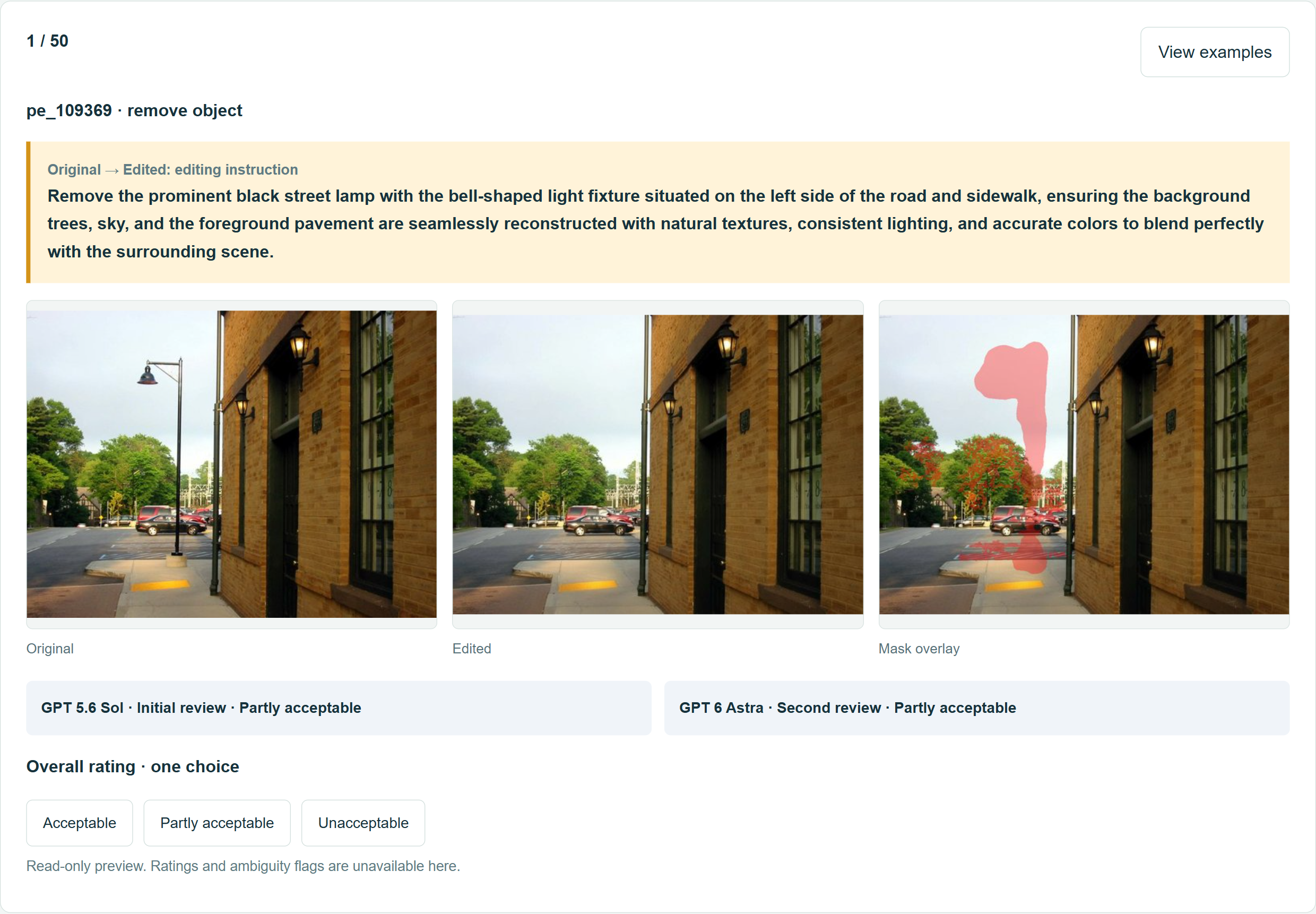}
    \caption{
    English read-only presentation of the mask review task.
    The image pair and editing instruction establish the modification;
    the overlay is inspected against the coverage criteria.
    Model assessments are shown as references, not human judgments.
    }
    \label{fig:audit-mask-interface}
\end{figure*}

\begin{figure*}[t]
    \centering
    \includegraphics[width=\textwidth]
        {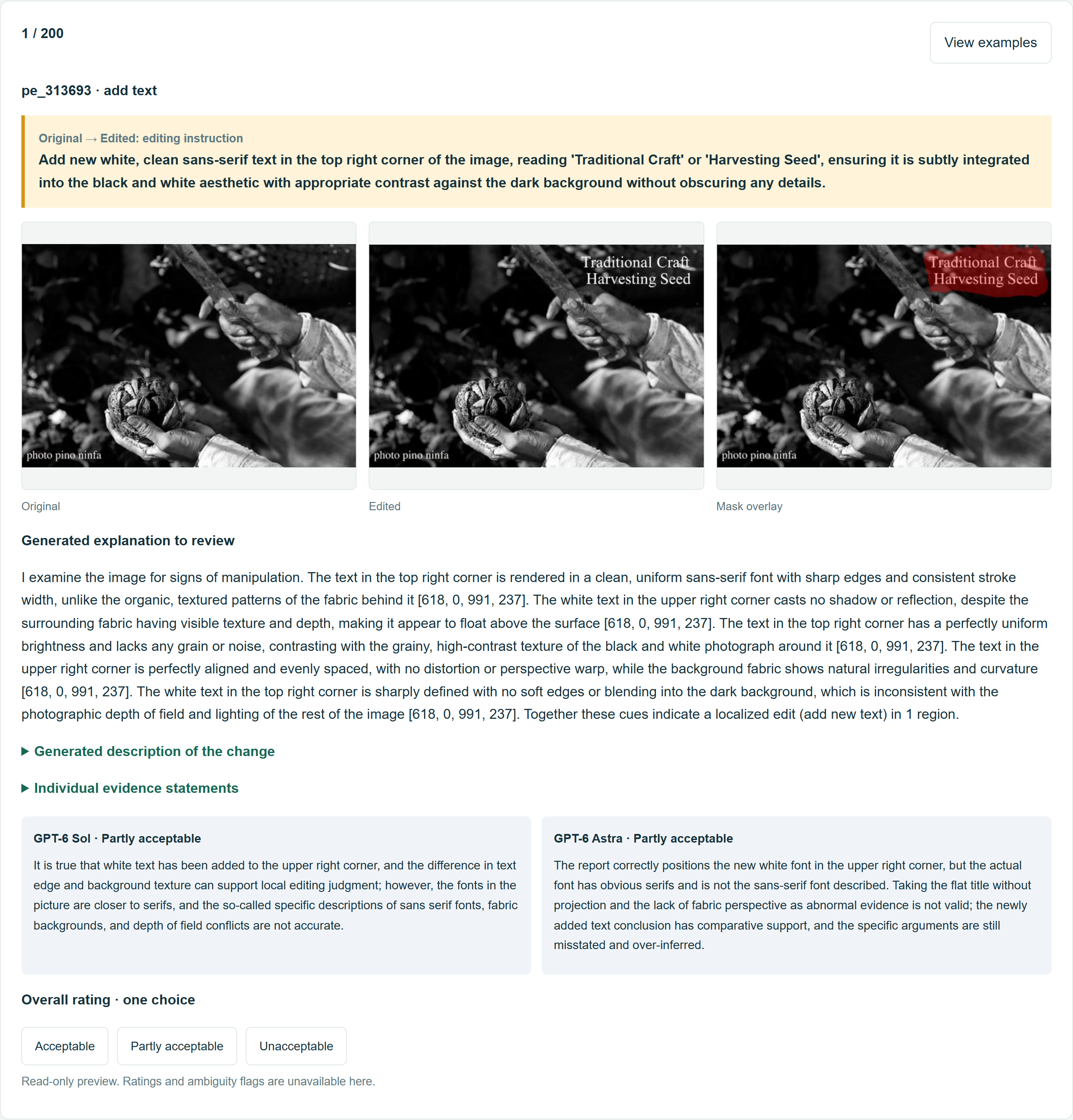}
    \caption{
    English read-only presentation of the reasoning review task.
    The complete generated explanation is shown with the image pair
    and mask overlay. Reviewers assess its overall evidential support;
    the displayed explanation and model opinions are not endorsed
    by this illustration.
    }
    \label{fig:audit-reasoning-interface}
\end{figure*}

\begin{figure*}[t]
    \centering
    \includegraphics[width=\textwidth]
        {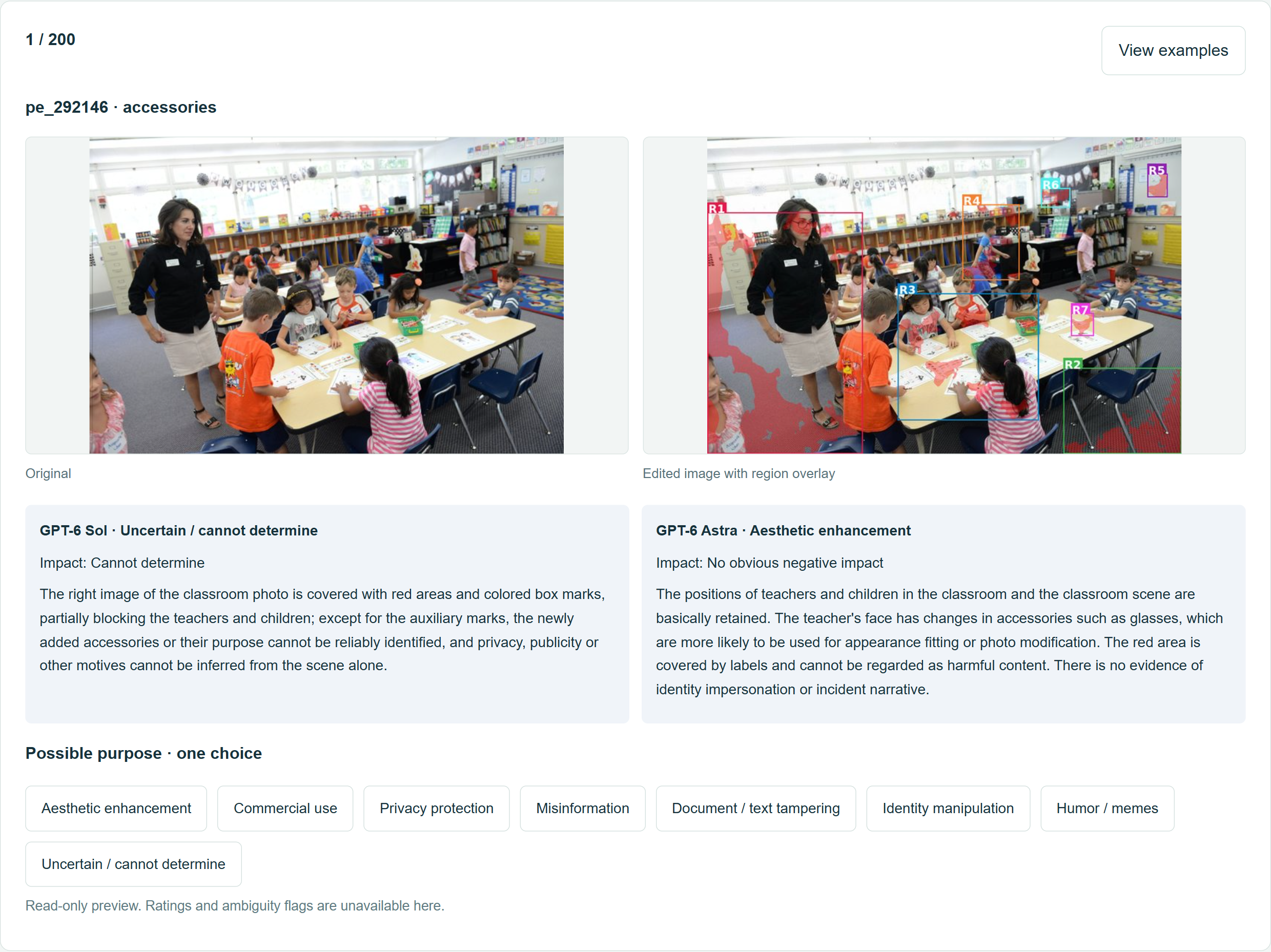}
    \caption{
    English read-only presentation of the intent task, including
    the supplied image pair, model opinions, and possible-purpose
    options. The right-hand guide lists the impact levels.
    The operational interface records primary purpose and impact
    separately, including abstentions; this preview does not expose
    all input controls.
    }
    \label{fig:audit-intent-interface}
\end{figure*}

Figures~\ref{fig:audit-mask-interface}--\ref{fig:audit-intent-interface}
show the English read-only presentation of the review materials.
The annotation interface uses Chinese instructions and records answers
separately for each reviewer. The English presentation is provided for
inspection and does not accept ratings. Its screenshots retain the
generated annotations and model assessments as displayed; these are
objects of review or reference opinions, not adjudicated human answers.
The operational interface additionally supports an ambiguity flag and
exports of saved judgments. No human review outcomes are inferred from
the examples shown here.

\end{document}